\documentclass[10pt,journal,compsoc]{IEEEtran}
\ifCLASSOPTIONcompsoc
  \usepackage[nocompress]{cite}
\else
  \usepackage{cite}
\fi

\ifCLASSINFOpdf
\else
\fi

\ifCLASSOPTIONcompsoc
  \usepackage[caption=false,font=footnotesize,labelfont=sf,textfont=sf]{subfig}
\else
  \usepackage[caption=false,font=footnotesize]{subfig}
\fi

\usepackage{color}
\PassOptionsToPackage{sort}{natbib}
\usepackage{amsmath, amsthm, amssymb,upgreek}
\usepackage{algorithm,algorithmicx,algpseudocode}
\usepackage{graphicx}
\usepackage{makecell}
\usepackage{multirow}
\usepackage{pifont}
\usepackage{ifthen}

\usepackage{tikz}
\newcommand*{\circled}[1]{\lower.7ex\hbox{\tikz\draw (0pt, 0pt)%
    circle (.5em) node {\makebox[1em][c]{\small #1}};}}  %circled number
\usepackage{ragged2e}
\newtheorem{definition}{Definition}[section]
\newtheorem{theorem}{Theorem}[section]
\newtheorem{assumption}{Assumption}[section]

\newtheorem{remark}[theorem]{Remark}

\newtheorem{lemma}[theorem]{Lemma}

\begin{document}
\title{Towards Efficient and Stable K-Asynchronous Federated Learning with Unbounded Stale Gradients on Non-IID Data}

\author{Zihao~Zhou,Yanan Li, Xuebin Ren, and Shusen Yang % <-this % stops a space
\IEEEcompsocitemizethanks{\IEEEcompsocthanksitem Z. Zhou, Y. Li and S. Yang are with School of Mathematics and Statistics, Xi'an Jiaotong University, Xi'an, Shaanxi 710049, China. \protect\\
E-mails: \{wszzh15139520600, gogll2,shusenyang\}@stu.xjtu.edu.cn.
\IEEEcompsocthanksitem X. Ren is with School of Computer Science and Technology, Xi'an Jiaotong University , Xi'an, Shaanxi 710049, China. \protect E-mail: xuebinren@mail.xjtu.edu.cn.}
%\IEEEcompsocthanksitem J. Doe and J. Doe are with Anonymous University.}
}

\IEEEtitleabstractindextext{
\justifying  %Align the ends of the abstract text
\begin{abstract}
  Federated learning (FL) is an emerging privacy-preserving paradigm that enables multiple participants collaboratively to train a global model without uploading raw data. Considering heterogeneous computing and communication capabilities of different participants, asynchronous FL can avoid the stragglers effect in synchronous FL and adapts to scenarios with vast participants. Both staleness and non-IID data in asynchronous FL would reduce the model utility. However, there exists an inherent contradiction between the solutions to the two problems. That is, mitigating the staleness requires to select less but consistent gradients while coping with non-IID data demands more comprehensive gradients. To address the dilemma, this paper proposes a two-stage weighted $K$ asynchronous FL with adaptive learning rate (WKAFL). By selecting consistent gradients and adjusting learning rate adaptively, WKAFL utilizes stale gradients and mitigates the impact of non-IID data, which can achieve multifaceted enhancement in training speed, prediction accuracy and training stability. We also present the convergence analysis for WKAFL under the assumption of unbounded staleness to understand the impact of staleness and non-IID data. Experiments implemented on both benchmark and synthetic FL datasets show that WKAFL has better overall performance compared to existing algorithms.
\end{abstract}

}

\maketitle

\IEEEdisplaynontitleabstractindextext

\IEEEpeerreviewmaketitle

\IEEEraisesectionheading{\section{Introduction}\label{sec:introduction}}

\IEEEPARstart{T}{he} availability of massive data has been the bottleneck of many machine learning (ML) algorithms, especially in many deep learning (DL) scenarios, such as video surveillance \cite{chen2019distributed}, and speech recognition \cite{deng2013recent}. In the 5G era, more and more data are being generated and scattered at massive users' devices (i.e., clients) locally. The aggregation of these distributed data promises a significant boost in the quality of ML models, prospering various artificial intelligence services. However, with the increasing awareness of data privacy, people are unwilling to share their own data, which would impede the DL development \cite{chen2018bi} (e.g., especially for data sensitive fields like autonomous driving \cite{chen2015deepdriving} and disease detection \cite{esteva2017dermatologist}).
%the lack of data protection in fields like autonomous driving \cite{chen2015deepdriving} and disease detection \cite{esteva2017dermatologist} has rolled back the frontiers unconsciously.
To address this issue, a novel distributed learning paradigm, federated learning (FL) \cite{mcmahan2016communication}, \cite{9141436}, \cite{9195793}, \cite{liu2020accelerating} has been proposed to achieve data utilization from massive clients while not seeing their local data.
%Throughout the remainder of the paper, we use clients to denote the participants such as mobile users or corresponding devices.

Generally, FL \cite{yang2019federated}, \cite{kairouz2019advances} enables multiple clients with decentralized data to collaboratively train a shared model under the concerted control of a center server (e.g. service provider), which complies with privacy regulations such as GDPR \cite{voigt2017eu}. By providing an effective framework to exploit the naturally generated data while respecting privacy, FL has not only attracted extensive interests from the academia \cite{liu2020accelerating}, but also been deployed in various practical applications, such as e-health diagnosis \cite{sheller2020federated}, \cite{brisimi2018federated}, fraud detection \cite{yang2019ffd}, and recommendation systems \cite{qi2020fedrec}, \cite{malle2017more}.

Many algorithms have been proposed for FL \cite{9098045}, \cite{li2019convergence}, \cite{wu2020safa}, among which, FedAvg is one of the most classical FL algorithms and runs in a synchronous manner \cite{mcmahan2016communication}. At each iteration, the clients complete several local training processes and upload the model parameter. The server randomly samples a portion of clients and aggregates their locally updated models. In such a case, the runtime of each FL iteration is determined by the stragglers \cite{cipar2013solving}, \cite{tandon2017gradient} who are the sampled slowest clients.
To alleviate the stragglers effect, the server can update the global model once receiving the fastest $K$ gradients out of total clients, namely, $K$-sync FL \cite{dutta2018slow} or $K$-batch-sync FL \cite{dutta2016short}. However, these approaches can not eradicate stragglers due to the unpredictable statuses of sampled clients and network communication.

Besides the above synchronous methods, the $K$ asynchronous FL ($K$-async FL) has also been extensively studied \cite{xie2019asynchronous}, \cite{lu2019differentially}, \cite{chen2019asynchronous}, \cite{li2019asynchronous}, in which the server updates the global model once receiving the gradients from the first $K$ clients and those clients who fail to participate in current iteration can continue their training for reducing the runtime in the next iteration. $K$-async FL can not only alleviate stragglers, but also save the total training time when the iteration time of clients follows a \textit{new-longer-than-used distribution} \cite{dutta2018slow}. Apart from that, Hannah et al. \cite{hannah2017more} has further pointed out that asynchronous FL (AFL) allows more iterations within the same time compared to synchronous FL. Therefore, compared to $K$-sync FL, $K$-async FL has demonstrated great benefits of achieving more efficient learning in highly heterogeneous systems. In this paper, we focus on achieving both efficient and high-utility $K$-async FL.

Though $K$-async FL can mitigate the stragglers effect and save total training time \cite{dutta2018slow}, there are still two obstacles to be overcome in practice.
On the one hand, non-IID datasets generated at different FL clients can impact the model utility \cite{zhao2018federated}, \cite{li2019gradient}.
On the other hand, stale gradients may harm the model utility, or even diverge the training process \cite{dai2018toward}, \cite{zhang2015staleness}.
The solutions to these two problems have been studied separately. For non-IID data, the essence of the existing solutions, e.g., momentum \cite{li2019gradient}, \cite{karimireddy2020scaffold} and variance reduction \cite{liang2019variance}, is to fully utilize all available information for estimating the global data distribution. Hence, gradients from as many clients as possible need to be aggregated to make the aggregated gradients represent the overall data comprehensively. For staleness, most studies pointed out that the server should aggregate the received gradients \cite{chen2019communication}, \cite{xie2019asynchronous} or adjust the learning rate \cite{zhang2015staleness}, \cite{xie2019local} negatively correlated with the staleness. Therefore, only a few low-stale gradients are maintained and most of the high-stale gradients will be filtered out.
%\emph{On the one hand}, to reduce the influence of non-IID data, the essence lies in the better utilization of all available information to estimate the global information, such as momentum \cite{li2019gradient}, \cite{karimireddy2020scaffold} and variance reduction \cite{liang2019variance}. Therefore, gradients from as many clients as possible need to be aggregated to make the aggregated gradients represent the overall data comprehensively.
%\textit{On the other hand}, regarding the influence of staleness, most studies pointed out that the server should aggregate the received gradients \cite{chen2019communication}, \cite{xie2019asynchronous} or adjust the learning rate \cite{zhang2015staleness}, \cite{xie2019local} negatively correlated with the staleness. Therefore, only a few low-stale gradients are maintained and most of the high-stale gradients will be filtered out.
Apparently, there will pose an essential contradiction when simply combining the existing methods to mitigate the impact of non-IID data and stale gradients. To better help comprehend the contradiction, we explore the interplay of mitigation strategies for non-IIDness and staleness on EMNIST MNIST \cite{cohen2017emnist}  in Section \ref{motivation}. The experimental results show that the mitigation effect for non-IIDness is negatively correlated with the mitigation effect for staleness, which indicates their inherent contradiction. Therefore, it is of significance to design a novel asynchronous FL approach that can effectively deal with both stale gradients and non-IID data.

Though stale gradients may cause the model to diverge, Mitliagkas et al. \cite{mitliagkas2016asynchrony} found that stale gradients can help converge faster and they leveraged staleness to accelerate the training process. However, not all gradients can speedup convergence. In fact, low-stale gradients are more likely to have a consistent descent direction and accelerate the training process. On the contrary, high-stale gradients may have different directions and the biased gradients will prevent model convergence. %will cause the model to diverge and only consistent gradients can accelerate the training process.
This motivates us to pick the consistent gradients, instead of only utilizing the low-stale gradients.
Therefore, based on the potential advantage of consistent gradients and treatments of non-IID data, we propose a two-stage weighted $K$-async FL algorithm (WKAFL) with adjusting the learning rate to improve the prediction accuracy, training speed and stability.
Our main contributions are as follows:
\begin{enumerate}
  \item We propose WKAFL, a novel $K$-async FL algorithm with two stages, which improves the model utility of asynchronous algorithm and shows good robustness in non-IID settings. In stage one, stale gradients with consistent descent direction will be picked to accelerate the training process. In stage two, stale gradients with large norm will be clipped to stabilize the model. In both stages, the server adjusts the learning rate according to the least staleness of $K$ gradients and accumulates selected consistent gradients to further mitigate the non-IIDness.
  \item We present the convergence analysis for the non-convex optimization problem in AFL to analyze the impact of staleness and non-IID data, under both the assumptions of bounded and unbounded staleness. The analytical results (Theorem \ref{stageone} and Theorem \ref{stagetwo}) validate that both staleness and non-IID data can decrease model utility.
  %In stage one, the analysis is based on the assumption of bounded staleness while in stage two, we provide analysis under the assumption of unbounded staleness   for the reason that the number of iterations in stage one is limited and the maximal staleness is smaller than the total iterations of stage one. In stage two, we provide analysis under the assumption of unbounded staleness due to the possible infinite number of iterations.
  \item We conducted extensive experiments on four federated learning datasets, including CelebA, EMNIST ByClass, EMNIST MNIST and CIFAR10 to validate the performance of WKAFL in terms of training speed, prediction accuracy and training stability. Ablation experiments indicate that each component is indispensable for WKAFL and a contradiction exists between mitigation strategies for staleness and non-IIDness. Comparison experiments show that WKAFL has the best overall performance, especially when the number of clients is large. In particular, WKAFL achieves up to 19.41\% accuracy gain on CIFAR10 and 92.8\% training stability improvement on EMNIST MNIST compared with existing staleness-aware algorithms. Besides, compared with existing non-IID mitigation algorithm on CIFAR10, WKAFL achieves 5.83-10.59\% accuracy gain while maintaining a similar training stability.
  %although WKAFL has the same training stability amplitude, it achieves 5.83-10.59\% accuracy gain.
\end{enumerate}

The rest of the paper is organized as follows. Section \ref{relatedwork} describes the related work on both staleness and non-IID data in AFL. Section \ref{preliminary} presents the system model and problem definition. Section \ref{WKAFL} describes the details of our proposed algorithm WKAFL. The corresponding theoretical analysis is presented in Section \ref{analysis}. Section \ref{experiment} shows the experiment results for validating the performance of WKAFL. Finally, we conclude the paper in Section \ref{conclusion}.

\section{Related Work}\label{relatedwork}
%Although $K$-async FL can alleviate the impact of stragglers and save training time, it still faces many challenges, such as privacy preservation, system framework, and scalability. In this paper, we focus on non-IID data and staleness.
In this section, we introduce the related work of FL in terms of non-IID data and staleness.
\subsection{Non-IID Data}

Non-IID data is an essential characteristic of FL. However, since most of the well-established statistical theories are based on the IID data assumption, there are only a few analytical results about convergence under the non-IID setting. Some empirical results have shown the negative influence of non-IID data on the model utility. For example, Zhao et al. \cite{zhao2018federated} empirically showed that the accuracy of FedAvg decreases significantly with the increase of data heterogeneity. Besides, Li et al. \cite{li2018federated} proved that non-IID data can decrease the model utility. They assumed that there is a bounded difference between the gradients uploaded by clients and global unbiased gradients to guarantee the convergence, similar to \cite{li2019convergence}, \cite{khaled2019first}, \cite{wang2019matcha}. In this paper, we follow this assumption and provide a convergence analysis under the non-IID setting.
%Given the aforementioned impediment to full achievement of non-IID data,
To mitigating the impact of non-IID data, a number of methods have been proposed to utilize all available information to estimate the global knowledge of the decentralized non-IID datasets.
For example, in \cite{zhao2018federated}, the server broadcasts some non-private data to all the clients to make the gradients to carry more common information. Apart from data sharing, each received gradient can also accumulate historical information by applying momentum \cite{li2019gradient}, \cite{karimireddy2020scaffold} or variance reduction \cite{liang2019variance} to ensure the aggregated gradient to be more representative of the global information. Li et al. \cite{li2019gradient} proposed GSGM which divides the whole training process into multiple rounds. In each round, the server eliminates the diversity of aggregated gradients by using a global momentum, which is accumulated by part of the gradients based on the scheduling strategy.

Since momentum can also help converge faster, WKAFL in this paper also takes advantages of the momentum to alleviate the impact of non-IID data as GSGM. However, WKAFL is essentially different from GSGM. Since GSGM needs all the clients to upload gradients, GSGM has a low scalability due to computational overhead in each round, while WKAFL adapts K-async FL and behaves robustly.
%Another is, GSGM only takes the model stability into consideration without theoretical analysis. Moreover, WKAFL can improve the training speed with theoretical analysis for convergence rate.

\subsection{Staleness}
Staleness is one of the conspicuous challenges in AFL. Dai et al. \cite{dai2018toward} found that stale gradients will slow down the training process and decease prediction accuracy. To guarantee model utility, most of existing methods make full use of low-stale gradients and prevent stale gradients to bring down the model utility.
One effective path is to adjust the learning rate \cite{zhang2015staleness}, \cite{mcmahan2014delay}. In \cite{zhang2015staleness}, Zhang et al. tuned the learning rate on a per-gradient basis inversely proportional to the staleness. However, when there are massive heterogeneous clients, the staleness may be very high, which leads to very small learning rate and extremely slow training.
%the training speed is a natural concomitant.
In \cite{mcmahan2014delay}, McMahan et al. extended the adaptive gradient methods \cite{mcmahan2010adaptive}, \cite{duchi2011adaptive} to the asynchronous parallel setting and revised the learning rate based on previous gradient steps.
Another feasible method is to aggregate the gradients with different weights negatively correlated with the staleness. Xie et al. \cite{xie2019asynchronous} and Chen et al. \cite{chen2019communication} aggregated the stale gradients based on exponential weights $(e/2)^{-\tau}$ and $e^{\tau}$ respectively, where $\tau$ represents the staleness.

To alleviate the effect of staleness, WKAFL adopts the exponential weighting method, which is the same as \cite{chen2019communication} for the reason of ideal performance in asynchronous scenarios. However, all above algorithms either provide no theoretical analysis, or need the assumption of bounded staleness \cite{lian2015asynchronous}, \cite{lee2015bssync} which has less practicality in many AFL scenarios \cite{hannah2018unbounded}. In this paper, WKAFL can provide a convergence analysis under the assumption of unbounded staleness.

%To better alleviate the effect of non-IID data, the server requires a greater demand of gradients. However, to prevent the stale gradients from bringing down the model utility, only confined low-stale gradients are suggested to participate in the aggregating process.

%It will bring out a new contradictory about the quantity of aggregated gradients when simply combining the corresponding mitigation methods concerning about non-IID data It will bring out a new contradictory about the quantity of aggregated gradients when simply

\subsection{Interplay of Non-IID Data and Staleness}
%In this paper, we aim to mitigate the impacts of both non-IID data and staleness. A straightforward idea is to simply combine the existing methods for non-IID data and staleness, which, however, are conflicting in choosing the aggregated gradients.
To better alleviate the impact of non-IID data, the server requires a greater number of gradients. However, to prevent the stale gradients from bringing down the model utility, only confined low-stale gradients are favorable in the aggregation.
Therefore, a contradiction about the quantity of aggregated gradients occurs when simply combining the corresponding mitigation methods concerned about non-IID data and staleness. Facing the same contradiction, the proposed WKAFL aims to exploit the idea of gradient selection to mitigate it. The idea of gradient selection has been explored in~\cite{xie2020zeno++}, which proposes to pick high-quality gradients for tackling with Byzantine attacks. Different from WKAFL, ZENO++ in \cite{xie2020zeno++} requires the central server to have some high-quality data and relies on the assumption of IID data, which is impractical in the federated learning scenario. Particularly, with non-IID datasets, it has much poorer performance than WKAF, as shown in Appendix \ref{Appendix:ZENO}.

%that can represents the distribution of global data, thus relying on the assumption of IID data among

%\cite{xie2020zeno++} proposed ZENO++ to improve the model utility by picking high-quality gradients. Different from WKAFL, ZENO++ focuses on tackling with byzantine workers. Apart from that, The server must collect some high-quality data for ZENO++, which is infeasible in federated scenarios, as shown in Appendix \ref{Appendix:ZENO}. However, WKAFL does not have these restrictions.

Despite the adverse impact of staleness on model utility, stale gradients can play a positive role in accelerating the training process. In \cite{mitliagkas2016asynchrony}, Mitliagkas et al. demonstrated that the stale gradients could have the same effect as momentum when the staleness follows a geometric distribution. %However, they ignored the adverse effect of stale gradients.
In fact, consistent stale gradients can boost the convergence, analogous to the function of a large momentum \cite{mitliagkas2016asynchrony}, \cite{li2019asynchronous}.
However, when the model is close to the optimal solution, the momentum may fluctuate the training process.
To prevent this fluctuation, WKAFL uses the two-stage strategy as \cite{chen2019efficient} to fully utilize the stale gradients. In \cite{chen2019efficient}, the two-stage strategy is used to control discrepancies between the global model and stale models, which is different from the aim of utilization of stale gradients in WKAFL.

\section{Preliminary}\label{preliminary}
\subsection{Stochastic Optimization}
Generally, the aim of a machine learning problem is to minimize the empirical risk function $F(w)$:
\begin{IEEEeqnarray}{c}
  \min_w F(w) =\frac{1}{N} \sum_{\xi_{i}\in \mathcal{D}} f(w,\xi_i), \notag
\end{IEEEeqnarray}
where $\mathcal{D} = \{\xi_1,\xi_2,\cdots,\xi_N\}\subseteq R^n$ is the training dataset and $f(w,\xi_i)$ defines the composite loss function at the $i$-th data point. In mini-batch stochastic gradient descent (SGD), we can minimize the objective function $F(w)$ through iteratively updating the model parameter vector $w$ using:
\begin{IEEEeqnarray}{c}\label{eq_secPreli_sgd}
  w_{j+1} = w_j - \frac{\eta_j}{|\mathcal{D}_j|} \sum_{\xi_i \in \mathcal{D}_j} \nabla f(w_j,\xi_i),
\end{IEEEeqnarray}
where $\mathcal{D}_j \subseteq \mathcal{D}$ and $\eta_j$ define the mini-batch data and the learning rate at the $j$-th iteration respectively.

\subsection{Problem Framework}
\begin{figure}[!t]
  \centering
  \includegraphics[width=1\textwidth,trim=15 355 150 10,clip]{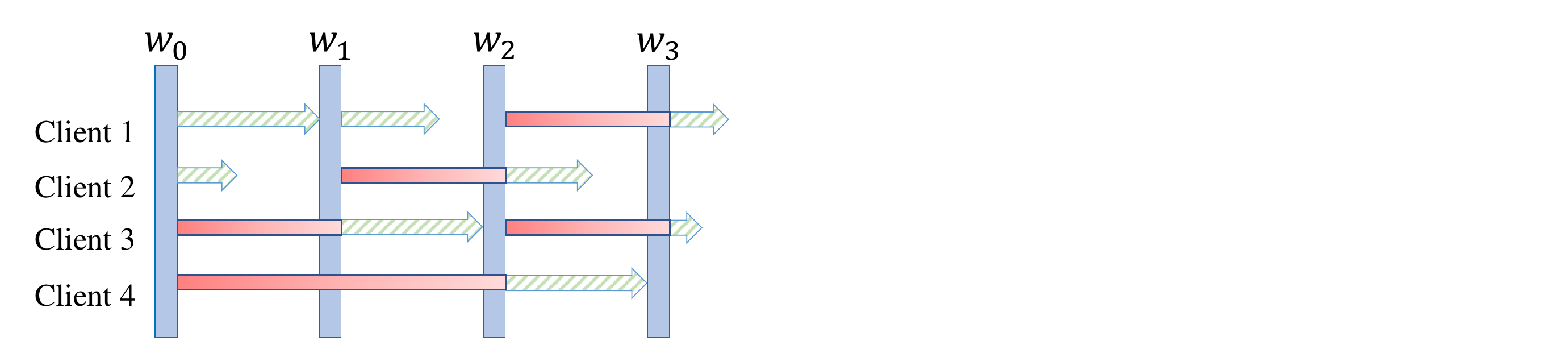}
  \caption{$K$-async FL with $K$=2 and $P$=4. The green shadow arrows indicate the gradients selected for updating. Red arrows indicate the gradients which fail to update the global model in current iteration.}\label{Async}
\end{figure}

We consider the $K$-async FL framework with a central server and $P$ clients. At the beginning, the server initializes the model parameter vector as $w_0$. Then, all the $P$ clients fetch current parameter vector $w_0$ and compute their respective gradients on a single mini-batch data independently. Clients who finish computing gradients upload their gradients without waiting for other clients. The server waits for the first $K$ out of $P$ clients and the rest of clients continue computing gradients. As a result, at every iteration, the gradients received by the server might be computed based on stale parameters.
Then, for the $K$-async FL, the updating formula is expressed by
\begin{IEEEeqnarray}{c}\label{eq_secPreli_Kasyn}
w_{j+1} = w_j - \frac{\eta_j}{K} \sum_{i=1}^{K} g(w_{j,i},\xi_{j,i}),
\end{IEEEeqnarray}
where $g(w_{j,i},\xi_{j,i})$ is the stale gradient received by the server at the $j$-th iteration, $\xi_{j,i}=\{\xi_{j,i}^{(1)},\cdots,\xi_{j,i}^{(m)}\}$ denotes the used $m$ samples of client $i$ at the $j$-th iteration and $w_{j,i}$ defines the model parameter vector used to compute the stale gradient $g(w_{j,i},\xi_{j,i})$ with staleness $\tau_{j,i}$.
\figurename \ref{Async} illustrates the two-async FL when the total number of clients is four. At each iteration, four clients independently upload gradients to the server after finishing computing gradients. The server will update the global model once receiving any two gradients (green and shadow arrows), and clients who fail to upload gradients in limited time continue computing gradients based on a local mini-batch (red arrows). After updating the global model, the server broadcasts the new model parameters to the clients who upload their gradients.

Considering the different levels of clients' staleness and the impact of non-IID data, it is unreasonable to use the same weight $1/K$ to aggregate the received gradients in Equation \eqref{eq_secPreli_Kasyn}. Therefore, in this paper, we will use the weighted $K$-async FL which is expressed as
\begin{IEEEeqnarray}{c}\label{aggregation}
w_{j+1} = w_j - \eta_j \sum_{i=1}^{K} p_{j,i} g(w_{j,i},\xi_{j,i}),
\end{IEEEeqnarray}
where $p_{j,i}$ defines the weight of the $i$-th client's gradient at the $j$-th iteration and $\sum_i^K p_{j,i} = 1$. If $p_{j,i} = 1/K$, Equation \eqref{aggregation} is back to Equation \eqref{eq_secPreli_Kasyn}. Our main purpose is showing how to determine $p_{j,i}$ based on the staleness and non-IID data to improve both the model utility.

\subsection{Definitions and Assumptions}
To determine $p_{j,i}$, we firstly define what kind of stale gradients $g(w_{j,i},\xi_{j,i}), i=1,\cdots,K$ are "consistent" or "inconsistent", based on the cosine similarity comparison between $g(w_{j,i},\xi_{j,i})$ and the globally unbiased gradient $\nabla F(w_j)$.

\begin{definition}[Inconsistent gradient]
\textit{If the cosine similarity of a gradient and the globally unbiased gradient is smaller than a given constant $sim_{min}\in [0,1]$, then the gradient is an inconsistent gradient.}
\end{definition}

\begin{definition}[Consistent gradient]
\textit{If the cosine similarity of a gradient and the globally unbiased gradient is larger than a given constant $sim_{min}\in [0,1]$, then the gradient is a consistent gradient.}
\end{definition}

Besides, some basic assumptions for convergence analysis are listed as follows.
\begin{assumption}[Lipschitz Continuity]\label{assumption1}
  \textit{Objective function $F(w)$ satisfies $L$-Lipschitz continuity:}
  $\forall w_1,w_2$, $\exists$ \textit{constant} $L$,
  \begin{IEEEeqnarray}{c}
    F(w_1) - F(w_2) \leq \nabla F(w_1)^T (w_2-w_1) + \frac{L}{2} ||w_2-w_1||_2^2.\notag
  \end{IEEEeqnarray}
\end{assumption}

\begin{assumption}[Client-Level Unbiased Gradient]\label{assumption2}
\textit{The gradient $g(w_j,\xi_{j,i})$ of client $i$ is a client-level unbiased gradient which means that the expectation of gradient $g(w_j,\xi_{j,i})$ is equal to $\nabla F_i(w_j)$:
\begin{IEEEeqnarray}{c}
  E[g(w_j,\xi_{j,i})]=\nabla F_i(w_j). \notag
\end{IEEEeqnarray}
}
\end{assumption}

\begin{assumption}[Gradients with Bounded Variance]	\label{assumption3}
\textit{The gradient $g(w_j,\xi_{j,i})$ of client $i$ has client-level bounded variance: $ \exists$ constants $\sigma_c,M_c, $
  \begin{IEEEeqnarray}{c}
    E[||g(w_j,\xi_{j,i})-\nabla F_i(w_{j})||_2^2] \leq \frac{\sigma_c^2}{m} + \frac{M_c}{m}||\nabla F_i(w_j)||_2^2.\notag
  \end{IEEEeqnarray}
where $\nabla F_i (w_j)$ is the unbiased gradient of client $i$. To guarantee the convergence of the model, we also need to assume $\nabla F_i(w_j)$ satisfies global-level bounded variance: $\exists$ constant $G $,
  \begin{IEEEeqnarray}{c}
    ||\nabla F(w_j) - \nabla F_i(w_j)||_2^2 \leq G^2, \notag
  \end{IEEEeqnarray}
which means that the difference between the unbiased gradient of each client and the unbiased gradient of global data is limited.}
\end{assumption}

\noindent \textbf{Notations.}
In Table \ref{notations}, we list some notations involved in the paper.
\begin{table}[!t]
  \renewcommand{\arraystretch}{1.3}
  \caption{Notations}\label{notations}
  \centering
   \begin{tabular}{cc}
     \hline
     Variables & Meaning \\
     \hline
     $\eta_j$ & Learning rate at the $j$-th iteration \\
     $P$      & Number of total clients  \\
     $K$      & Number of gradients needed at each iteration \\
     $m$      & Mini-batch size \\
     $L$      & Lipschitz constant  \\
     $J$      & Total number of iteration \\
     $w^*$    & Global optimal solution \\
     $w_j$    & Global model at the $j$-th iteration \\
     $\tau_{max}$ & The maximal staleness of all the gradients in stage one \\
     $\tau_{j,i}$ & The staleness of the $i$-th gradients at $j$-th iteration  \\
     $p_{j,i}$ & Weight of the $i$-th gradient at $j$-th iteration \\
     \hline
   \end{tabular}
\end{table}

\section{Weighted $K$-async FL with Learning Rate Adaptation}
\label{WKAFL}

In this section, we introduce WKAFL to alleviate the contradiction between the existing strategies for staleness and non-IIDness, thus improving the model utility.
We first motivate our work by showing the experimental results about this contradiction in Section \ref{motivation}.
In Section \ref{BasicIdea}, we briefly introduce the basic idea. Section \ref{ModelAggregation} then presents the algorithm details of WKAFL.

\subsection{Motivation}\label{motivation}

To demonstrate the aforementioned contradiction, we implemented the $K$-async FL algorithm that simply integrates the mitigating strategies for staleness and non-IIDness, and trained it on EMNIST MNIST \cite{cohen2017emnist}.
Specifically, the mitigation strategy for non-IIDness here is to accumulate historical gradients \cite{li2019gradient}, while that for staleness is to assign exponential weights to the gradients in aggregation \cite{chen2019communication}.
We use the average staleness and average number of aggregated gradients to represent the impact of staleness and non-IIDness, respectively. The detailed descriptions are as follows.
\begin{itemize}
  \item Average staleness: the mean value of weighted staleness at each iteration and is defined as,
  \begin{IEEEeqnarray}{c}
    \tau_{ave} = \frac{1}{J} \sum_{j=1}^{J} \sum_{i=1}^{K} p_{j,i} \tau_{j,i},   \notag
  \end{IEEEeqnarray}
  where $p_{j,i}$ and $\tau_{j,i}$ denote the weight and staleness of the gradient. Large $\tau_{ave}$ reflects the poor mitigation effect for staleness.
  \item Average number of aggregated gradients: the number of gradients whose weights are larger than a threshold and is defined as,
    \begin{IEEEeqnarray}{c}
      N_{ave} = \frac{1}{J} \sum_{j=1}^{J} {\Big|} \left \{ p_{j,i}, i \in [K] | p_{j,i} \geq p_{j,max} / \mu  \right \} {\Big|}, \notag
    \end{IEEEeqnarray}
    where $p_{j,max} = \max \{ p_{j,i}, i \in [K] \}$ and $\mu \in \mathbb{R}^{+}$ is a constant. Similarly, large $N_{ave}$ reflects the poor mitigation effect for non-IIDness.
\end{itemize}

\begin{figure}[!t]
  \centering
  \subfloat[K=20.]{
		\includegraphics[width=1.5in]{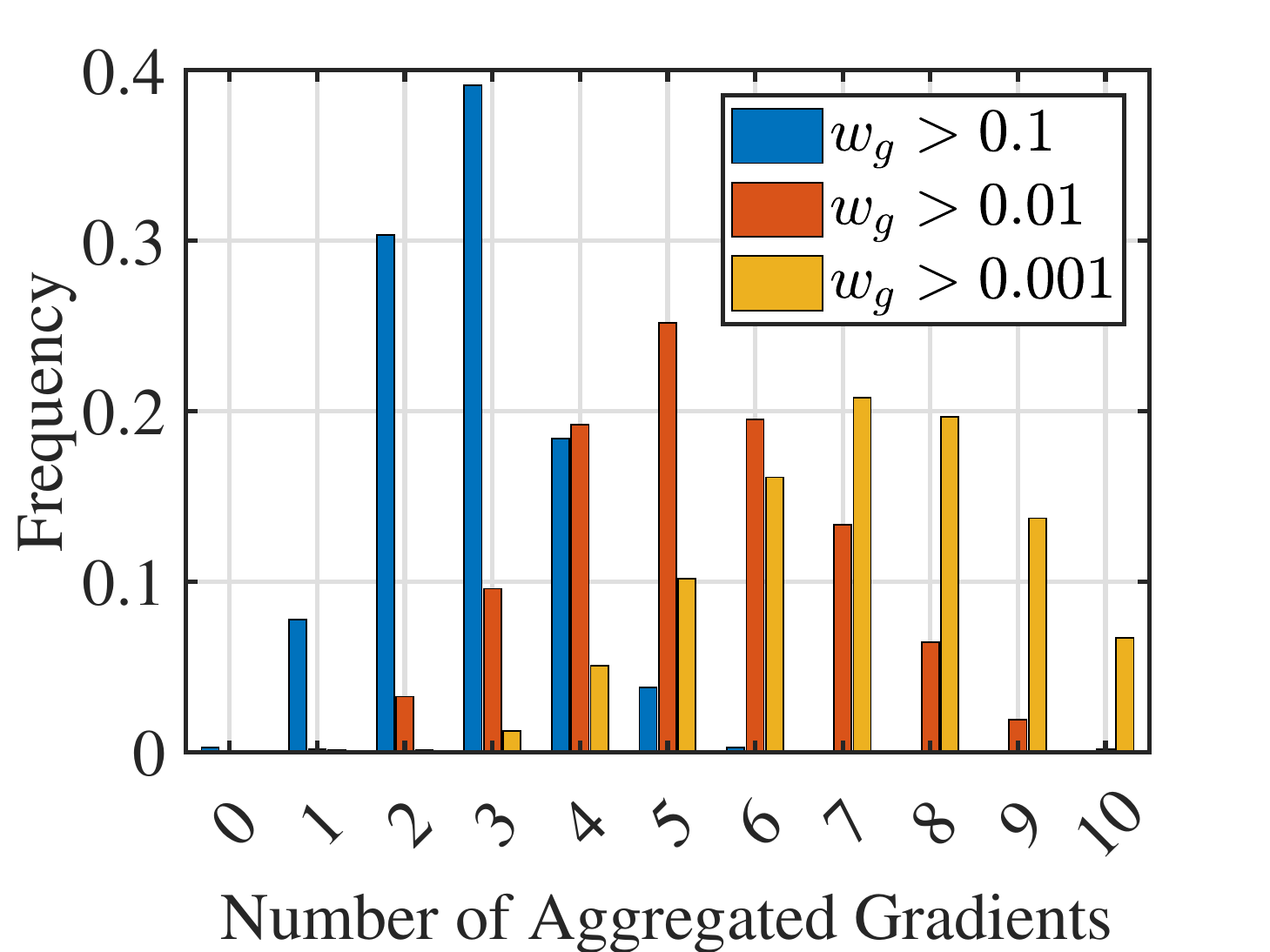}
		\label{moti:Fig:a}
	}
  \hfil
  \subfloat[K=50.]{
		\includegraphics[width=1.5in]{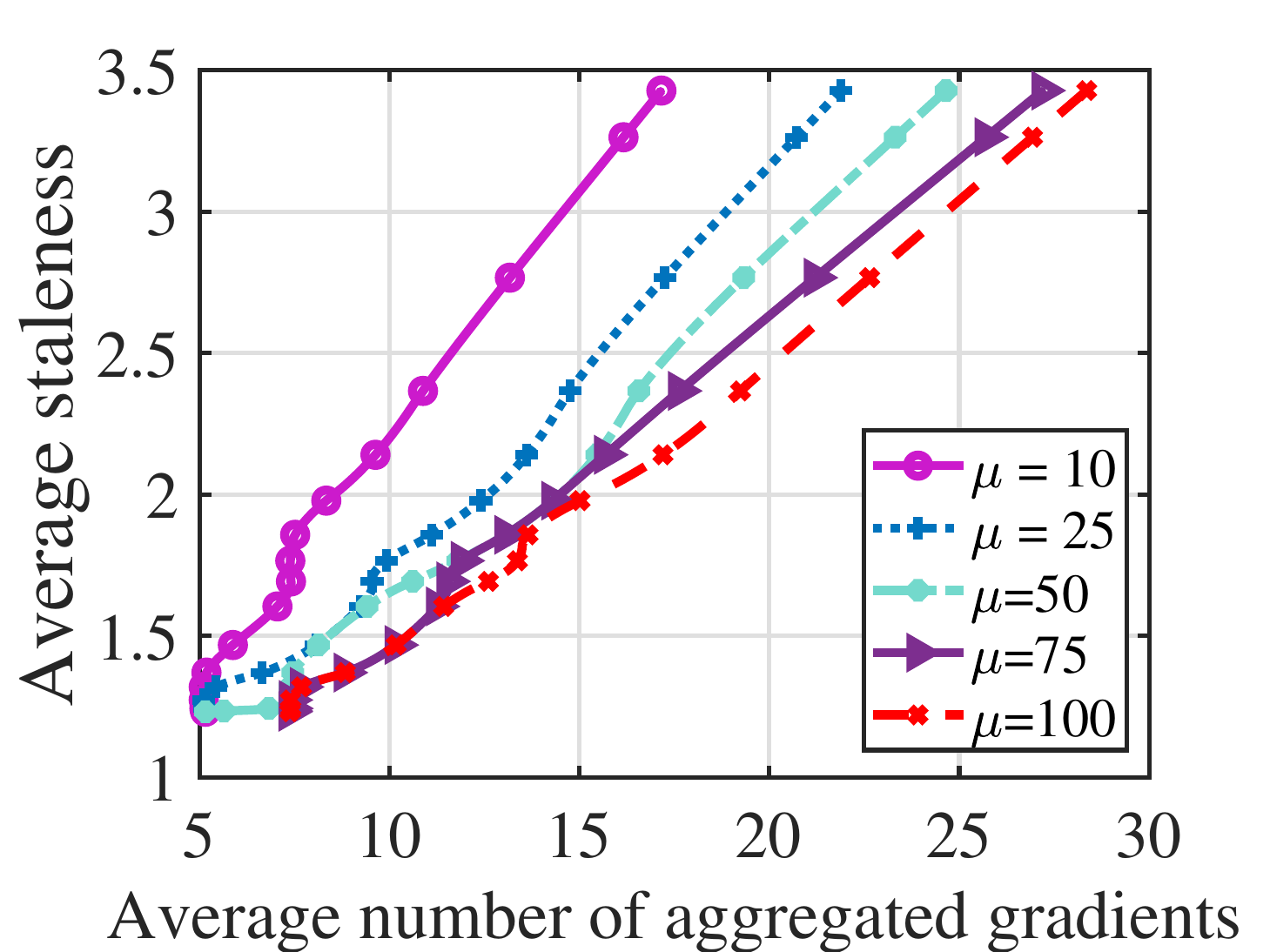}
		\label{moti:Fig:b}
	}
  \caption{Motivation experiments. (a) Exploration of the effect of non-IIDness; (b) Relation of non-IID data (number of aggregated gradients) and staleness (average staleness).}\label{moti-comp}
\end{figure}

Fig. \ref{moti-comp}(a) depicts the histogram over the number of predominated gradients in all aggregations during training. The predominated gradients refer to the gradients whose weights are larger than a threshold $w_g \in \mathbb{R}^{+}$. The more predominated gradients means the aggregated gradients are more representative, which enhances the mitigation for non-IIDness.
However, in almost all iterations, there are at most five gradients with weights larger than 0.1. This reflects that only a small number of gradients dominate the aggregation, incurring an adverse effect on non-IIDness mitigation.
Fig. \ref{moti-comp}(b) shows that the average staleness grows approximately linearly with the number of aggregated gradients, indicating a negative correlation between the two. In conclusion, there exists a contradiction between the mitigation strategies for non-IIDness and staleness.

\subsection{Basic Idea of WKAFL}
\label{BasicIdea}
To break the above contradiction, we propose WKAFL to pick consistent gradients to alleviate the impact of non-IIDness while accelerating the training process.

Generally, gradients with low-stale are relatively reliable and high-stale gradients are more likely to be biased. However, some high-stale gradients may also have consistent descent direction. Then, we can estimate the unbiased gradients based on low-stale gradients and then pick the consistent high-stale gradients. If these gradients are selected, more gradients are aggregated to improve the mitigation effect of non-IID data while guaranteeing the stale gradients not to diverge the model. In such a case, the contradictory of mitigating the effect of non-IID data and staleness will be mitigated.
Apart from this advantage, the selected gradients can also help converge faster. Since stale model is possibly farther from optimal solution and its derivative has a larger norm, stale gradients generally have larger norm for convex loss function. Therefore, the selected consistent stale gradients can help converge faster \cite{mitliagkas2016asynchrony}, \cite{li2019asynchronous}.
%However, when all of the gradients have large staleness, the biased estimated gradients will decrease model utility and destabilize the training process. One feasible method is to adjust the learning rate inversely proportional to the staleness \cite{mcmahan2014delay}, \cite{even2003learning}.

%Based on these observations, WKAFL aims to pick consistent gradients to address the contradictory and utilizes consistent gradients to accelerate the training process at the beginning and then clip gradients whose norm is larger than a given threshold to stabilize training process at the remainder process.

\subsection{Detailed Descriptions}
\label{ModelAggregation}
\begin{figure}
  \centering
  \includegraphics[scale=0.26]{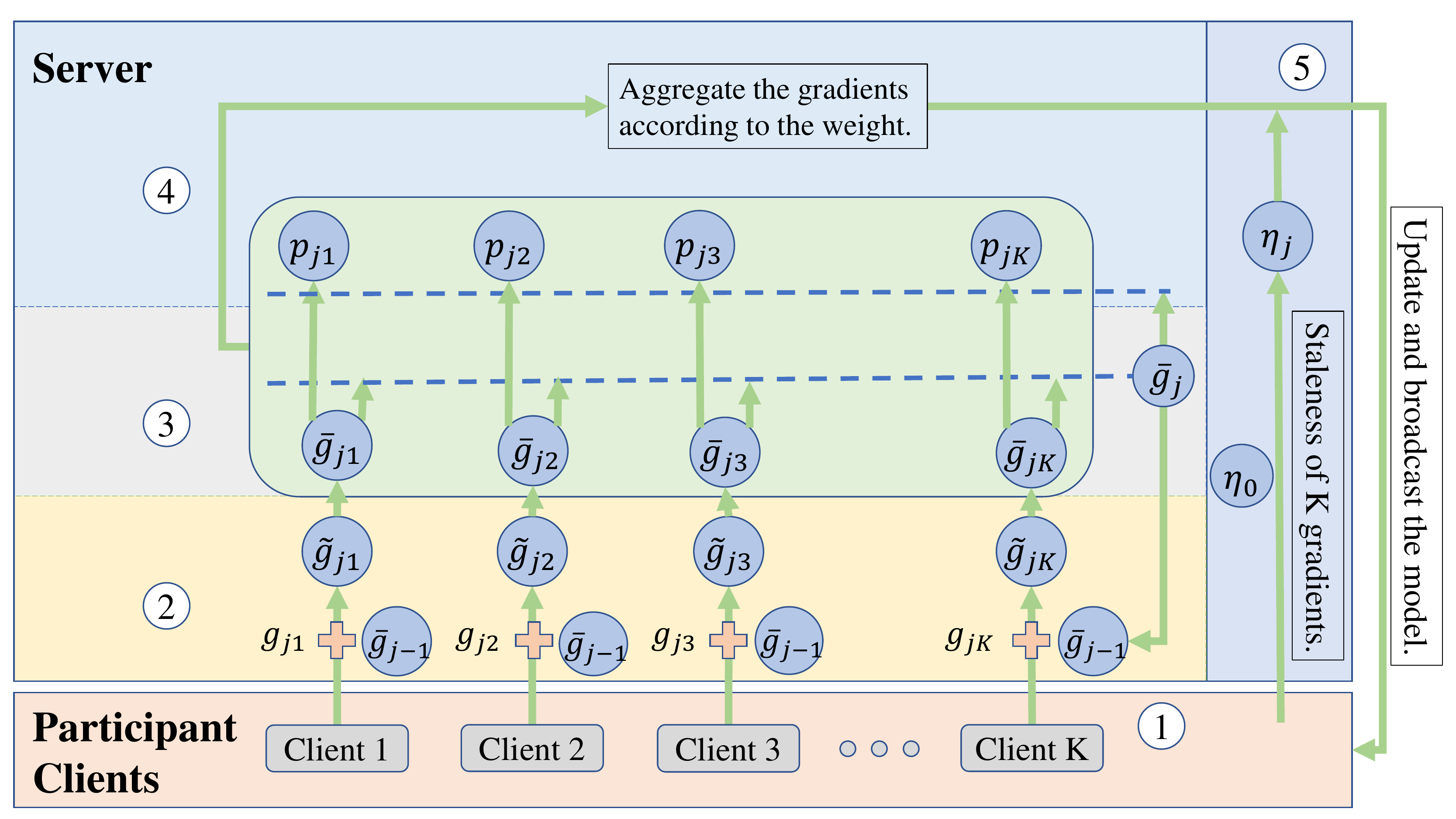}\\
  \caption{WKAFL with five main components, one for clients ({\normalsize \ding{172}}) and four for the server ({\normalsize \ding{173}} - {\normalsize \ding{176}}). Gradients computation ({\normalsize \ding{172}}). Improvement of stability on non-IID data ({\normalsize \ding{173}}). Estimation of globally unbiased gradient ({\normalsize \ding{174}}). Selection and aggregation of consistent gradients ({\normalsize \ding{175}}). Learning rate adaptation ({\normalsize \ding{176}}).}
  \label{workflow}
\end{figure}

Based on above basic idea, we proposed WKAFL to accelerate the training process and improve the training stability.
Pseudo-code of WKAFL consists of two parts, Algorithm \ref{Clients} for clients and Algorithm \ref{Server} for the server. The workflow of the whole process for WKAFL is shown in \figurename \ref{workflow}, which includes gradients computation ({\large \ding{172}}) for clients and four main parts for the server, improvement of stability on non-IID data ({\large \ding{173}}), estimation of globally unbiased gradient ({\large \ding{174}}), selection and aggregation of consistent gradients ({\large \ding{175}}) and learning rate adaptation ({\large \ding{176}}). Detailed description of the four parts can refer to Section \ref{decrease} - Section \ref{adjust} respectively.

For clients ({\large \ding{172}}), as shown in Algorithm \ref{Clients}, they first download the current model parameters and the iteration index from the server (Line 1). Then they compute gradients and upload the results to the server (Lines 2-4).

For the server ({\large \ding{173}} - {\large \ding{176}}), as shown in Algorithm \ref{Server}, after initializing the model (Lines 1-3), the server broadcasts it to all clients for local computation (Line 4). The learning rate adaptation strategy and weighted strategy are presented in Lines 6-16. Particularly, the weighted strategy includes three components (Lines 8, 10-15), alleviating the impact of non-IID data (Line 8), estimating globally unbiased gradient (Lines 13-14), selecting and aggregating consistent gradients (Lines 15) while the learning rate adaptation strategy consists of one component, adjustment of learning rate (Line 16). We will describe the four components of  Algorithm \ref{Server} and explain corresponding rationality in Sections \ref{decrease} to \ref{adjust}.

\begin{algorithm}[!t]
  \caption{WKAFL: Clients side.}\label{Clients}
  \begin{algorithmic}[1]
    \Require mini-batch size $m$.
	\Ensure loss $l_{j,i}$, gradient $g(w_{j,i},\xi_{j,i})$ , delay $\tau(j)$.
    \State Receive model parameters $w_j$ and iteration $j$ from the server.
    \State Compute loss $l_{j,i}$ and gradient $g(w_{j,i},\xi_{j,i})$ based on $m$ samples $\xi_{j,i}$.
	\State Set $\tau(j) \gets j$.
	\State Upload $l_{j,i}$, $g(w_{j,i},\xi_{j,i})$ and $\tau(j)$.
  \end{algorithmic}
\end{algorithm}

\begin{algorithm}[!t]
	\caption{WKAFL: Server side.}\label{Server}
	\begin{algorithmic}[1]
        \Require learning rate $\eta_0$, number of gradients received by server $K$, gradients adjustment parameter $B$, learning rate adjustment parameter $\gamma$, clip-bound parameter $CB$, weighted parameter $\beta$, constant $\alpha$.
		\Ensure Optimal solution $w^*$.
		\State Initialize model parameter $w_0$ and iteration $j=1$.
		\State Initialize the estimated gradient $\bar{g}(w_0)=0$.
		\State Initialize the stage state $stage=1$;
        \State Broadcast $w_0$ and $j$ to all clients.
		\While {the stopping criteria is not satisfied}
		\For {$i=1 \to K$}
        \State Receive loss value, the gradient and its staleness $(l_{i}, g(w_{j,i},\xi_{j,i}),\tau_i)$ from the $i$-th client.
        \State $\tilde{g}(w_{j,i},\xi_{j,i}) = g(w_{j,i},\xi_{j,i}) + \alpha \bar{g}(w_{j-1})$;
		\EndFor
		\If{$\sum_{i=1}^K l_{i} \leq \epsilon$}
		  \State $stage=2;$
		\EndIf
		\State Clip all the gradients with bound $CB$.
        \State Estimate globally unbiased gradient $\bar{g}(w_{j})$ according to Equation \eqref{firstAggregation}.
		\State $g(w_j) = SAGrad(B, \beta, \epsilon, \bar{g}(w_{j,i}, \xi_{j,i}), \bar{g}(w_{j}), stage)$;
        \State Adjust the learning rate $\eta_j$ adaptively according to Equation \eqref{LR}.
		\State $w_{j+1} \gets w_j-\eta_j g(w_j).$
		\State $j\gets j+1$.
		\EndWhile
	\end{algorithmic}
\end{algorithm}

\subsubsection{Improvement of Stability on Non-IID Data}\label{decrease}
It is efficient to decrease the effect of non-IID data by accumulating all the historical gradients. Then, the estimated gradient at $(j-1)$-th iteration can be added to each of $K$  gradients at $j$-th iteration to decrease the impact of non-IID data:
\begin{IEEEeqnarray}{c}\label{Non-IID}
	\tilde{g}(w_{j,i},\xi_{j,i}) = g(w_{j,i},\xi_{j,i}) + \alpha \bar{g}(w_{j-1}),
\end{IEEEeqnarray}
where $\alpha > 0$ is a constant and $\bar{g}(w_{j-1})$ is the estimated gradient at $(j-1)$-th iteration.

Now, we explain the rationality from theoretical perspective.
Every gradient received by the server is biased because of staleness and non-IID data. Then, we can decompose the gradient $g(w_{j,i},\xi_{j,i})$ into three parts.
\begin{IEEEeqnarray}{c}
	E[g(w_{j,i},\xi_{j,i})] = \nabla F(w_j) + E[g_{s}(w_{j,i},\xi_{j,i})] + E[g_{n}(w_{j,i},\xi_{j,i})], \notag
\end{IEEEeqnarray}
where $\nabla F(w_j)$ is the unbiased gradient based on global data at $j$-th iteration, $E[g_{s}(w_{j,i},\xi_{j,i})]$ is the error resulted from staleness, and $E[g_{n}(w_{j,i},\xi_{j,i})]$ is the error resulted from non-IID data.

Similarly, the estimated gradient can also be decomposed into three parts,
\begin{IEEEeqnarray}{c}
  E[\bar{g}(w_{j})] = \nabla F(w_j) + E[\bar{g}_{s}(w_{j,i},\xi_{j,i})] + E[\bar{g}_{n}(w_{j,i},\xi_{j,i})].\notag
\end{IEEEeqnarray}
Then, the expectation of $\tilde{g}(w_{j,i},\xi_{j,i})$ is:
\begin{IEEEeqnarray}{cl}
	E & [\tilde{g}(w_{j,i},\xi_{j,i})] = \nabla F(w_j)+ \alpha \nabla F(w_{j-1})  \notag \\
	& + \underbrace{ E[g_{s}(w_{j,i},\xi_{j,i})] + \alpha E[\bar{g}_{s}(w_{j-1,i},\xi_{j-1,i})]}_{E_S}  \notag \\
	& + \underbrace{ E[g_{n}(w_{j,i},\xi_{j,i})]+ \alpha E[\bar{g}_{n}(w_{j-1,i},\xi_{j-1,i})]}_{E_N}  \notag
\end{IEEEeqnarray}
where $E_N$ is resulted from non-IID data. $E[g_{n}(w_{j,i},\xi_{j,i})]$ only contains the information of client $i$. While $E_{N}$ contains the information of more clients and the information can accumulate as the iterative number increases. Therefore, Equation \eqref{Non-IID} can decrease the effect of non-IID data and stabilize the model.

\subsubsection{Estimation of Globally Unbiased Gradient}\label{estimate}
\textit{It is assumed that gradients with lower staleness have high probability to be consistent and gradients with higher staleness have high probability to be inconsistent} because gradients based on stale model generally deviate from consistent direction. Therefore, we can aggregate gradients by weighting them according to their staleness to estimate globally unbiased gradient. The rule of aggregation is
\begin{IEEEeqnarray}{c}\label{firstAggregation}
	\bar{g}(w_j) = \sum_{i=1}^{K} \frac{a_{j,i}}{a} \bar{g}(w_{j,i},\xi_{j,i}),
\end{IEEEeqnarray}
where $\bar{g}(w_{j,i},\xi_{j,i})$ is a clipped gradient of $\tilde{g}(w_{j,i},\xi_{j,i})$, $a= \sum_{i=1}^K a_{j,i}$ and $a_{j,i}$ is a function of staleness. Cong et al. \cite{xie2019asynchronous} empirically figured out that a better utility was achieved when $a_{j,i}$ was chosen as an exponential function. Chen et al. \cite{chen2019communication} found that $a_{j,i}$ was more properly set as $(\frac{e}{2})^{-\tau_{j,i}}$ than $e^{-\tau_{j,i}}$, where $\tau_{j,i}$ is the staleness of gradient $g(w_{j,i}, \xi_{j,i})$ and $e$ is natural logarithm. Based on their work, we set $a_{j,i}$ to $(\frac{e}{2})^{-\tau_{j,i}}$ in our experiments.

\begin{figure}[!ht]
  \centering
  \includegraphics[width=0.35\textwidth, trim={0 25 0 0}, clip]{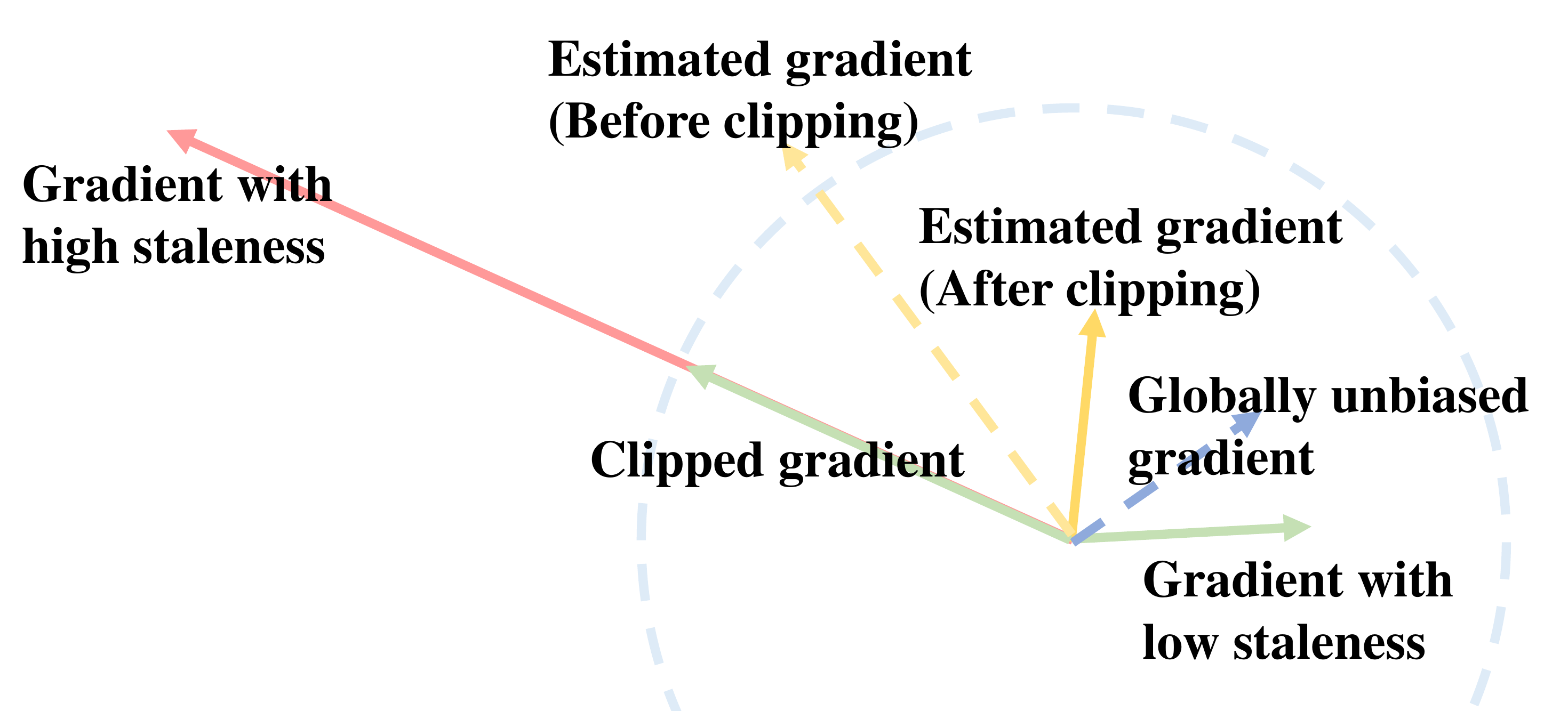}  \\
  \caption{Estimated gradient before and after clipping the gradients respectively.}\label{Clip}
\end{figure}

Before estimating the globally unbiased gradient, it is necessary to clip the gradients for two reasons. On the one hand, it is known that clipping can prevent gradient explosion. On the other hand, under the assumption that gradients with lower (higher) staleness have high probability to be consistent (inconsistent), the aggregated gradient after clipping will be closer to the globally unbiased gradient. A straightforward illustration is shown in \figurename \ref{Clip}. When aggregating the gradients before clipping, the norm of stale gradients may be so large that the aggregated gradient deviates from the globally unbiased gradient. When aggregating the gradients after clipping, the aggregated gradient is closer to the globally unbiased gradient due to the alleviation of impact of gradients with higher staleness.

\subsubsection{Selection and Aggregation of Consistent Gradients}\label{select}
\begin{algorithm}[!t]
	\caption{Select and aggregate consistent gradients.}\label{selectGradients}
	\begin{algorithmic}[1]
		\Require $B$, $\beta$, $\epsilon$, $\bar{g}(w_{j,i},\xi_{j,i})$, $\bar{g}(w_{j}), stage$.
		\Ensure $g(w_{j})$.
		\Function{SAGrad}{$B$, $\beta$, $\epsilon$, $\bar{g}(w_{j,i},\xi_{j,i})$, $\bar{g}(w_{j})$, $stage$}
		\For {$i=1,\cdots,K$}
		\State $sim_{j,i} = \cos \left \langle \bar{g}(w_{j,i},\xi_{j,i}), \bar{g}(w_{j})\right \rangle;$
		\If {$sim_{j,i}\geq sim_{min}$}
		\State $p_{j,i}' = \exp(\beta*sim_{j,i})$;
		\Else
		\State $p_{j,i}' = 0$;
		\EndIf
		\EndFor
		\If {$stage=2$}
		\If {$B||\bar{g}(w_{j})||_2 \leq ||\bar{g}(w_{j,i},\xi_{j,i})||_2$}
		\State $\bar{g}(w_{j,i},\xi_{j,i}) = \frac{B||\bar{g}(w_{j})||_2}{||\bar{g}(w_{j,i},\xi_{j,i})||_2} \bar{g}(w_{j,i},\xi_{j,i})$;
		\EndIf
		\EndIf
		\State $p_{j,i} = p_{j,i}'/sum_{i=1}^K p_{j,i}'$;
		\State $g(w_j) = \sum_{i=1}^K p_{j,i} \bar{g}(w_{j,i},\xi_{j,i})$.
		\State \Return{$g(w_j)$}
		\EndFunction
	\end{algorithmic}
\end{algorithm}

Some stale gradients may have consistent update direction. Then, we can pick these gradients to accelerate the training process. When the staleness is low, the estimated gradient is generally consistent with globally unbiased gradient. Therefore, when judge whether a gradient is consistent or inconsistent, we treat the estimated gradient as globally unbiased gradient for convenience. To determine which gradients are consistent and the corresponding degrees, the server will first compute the cosine similarity $sim_{j,i}$ of the estimated globally unbiased gradient $\bar{g}(w_j)$ and corresponding gradient $\bar{g}(w_{j,i},\xi_{j,i})$. Then the server assigns each gradient with weight according to the cosine similarity. The rule for assigning weight is given by:
\begin{align}\label{sim}
	sim_{j,i} = \cos \left \langle \bar{g}(w_{j,i},\xi_{j,i}), \bar{g}(w_{j})\right \rangle;
\end{align}
\begin{equation} \label{weight}
 p_{j,i}' =
 \begin{cases}
 \exp(\beta sim_{j,i}), & \text{if} \quad sim_{j,i}\geq sim_{min} \,;\\
 0,  &  \text{if} \quad sim_{j,i}<sim_{min}. \\
\end{cases}
\end{equation}
\begin{align}
	p_{j,i} = p_{j,i}'/sum_{i=1}^K p_{j,i}';
\end{align}
where $\beta$ is a constant, $p_{j,i}$ defines the weight of gradient $g(w_{j,i},\xi_{j,i})$, and $sim_{min}$ is a given threshold used to judge whether a gradient is consistent or not. Gradients whose cosine similarity is lower than the threshold will be judged as inconsistent gradients and abandoned. On the contrary, consistent gradients will be aggregated according to their weights. The aggregation rule is as follow:
\begin{IEEEeqnarray}{c}
  g(w_j) = \sum_{i=1}^K p_{j,i} \bar{g}(w_{j,i},\xi_{j,i}). \nonumber
\end{IEEEeqnarray}

However, when the model is going to converge, the norm of stale gradients may be too large to make the model fluctuate around the optimal solution.
To deal with this problem, we divide the whole training process into two stages. In stage one, the training process can be accelerated by taking advantage of stale gradients. When the average loss of $K$ participant clients is smaller than a constant threshold $\epsilon$ , the training process enters into stage two. Though the norm of gradient also reflects the state of global model, it is improper to be selected as the stage division criterion because the global model may enter the local optimization area where the norm of gradient is approximate to zero but loss value is not the minimum. In stage two, to guarantee the training stability, consistent stale gradients will be clipped before aggregating based on the norm of estimated gradients. The clipping rule is given by:
\begin{equation} \label{stageTwoClip}
 \bar{g}(w_{j,i},\xi_{j,i}) =
 \begin{cases}
 &\frac{B||\bar{g}(w_{j})||_2}{||\bar{g}(w_{j,i},\xi_{j,i})||_2}  \bar{g}(w_{j,i},\xi_{j,i}),
 \\
 & \quad \text{if} \quad B||\bar{g}(w_{j})||_2 \leq ||\bar{g}(w_{j,i},\xi_{j,i})||_2 \,; \\
 &\bar{g}(w_{j,i},\xi_{j,i}),
 \\
 & \quad \text{if} \quad B||\bar{g}(w_{j})||_2 > ||\bar{g}(w_{j,i},\xi_{j,i})||_2.\\
\end{cases}
\end{equation}
where the gradients adjustment parameter $B \in (0,\infty)$ is a constant.

The workflow can refer to Algorithm \ref{selectGradients}. The server first computes the cosine similarity between estimated gradient and every gradient received by the server (Line 3). Then the server assigns gradient $g(w_{j,i},\xi_{j,i})$ with a weight $p_{j,i}$ according to cosine similarity (Lines 4-8) and aggregates the gradients (Lines 15-16) eventually. If the model enters stage two, stale gradients will be clipped (Lines 10-14).

\subsubsection{Learning Rate Adaptation}\label{adjust}
In WKAFL, the estimated gradient may deviate seriously from the globally unbiased gradient when most of the $K$ gradients have high staleness, leading to a seriously biased estimated gradient and a poor improvement of model utility. To alleviate it, a staleness-based adaptive learning rate is more reasonable than using a constant learning rate. Intuitively, a small learning rate is required for alleviating the impact of high staleness. In \cite{zhang2015staleness}, Zhang et al. divided the initial learning rate by the staleness to adjust the learning rate. However, it can not be extended to large scale AFL due to the possibly high staleness. In such a case, the adjusted learning rate will be so small that the training process will be prolonged due to the very slight improvement at each iteration. To address it, we adjust the learning rate according to the minimal staleness of the $K$ gradients because the estimated gradient is mainly determined by gradients with low staleness, which is expressed as follows.
\begin{IEEEeqnarray}{c}\label{LR}
\eta_j = \eta_0 * \frac{1}{\tau_{min,j}*\gamma+1},
\end{IEEEeqnarray}
where $\eta_0$ is the initial learning rate, $\tau_{min,j}$ is the minimal staleness of $K$ gradients at $j$-th iteration, and $\gamma \in (0,1)$ is a constant. Note that $\eta_j\leq\eta_{0}$ and the equality holds only if $\tau_{min,j}=0$ (i.e., no staleness).

\section{Analysis} \label{analysis}
In Section \ref{WKAFL}, we have explained the rationality of weighting and adjustment for learning rate to improve the prediction accuracy and stabilize the training process. In this section, we further analyze the convergence rate of the proposed WKAFL in which the loss function is non-convex, considering both non-IID data and unbounded staleness. Theorems \ref{stageone} and \ref{stagetwo} show the convergence analysis results with respect to two stages in WKAFL respectively. \textit{In stage one, the staleness is bounded while in stage two, the staleness is unbounded.}
Most convergence analysis of AFL algorithm is based on the theory of stochastic optimization and one challenge is how to bridge the stale gradient and globally unbiased gradient.
In stage one, the aggregated gradient at $j$-th iteration is $g(w_j) = \sum_{l=1}^{j} \alpha^{j-l} \sum_{i=1}^K p_{l,i} g(w_{l,i},\xi_{l,i})$. Consequently, the difference between global unbiased gradient $\nabla F(w_{j})$ and the aggregated gradient is the sum of the difference between each component gradient $g(w_{l,i}, \xi_{l,i})$ and the unbiased gradient. We connect the local gradient $g(w_{l,i},\xi_{l,i})$ with the theoretical unbiased gradient $\nabla F(w_j)$ by two steps.
\begin{itemize}
  \item Step 1: We connect the local gradient $g(w_{l,i},\xi_{l,i})$ with globally unbiased gradient $\nabla F(w_{l-\tau_{l,i}})$ by taking expectation of $g(w_{l,i},\xi_{l,i})$ and bounding the locally unbiased gradient $\nabla F_i(w_{l-\tau_{l,i}})$ with $\nabla F(w_{l-\tau_{l,i}})$ based on Assumption \ref{assumption3}.
  \item Step 2: We connect the unbiased gradient $\nabla F(w_{l-\tau_{l,i}})$ with the unbiased gradient $\nabla F(w_{j})$. The difference can be bounded by accumulating all aggregated gradients from $\max (l-\tau_{max}, 1)$ to $j-1$ iterations.
\end{itemize}

However, in stage two, the staleness can be infinite in theory due to the possibly infinite number of total iterations. Therefore, the method for dealing with finite staleness will not hold in stage two. To deal with it, we use three steps to connect the local gradient $g(w_{l,i},\xi_{l,i})$ of $l$-th iteration contained in the momentum term with $\nabla F(w_j)$.
\begin{itemize}
  \item Firstly, we connect the local gradient $g(w_{l,i},\xi_{l,i})$ with estimated gradient $\bar{g}(w_j)$ by clipping $g(w_{l,i},\xi_{l,i})$ and selecting consistent gradients. Based on Equation \eqref{stageTwoClip}, we have $||\bar{g}(w_j,\xi_{j,i})||_2^2 \leq B^2||\bar{g}(w_j)||_2^2$. Apart from that, since the accumulated gradients are consistent with the estimated gradients $\bar{g}(w_l)$ and inconsistent gradients will be endowed with weight 0, then the weighted differences between clipped gradients at $l$-th iteration and the estimated gradients can be bounded by $||\bar{g}(w_l)||_2^2$, i.e., $\exists \sigma_e^2,M_e$,
      \begin{IEEEeqnarray}{Cl}	\label{assumpTwo1}
        \sum_{i=1}^{K} p_{l,i}||\bar{g}(w_l) & - \bar{g}(w_{l,i}, \xi_{l,i})||_2^2   \nonumber  \\
        & \leq \sigma_e^2 + M_e||\bar{g}(w_l)||_2^2.
      \end{IEEEeqnarray}
  \item Secondly, after alleviating the effect of non-IID data and staleness, for the reason that the estimated gradient $\bar{g}(w_{l})$ is closed to the unbiased gradients, we can assume
      \begin{IEEEeqnarray}{c}	
        ||\bar{g}(w_l) - \nabla F(w_l)||_2^2 \leq \sigma_c^2, \quad \exists \sigma_c^2, \label{assumpTwo2}
      \end{IEEEeqnarray}
      which bridges the connection between $\bar{g}(w_l)$ and $\nabla F(w_l)$.
  \item Thirdly, the differences between $\nabla F(w_{l})$ and $\nabla F(w_{j})$ can be bounded by the aggregated gradients from $l$ to $j$ iterations.
\end{itemize}

After providing the proof sketch for the two stages, we will present the analysis results of WKAFL. The convergence result of stage one is as follows.

\begin{theorem}[Stage One]  \label{stageone}
  \textit{Assume Assumption \ref{assumption1}, \ref{assumption2}, \ref{assumption3} hold.
  To guarantee the convergence, learning rate $\eta_j$ satisfies that $\eta_j \leq \frac{1}{2L(1+\frac{M_G}{m})}$ and $L^2 \sum_{t=l}^{J}  (J-\tau_t) \alpha^{-t} \sum_{k=l}^{t} \eta_{k}^2 s_{k} \alpha^{k} ( 1 + \frac{M_c}{m} ) < \frac{1}{4}, l \in [J]$.
  Then, we have the following convergence result:}
  \begin{IEEEeqnarray}{lc}\label{conclusionStageone}
    & \frac{1}{J} \sum_{j=1}^J \frac{\eta_j s_j}{2} \mathbb{E}[||\nabla F(w_j)||_2^2] \leq  \frac{1}{J} \sum_{j=1}^J  (A_j \eta_j^2 + \eta_j s_j G^2+ \nonumber  \\
    & \quad \eta_j \sum_{l=1}^{j} \alpha^{j-l} (j-\tau_{l}) \sum_{t=\tau_l}^{j-1} \eta_t^2 B_j) + \frac{F(w_1)-F(w^*)}{J},
  \end{IEEEeqnarray}
  where $s_j = \sum_{k=1}^{j} \alpha^{j-k}$, $\tau_{l} = \max (1, l-\tau_{max})$, $A_j = \frac{L \sigma_c^2}{2m} s_j \sum_{l=1}^{j} \alpha^{j-l} \mathbb{E} [\sum_{i=1}^{K} p_{l,i}^{2}]$ and $B_j = L^2 s_t \sum_{k=1}^{t} \alpha^{t-k}  \frac{\sigma_c^2}{m} \mathbb{E}[\sum_{i=1}^{K} p_{k,i}^{2}]$.
\end{theorem}
\begin{IEEEproof}
See Appendix \ref{proofStageone}.
\end{IEEEproof}

Based on Theorem \ref{stageone}, we can qualitatively analyze the impact of staleness and different levels of non-IID data.
\textit{With respect to staleness}, it is observed that the maximal staleness $\tau_{max}$ has a negative impact on convergence rate. In AFL, the staleness is increasing with the increase of $P/K$, the ratio of number of total clients to the participants. Therefore, for large scale AFL system, it is critical to decrease $P/K$ to ensure a high model utility. Alternatively, we demonstrate how to further improve the model utility under the given $P/K$ by exploiting the stale gradients. \textit{With respect to non-IID data} measured by the parameter $G^2$, it is observed that the convergence rate is decreasing with $G^2$. That is, the non-IID data has a negative impact on the model utility.
However, the convergence result is too complex to acquire the convergence rate. Therefore, we consider the constant learning rate to simplify Equation \eqref{conclusionStageone}.

\begin{remark} \label{remarkOne}
 If we set the weight of each received gradients to $1/K$ and set the learning rate to a constant $\eta$ in Theorem \ref{stageone}, the right part of the Equation \eqref{conclusionStageone} is $\mathcal{O} (\frac{\eta^2}{K} + \eta G^2 + \frac{\eta^3(1+J^2 \alpha^J)}{K} + \frac{1}{J})$. If we further set the learning rate to $\eta = \frac{ \sqrt{K} }{ \sqrt{J} }$, then for any large enough $J$ that satisfies $J \geq 4KL^2(1+\frac{M_G}{m})^2$ and $L^2 K \sum_{t=1}^{J}  (J-\tau_t) \alpha^{-t} \sum_{k=l}^{t}  s_{k} \alpha^{k} ( 1 + \frac{M_c}{m} ) < \frac{J}{4}$, the convergence rate is $\mathcal{O} (\frac{1}{\sqrt{KJ}} + \frac{1}{J} + G^2)$ where $G$ measures the level of non-IID data. We can notice that the the non-IID data will cause the model to fluctuate near the optimal solution. In IID settings, we have $G=0$ and the convergence rate is $\mathcal{O} (\frac{1}{\sqrt{KJ}} + \frac{1}{J})$ which indicates that WKAFL achieves a linear speedup.
\end{remark}

%$\eta^2 4 s_J^3 \alpha^{J} \tau_{max} L^2 (1+M_c/m)\leq 1$
%$\eta^2 4 \tau_{max} L^2 (1+M_c/m)\leq (1-\alpha)^3$

Next, we will present the convergence result for stage two. Since the staleness can be unbounded and the gradients are clipped which is different from stage one, two additional assumptions are required and the rationality has been explained in the proof sketch.

\begin{theorem}[Stage Two]  \label{stagetwo}
  \textit{Assume Assumption \ref{assumption1}, \ref{assumption2}, \ref{assumption3} hold. To guarantee the convergence, we also assume Equation \eqref{assumpTwo1} and \eqref{assumpTwo2} holds. Learning rate satisfies that $\eta_j \leq \frac{1}{Ls_j}$ and subjects to $V_{j}\geq 0$, where $V_{j}$ is
  \begin{IEEEeqnarray}{Cl}
    V_j = \frac{\eta_j s_j}{2} - & \sum_{l=j}^{J} 3M_e \eta_l \alpha^{l-j} - \sum_{l=j}^{J} \eta_l  \sum_{t=j}^{l-1} \eta_t^2  \sum_{k=1}^{t}  \alpha^{t-j} I_{l,k,t}.  \nonumber
  \end{IEEEeqnarray}}
	\textit{Then, we can obtain the following convergence result:}
	\begin{IEEEeqnarray}{lc}\label{conclusionStagetwo}
      & \frac{1}{J} \sum_{j=1}^J V_j ||\nabla F(w_j)||_2^2 \leq \frac{1}{J} \sum_{j=1}^J (\frac{3\eta_j s_j}{2} C + \eta_j   \nonumber  \\
      & \quad \sigma_c^2  \sum_{l=1}^{j} \sum_{t=l}^{j-1}\eta_t^2 s_t I_{j,l,t}) + \frac{F(w_1) - F(w^*)}{J},
	\end{IEEEeqnarray}
  where $C = \sigma_c^2 + 2M_e\sigma_c^2+\sigma_e^2$ and $I_{l,k,t} = 3 L^2 B^2 s_t \alpha^{l-k} (l-k)$.
\end{theorem}
\begin{IEEEproof}
    See Appendix \ref{proofStagetwo}.
\end{IEEEproof}

Parameters $\sigma_c^2$ represents the difference between the estimated gradient and the globally unbiased gradient. When $\sigma_c^2$ is small, the estimated gradient is closer to the globally unbiased gradient. The right part of Equation \eqref{conclusionStagetwo} becomes small and the model will converge faster. Similarly, as $B$ decreasing, the left part of Equation \eqref{conclusionStagetwo} becomes large while the right part becomes small. Then, the model will also converge faster. In summary, both $\sigma_c^2$ and $B$ have negative impacts on the convergence rate.
The same as stage one, we will also analyze the settings with a constant learning rate for stage two to present the convergence rate.

\begin{remark}
 If we set the learning rate to a constant $\eta$ in Theorem \ref{stagetwo}, the right part of Equation \eqref{conclusionStagetwo} is $\mathcal{O} (\eta C + \eta^3 +\frac{1}{J})$. If we further set the learning rate to $\eta = \frac{1}{\sqrt{J}}$, then for any  $J \geq \frac{L^2}{(1-\alpha)^2} \geq L^2 S_J^2$ and large enough $J$ to make $\frac{1}{J} \sum_{l=j}^J \sum_{t=j}^{l-1} \sum_{k=1}^t \alpha^{t-j} I_{l,k,t}$, we have $ \mathcal{O} (\frac{1}{ \sqrt{J} } + C)$. Since the term $\sum_{l=j}^J \sum_{t=j}^{l-1} \sum_{k=1}^t \alpha^{t-j} I_{l,k,t}$ contains $\alpha<1$, then the term is $\mathcal{O} (\alpha^J J^y), y \in \mathbb{R}$. Therefore, there exists large enough $J$ to meet the conditions. The parameter $C$ reflects the level of non-IID data and the convergence rate demonstrates that non-IID data will decrease the prediction accuracy. In IID settings, we have $C=0$ and the convergence rate will be $\mathcal{O} (\frac{1}{\sqrt{J}})$.
\end{remark}

\section{Experiment} \label{experiment}
In this section, we conducted extensive experiments \footnote{Source code is available at https://github.com/zzh816/WKAFL-code.} under the PySyft framework \cite{ryffel2018generic} to validate the performance of proposed algorithm WKAFL in terms of the training speed, prediction accuracy and training stability.
The proposed WKAFL was compared with three existing algorithms, which will be described in Section \ref{sec_Experi_algorithms_datasets}. Meanwhile, several benchmark datasets were adopted with varying levels of non-IIDness.
In Section \ref{sec:ablation}, ablation experiments demonstrate the impacts of each component of WKAFL.
In Section \ref{ExpWKAFL}, detailed comparison results are illustrated to validate the advantage of WKAFL.

\subsection{Experiment Setup}\label{sec_Experi_algorithms_datasets}

\subsubsection{Datasets and Models}\label{sec_Experi_datasets}
In our experiments, we used four datasets: CelebA \cite{caldas2018leaf}, EMNIST ByClass \cite{cohen2017emnist}, EMNIST MNIST \cite{cohen2017emnist} and CIFAR10 \cite{cifar10}. The first two benchmark FL datasets were directly used. EMNIST MNIST and CIFAR10 were manually split to have different levels of non-IIDness.

\begin{itemize}
  \item CelebA partitions the Large-scale CelebFaces Attributes Dataset by the celebrity in the image.
  \item EMNIST ByClass contains 81,4255 character images of 62 unbalanced classes. It is partitioned over around 3,500 clients according to the writers in FL.
  \item EMNIST MNIST has 70,000 characters in ten balanced labels. We partitioned it into 10 subsets by the label. Then, the digit images were divided over clients, which sampled local data according to Algorithm \ref{dataSampler}. Every client has $L_{num}$ classes of labels and the total number of data ranges from $D_{min}$ to $D_{max}$.
  \item CIFAR10 is a labeled dataset of 60,000 tiny color images in 10 classes, and consists of 50,000 training images and 10,000 test images. We partitioned CIFAR10 into FL according to Algorithm \ref{dataSampler} similarly. % similar to EMNIST MNIST.
\end{itemize}

\begin{algorithm}[!htbp]
    \caption{Generator of non-iid datasets.}\label{dataSampler}
    \begin{algorithmic}[1]
        \Require $D_{min}$, $D_{max}$, $L_{num}$, $P$.
        \Ensure $Datas=(D_1, D_2, \cdots, D_P)$.
        \For {$i$ $\in$ $\{1,2,\cdots,P\}$}
            \State $Classes$ $\gets$ Sample $L_{num}$ labels from $\{0,1,2,\cdots,9\}$;
            \State $D_{num}$ $\gets$ RandInt($D_{min}$,$D_{max}$);
            \State $weights$ $\gets$ Sample $L_{num}$ number in interval (0,1);
            \State $Data_{num}$ $\gets$ $D_{num}*weights/sum(weights)$;
            \State $D_{i}$ $\gets$ Sample data according to $Data_{num}$ and $Classes$.
        \EndFor
    \end{algorithmic}
\end{algorithm}

For the above four datasets, we deployed different CNN models.
\textit{For CelebA}, the model contains one convolutional layer with ReLU activation function followed by a maxpooling layer. The final pooled vector is passed to a dense layer with 16000 units. The loss was measured by the logarithmic softmax function.
\textit{For EMNIST ByClass}, it contains two convolutional layers with ReLU activation function followed by maxpooling layers. The final pooled vector is passed to a dense layer with 512 units. The loss was measured by the logarithmic softmax function.
\textit{For EMNIST MNIST}, the same CNN model for EMNIST ByClass's is deployed.
\textit{For CIFAR10}, we adopted LeNet \cite{lecun1998gradient}, a simple and classical CNN model.

\subsubsection{Comparison Algorithms}
We compared three asynchronous algorithms in our experiments, including temporally weighted asynchronous federated learning (TWAFL \cite{chen2019communication}), staleness-aware async-SGD (SASGD, \cite{zhang2015staleness}), and gradient scheduling with global momentum (GSGM, \cite{li2019gradient}). TWAFL and SASGD, were proposed to alleviate the effect of staleness to improve the prediction accuracy, and GSGM was proposed to alleviate the effect of non-IID data to improve training stability. %However, the proposed WKAFL aims to improve both the prediction accuracy and training stability considering both staleness and non-IID data.

\emph{TWAFL.}
Unlike aggregating stale gradients in WKAFL, the server in TWAFL directly aggregates the $K$ stale model parameters by assigning the stale model parameters with staleness-based weights to improve the prediction accuracy. For the sake of fair comparison, we modify TWAFL in which the server aggregates $K$ stale gradients instead of model parameters. In particular, the model update formula is expressed as follow.
\begin{IEEEeqnarray}{c}
  w_{j+1} = w_{j} - \eta_{j} \sum_{i=1}^K \frac{m_i}{m} \left(\frac{e}{2}\right)^{-\tau_{j,i}}*g(w_{j,i},\xi_{j,i}), \nonumber
\end{IEEEeqnarray}
where $m_i,\tau_{j,i}, g(w_{j,i},\xi_{j,i})$ are the mini-batch size, staleness, and gradients uploaded by client $i$ at the $j$-th iteration with $m=\sum_{i=1}^K m_i$.

\emph{SASGD.}
In SASGD, the sever first aggregates the $K$ received stale gradients $g(w_{j,i},\xi_{j,i})$ which are assigned with staleness-based weights $\eta_{j,i}$. Then, the server applies the aggregated gradient to update the model parameters. The specific updating formula is presented as follows.
\begin{IEEEeqnarray}{rCl}
  \eta_{j,i} & = & \eta_0 / \tau_{j,i};  \nonumber \\
w_{j+1} & = & w_{j} - \frac{1}{K} \sum_{i=1}^K \eta_{j,i} g(w_{j,i},\xi_{j,i}),  \nonumber
\end{IEEEeqnarray}
where $\eta_{0}$ is the initial learning rate and $\tau_{j,i}$ is the staleness for $i$-th client at $j$-th iteration.

\emph{GSGM.}
In GSGM with $P$ clients totally, the server only uses one gradient at each iteration (fully AFL) and the whole training process is divided into several rounds. In each round, the server averages part gradients used in gradient scheduling strategy to modify the current inconsistent direction. According to this strategy, once the gradients submitted by a fast client are used for updating in one round, the client is blocked until next round begins.
For the sake of fair comparison, we modify GSGM in which $K$ gradients are selected to update the model at each iteration with other protocols unchanged and GSGM is also modified in the way that each client uploads only one gradient in one round to improve the accuracy of GSGM.

\subsubsection{Metrics}\label{sec_Experi_algorithms}

In our experiments, we compared WKAFL with above algorithms in terms of training speed, model accuracy and training stability.
The training speed is directly measured by the number of iterations before convergence.
The model accuracy is measured by the prediction accuracy on test set. We describe the metric of training stability as follows.

Most studies focus more on the convergence rate and less on whether the training is stable. Particularly, a stable training indicates the training can be terminated with a relative stable model accuracy, instead of a fluctuated accuracy curve.
In this paper, we use training stability to refer to the stability of last few model prediction accuracy on the test set before the final convergence. Particularly, it is measured by the standard deviation of the last $A_{num}$ (which was set as 10) model accuracy values under logarithmic coordinates.

%For most algorithms, we focus more on convergence rate and ignore the training stability. However, a stable training process can produce an accurate judgment to decide whether the training is to be terminated, while a fluctuant training process confuses us whether the current model converges. Therefore, it is necessary to consider training stability. Generally, we focus more attention on the stability of last few test accuracy results when the model converges. In the experiments, training stability is measured by the standard deviation of the last $A_{num}$ model accuracy values under logarithmic coordinates and $A_{num}$ was set as 10.

\subsection{Ablation Experiment Result and Analysis}\label{sec:ablation}

This part provides some ablation studies to demonstrate the impacts of each component of WKAFL. %In particular, we deployed ablation studies on CelebA and EMNIST ByClass datasets, which are both benchmark FL datasets.
Specifically, WKAFL consists of three main components: gradients selection, and the strategies for staleness and non-IIDness. Therefore, in our ablation experiments, we removed them one by one, and the corresponding algorithms are named as WKAFL(No Grad-Sel), WKAFL(No Staleness) and WKAFL(No Non-IIDness) respectively.

\begin{figure}[!t]
  \centering
  \subfloat[CelebA.]{
		\includegraphics[width=1.6in]{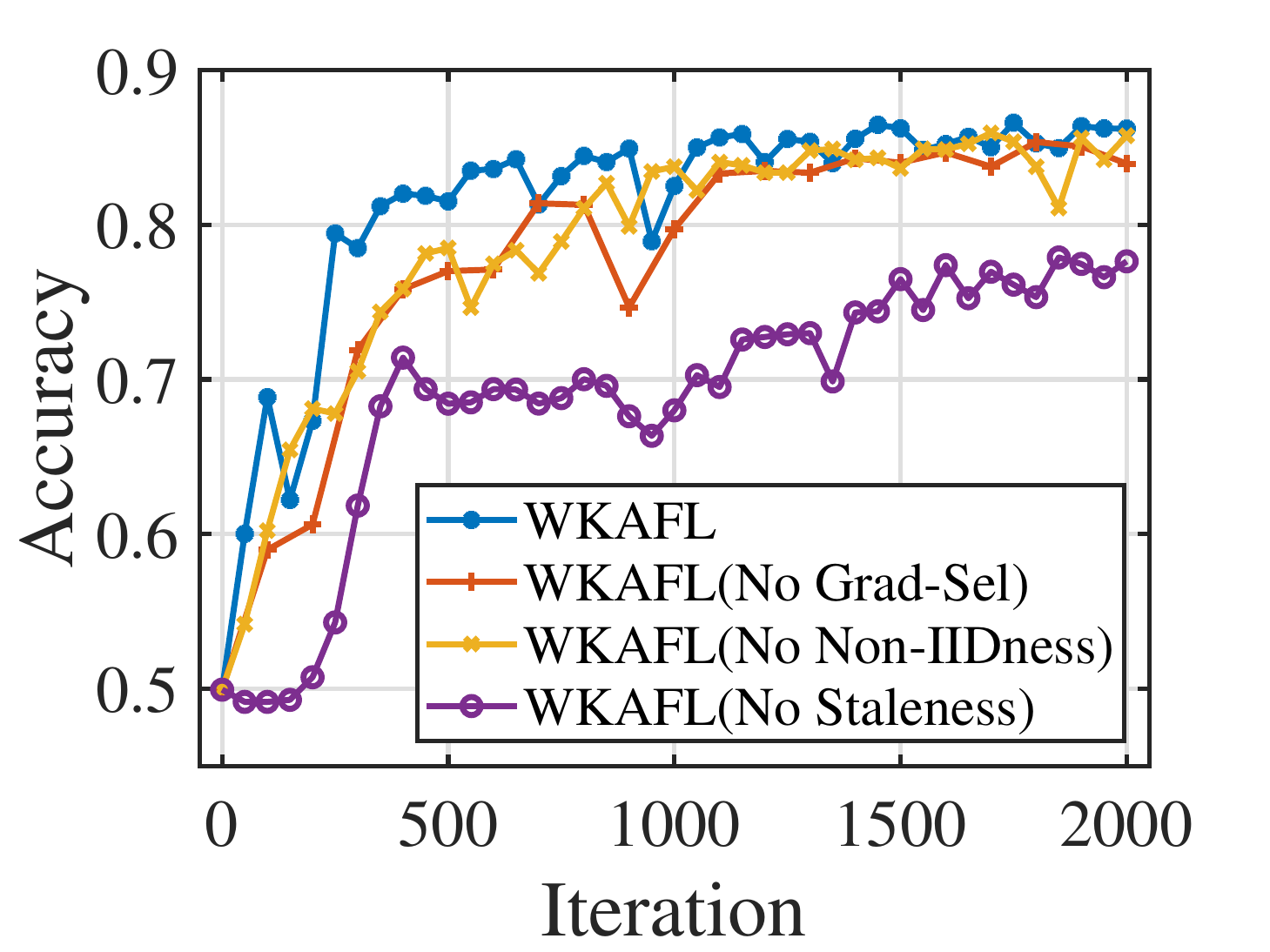}
		\label{Ablation:CelebA}
	}
	\hfil
  \subfloat[EMNIST ByClass.]{
		\includegraphics[width=1.6in]{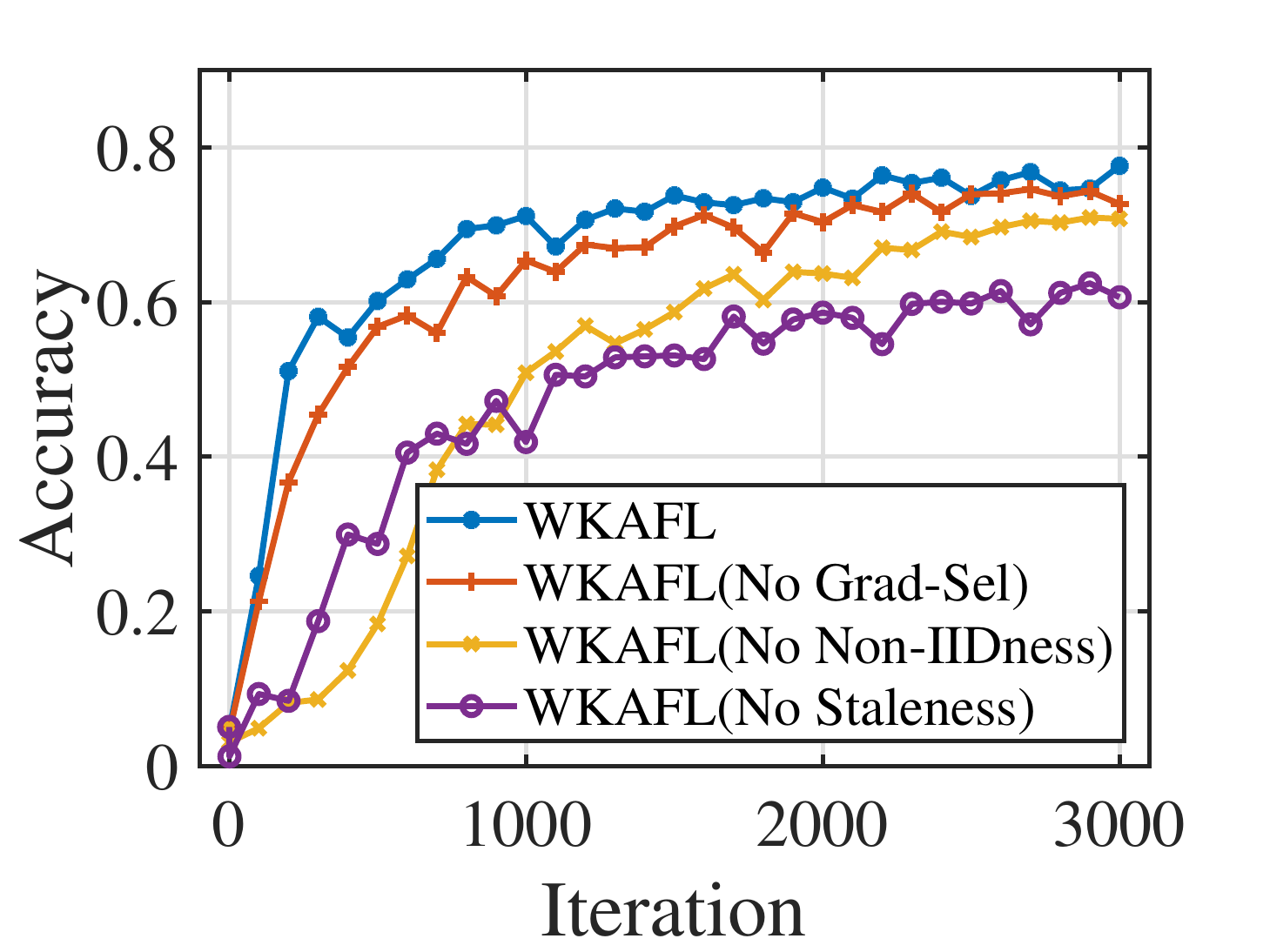}
		\label{Ablation:EMNIST}
	}
  \caption{Relation between test accuracy and iteration on CelebA and EMNIST ByClass.}\label{ablationExp}
\end{figure}

\figurename \ref{ablationExp} shows the test accuracy of four algorithms on both benchmark FL datasets CelebA and EMNIST ByClass. As shown, WKAFL can always converge fastest to the highest prediction accuracy, which indicates that every component in WKAFL is indispensable. WKAFL converges faster and more stable than WKAFL(No Grad-Sel), validating the existence of contradiction and the effectiveness of gradients selection. WKAFL(No Staleness) always has the poorest performance, which illustrates that staleness has more impacts on the model utility.

\subsection{Comparison Experiment Result and Analysis}\label{ExpWKAFL}
In this part, our purpose is to validate that the proposed WKAFL performs better not only on benchmark FL datasets, but also on modified FL datasets in terms of prediction accuracy and training stability.
Firstly, we implemented WKAFL and the comparison algorithms on two benchmark FL datasets, CelebA and EMNIST ByClass. Secondly, to comprehensively evaluate the performance of WKAFL, we implemented it on EMNIST MNIST and CIFAR10 which were modified to satisfy three levels of staleness and non-IID data respectively.

\begin{figure}[!t]
  \centering
  \subfloat[CelebA.]{
		\includegraphics[width=1.6in]{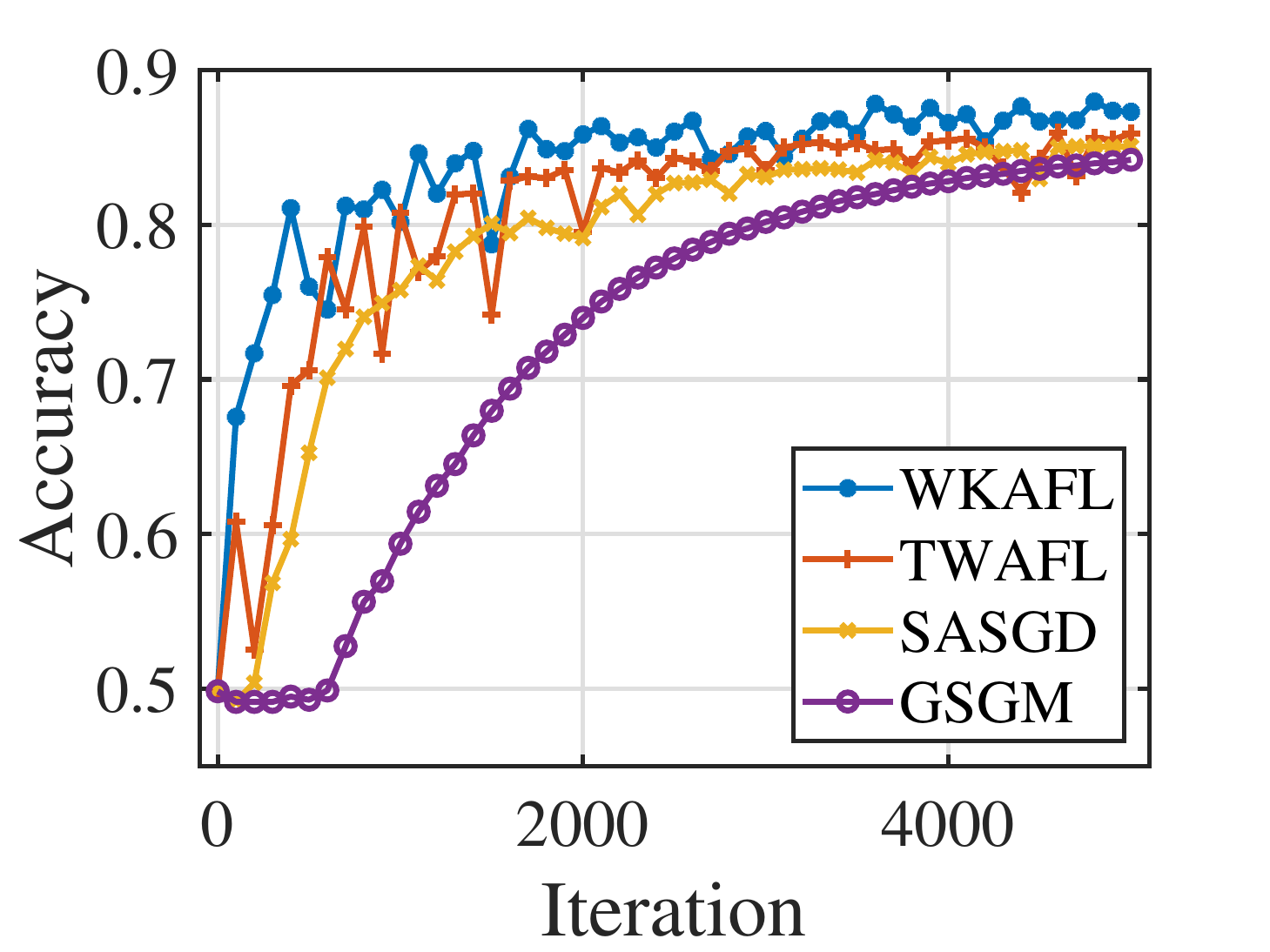}
		\label{Com:CelebA}
	}
	\hfil
  \subfloat[EMNIST ByClass.]{
		\includegraphics[width=1.6in]{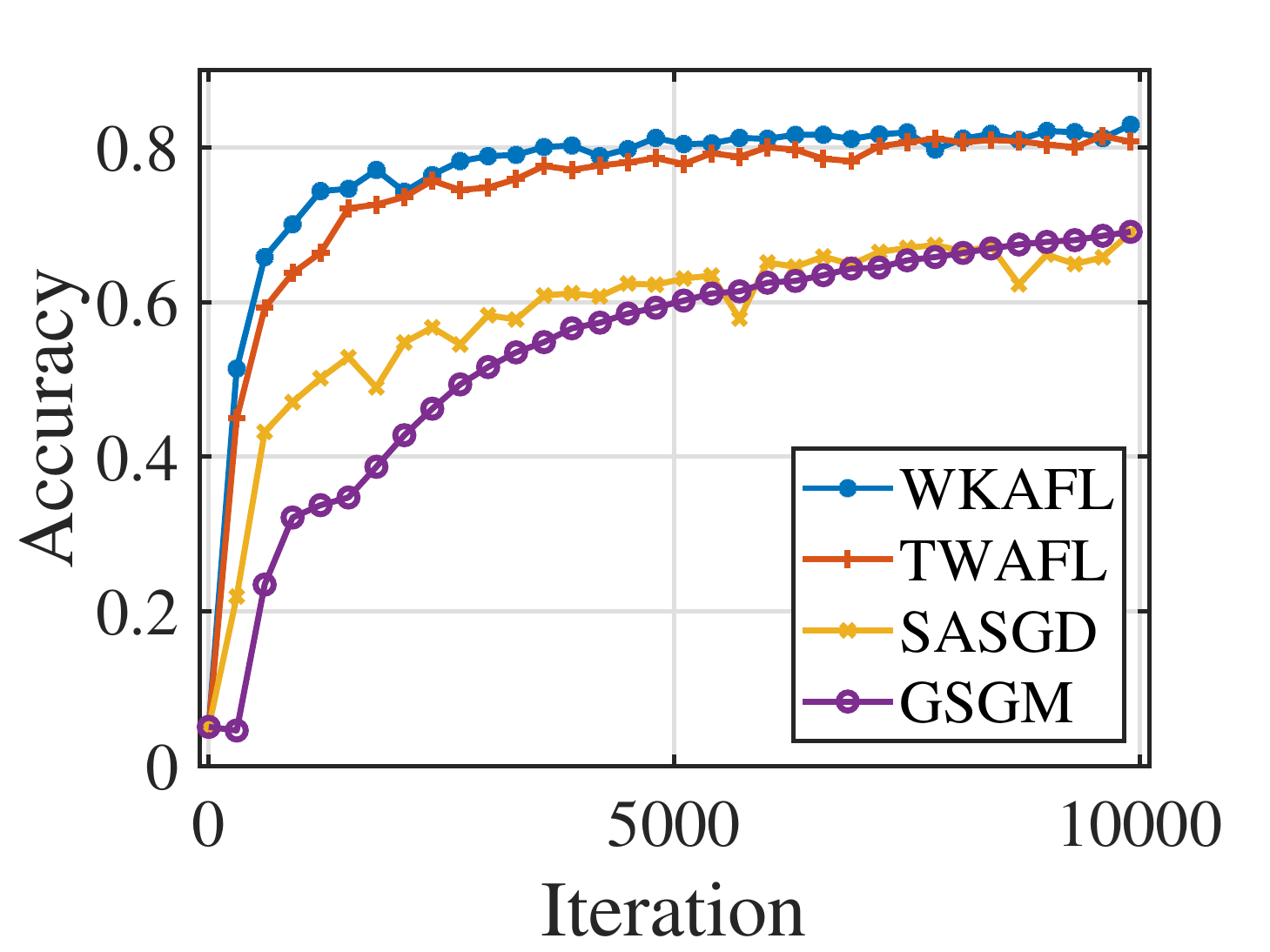}
		\label{Com:EMNIST}
	}
  \caption{Relation between test accuracy and iteration on CelebA and EMNIST ByClass.}\label{ER}
\end{figure}

\figurename \ref{ER} shows the test accuracy of four algorithms on CelebA and EMNIST ByClass. The training speed of WKAFL is faster than the other three algorithms in the beginning process and WKAFL can always achieve a higher accuracy which validates that the proposed WKAFL behaves well. The more comprehensive comparisons in terms of staleness and non-IID data are illustrated in Sections \ref{difStaleness} and \ref{difNonIID} respectively.

\begin{figure*}[!t]
  \centering
  \subfloat[$P/K = 100$ ($1000/10$).]{
      \includegraphics[width=2.3in]{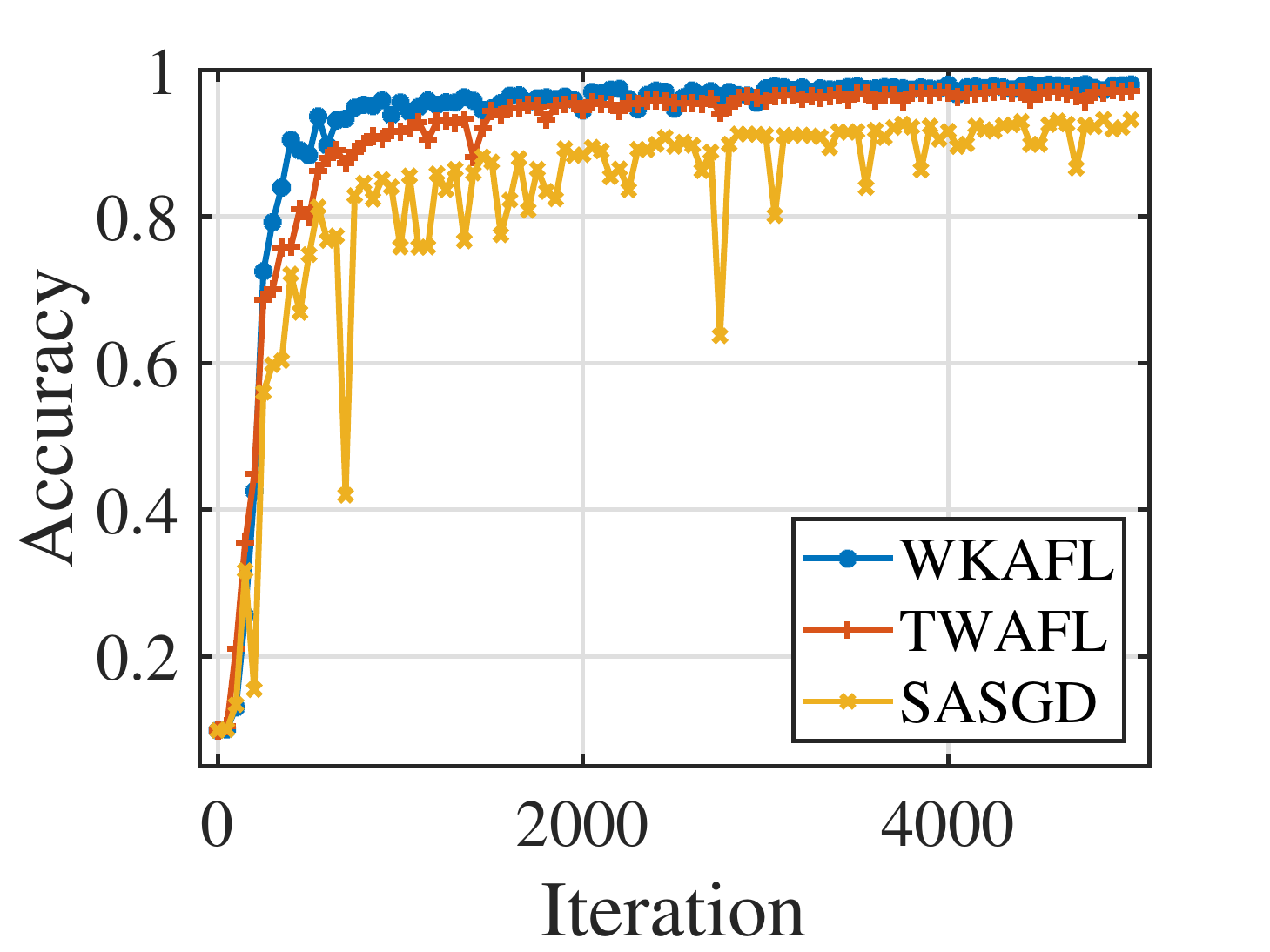}
	\label{EMNIST_1}
  }
  \hfil
  \subfloat[$P/K = 150$ ($3000/20$).]{
      \includegraphics[width=2.3in]{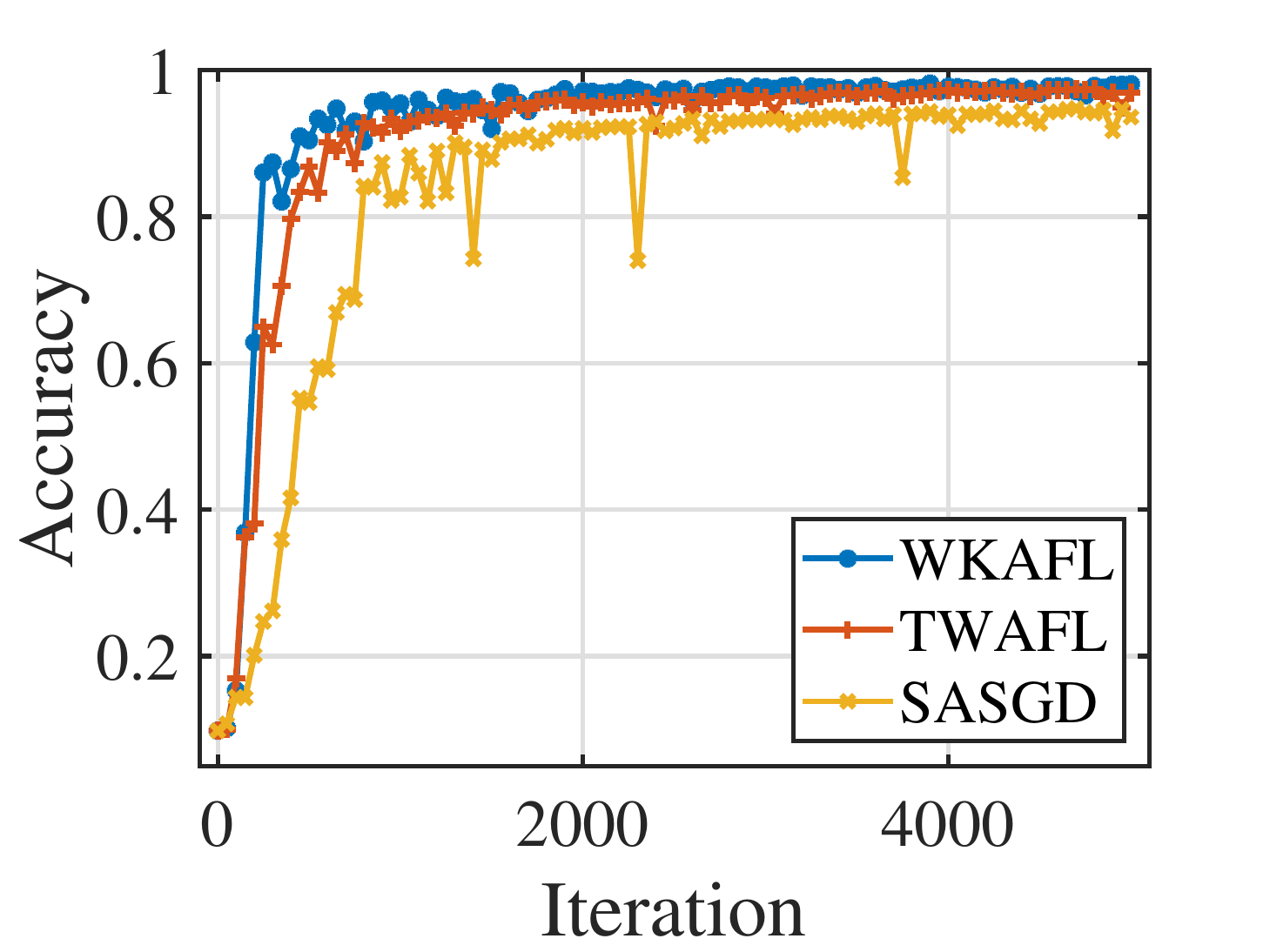}
	\label{MNIST_2}
  }
  \hfil
  \subfloat[$P/K = 300$ ($3000/10$).]{
      \includegraphics[width=2.3in]{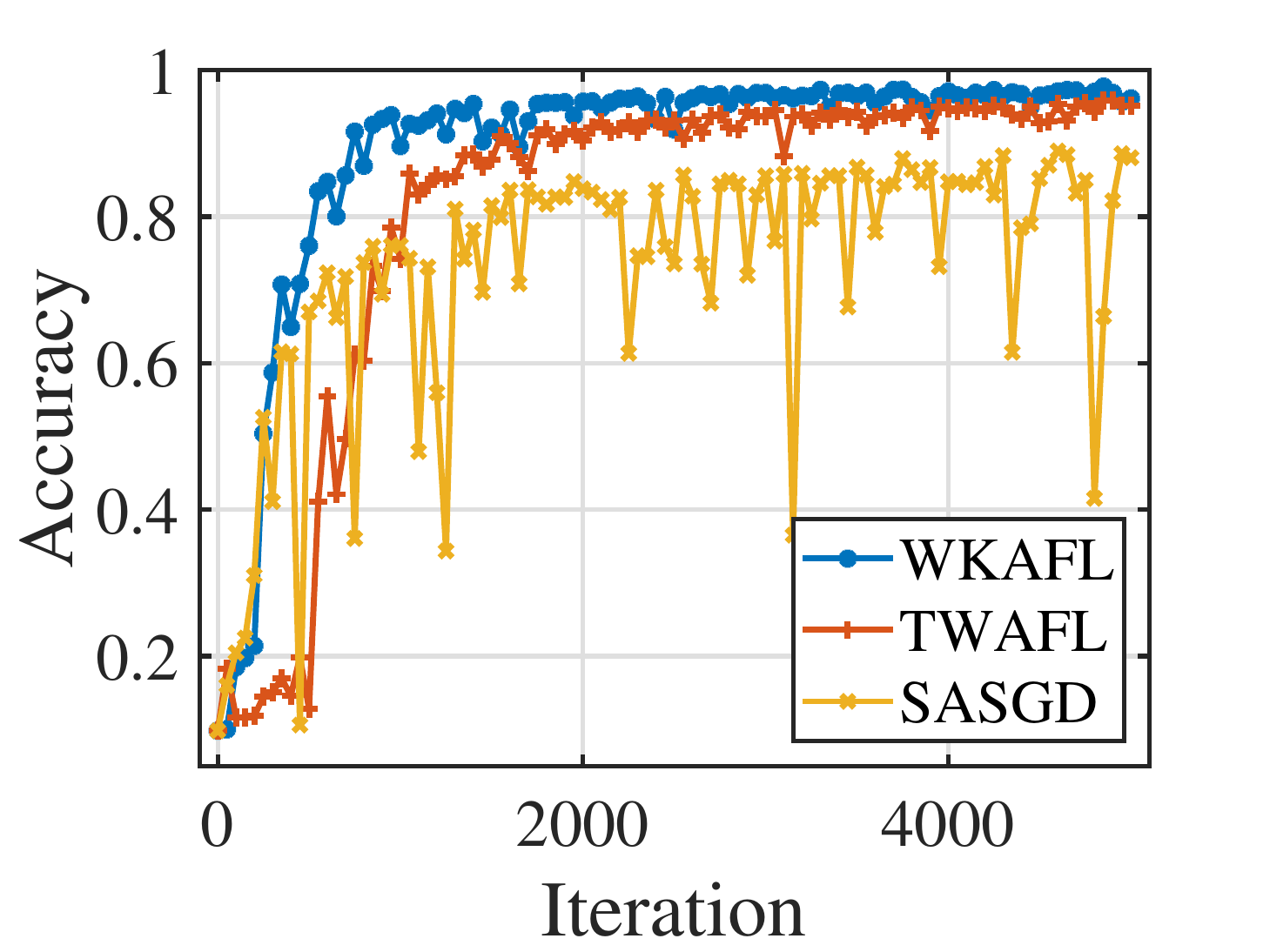}
	\label{EMNIST_3}
  }
  \caption{Prediction accuracy on EMNIST MNIST with different levels of staleness measured by $P/K$. Here $P$ is the number of total clients and $K$ is number of participants at each iteration.}\label{MNISTPK}
\end{figure*}
\begin{figure*}[!t]
  \centering
  \subfloat[$P/K = 50$ ($1000/20$).]{
      \includegraphics[width=2.3in]{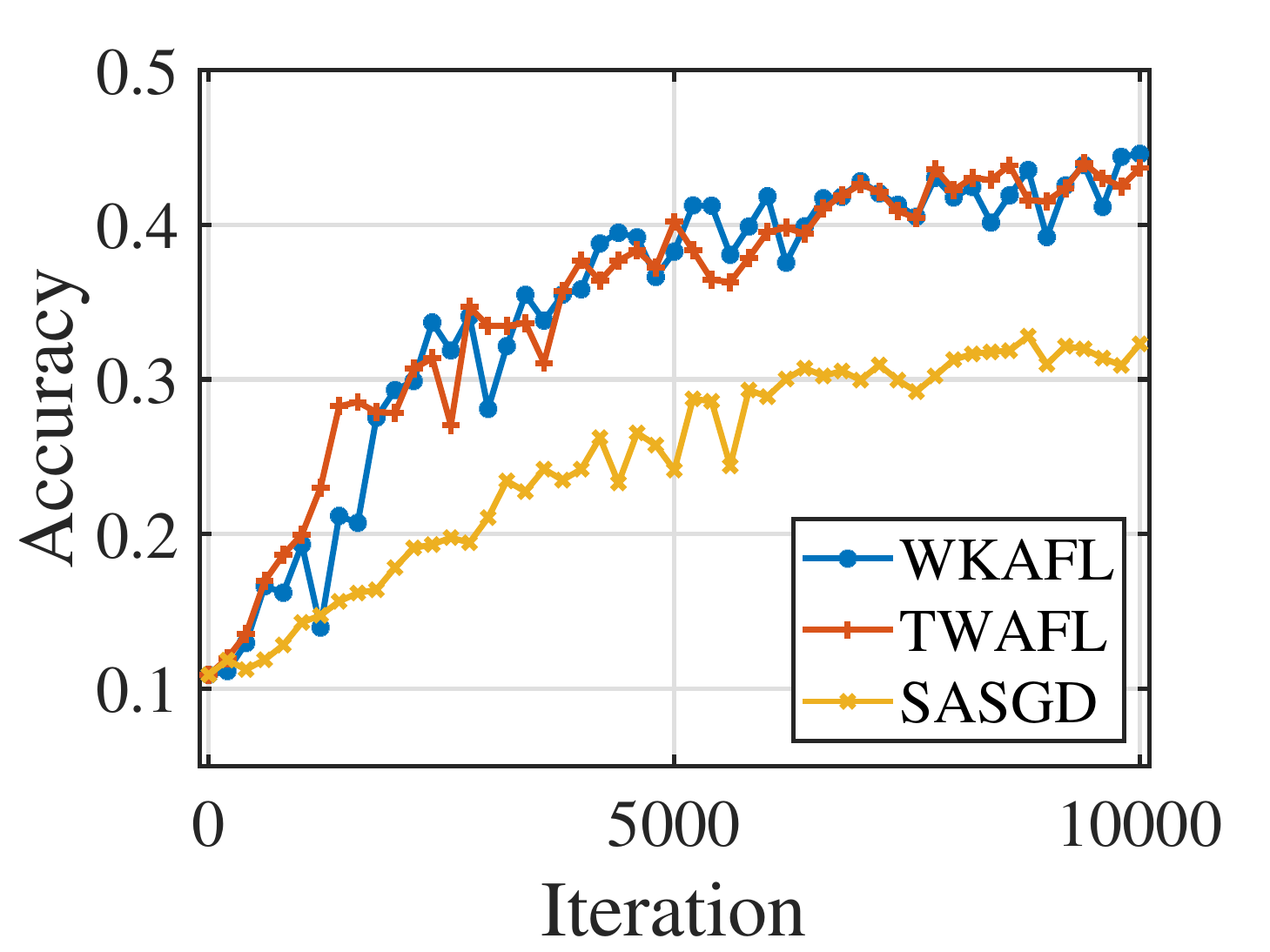}
	\label{REDDIT_1}
  }
  \hfil
  \subfloat[$P/K = 150$ ($3000/20$).]{
      \includegraphics[width=2.3in]{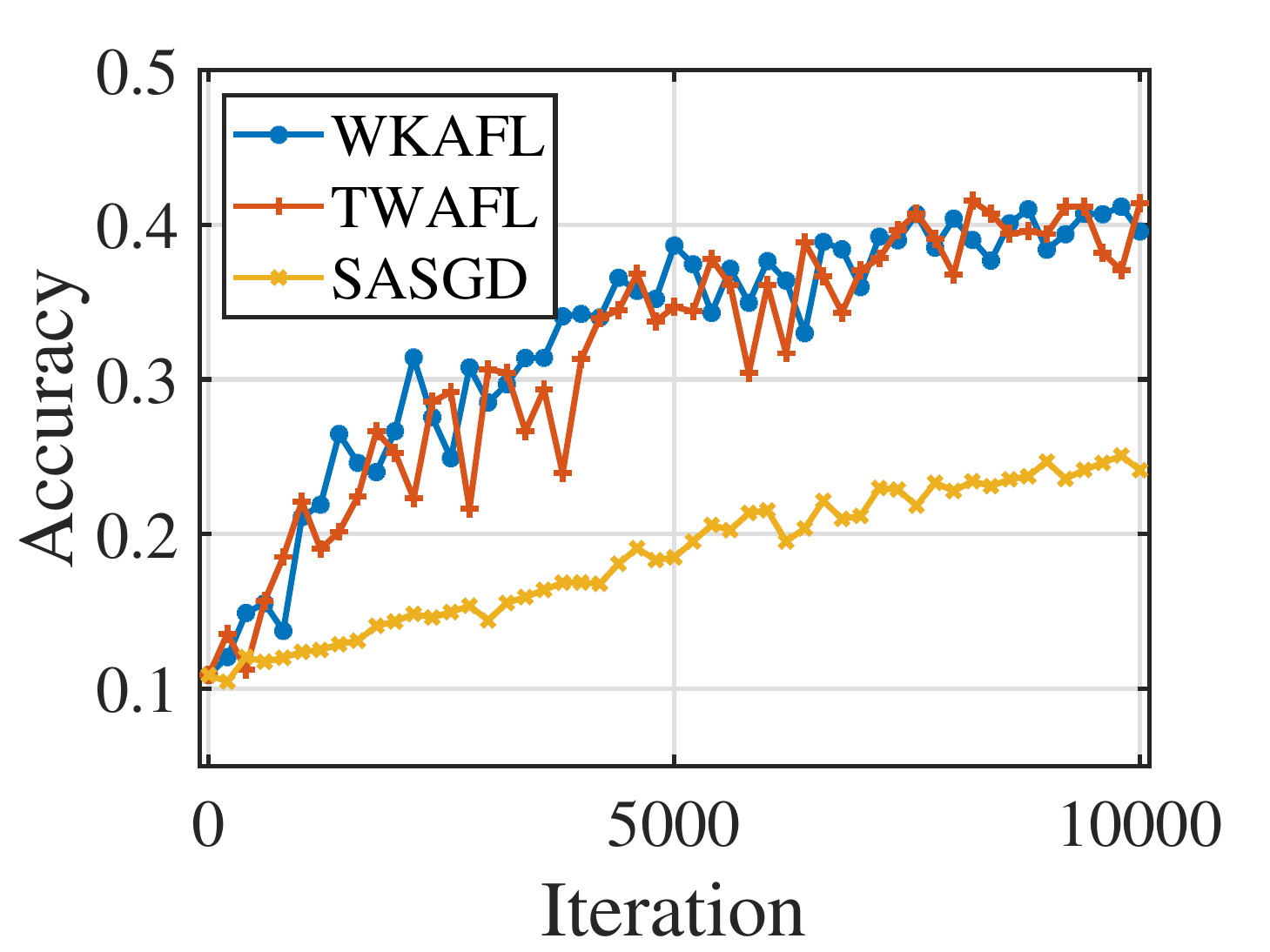}
	\label{REDDIT_2}
  }
  \hfil
  \subfloat[$P/K = 300$ ($3000/10$).]{
      \includegraphics[width=2.3in]{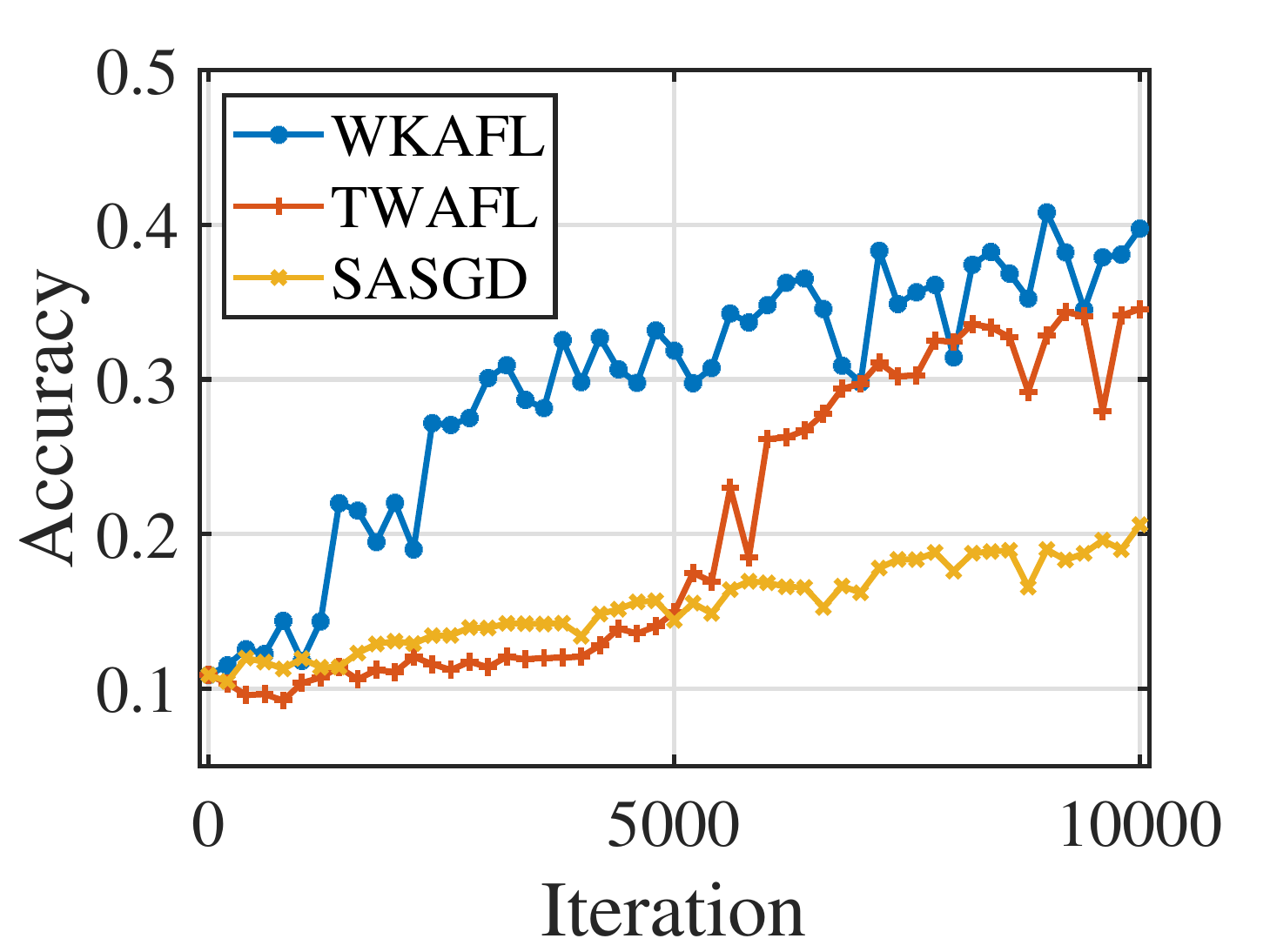}
	\label{REDDIT_3}
  }
  \caption{Prediction accuracy on CIFAR10 with different levels of staleness measured by $P/K$. Here $P$ is the number of total clients and $K$ is number of participants at each iteration.}\label{CIFAR10PK}
\end{figure*}

\subsubsection{Different Degrees of Staleness}\label{difStaleness}
Based on that the minimum average of staleness approximately equals $P/(2K)$ (refer to Appendix \ref{experiproof}), we achieve different levels of staleness by setting different values of $P$ (number of total clients) and $K$ (number of participants at each iteration) and the level of staleness can be measured as $P/K$ for convenience. For both EMNIST MNIST and CIFAR10, we set three levels of staleness, $1000/10=100, 3000/20=150$, $3000/10=300$ on EMNIST MNIST and $1000/20=50, 3000/20=150$, $3000/10=300$ on CIFAR10. The experimental results of test accuracy under the three levels on EMNIST MNIST and CIFAR10 is shown in \figurename \ref{MNISTPK} and \figurename \ref{CIFAR10PK} respectively and the final prediction accuracy is summarized in Table \ref{ASStaleness}.

\begin{table}[!ht]
  \centering
  \caption{Prediction Accuracy with Different Levels of Staleness.}\label{ASStaleness}
  \begin{tabular}{ccccc}
	\hline
	\multirow{2}*{Datasets} & \multirow{2}*{Staleness} &
	\multicolumn{3}{c}{Prediction Accuracy}  \\ \cline{3-5}
	& & WKAFL  & TWAFL & SASGD  \\\hline
	\multirow{3}*{EMNIST MNIST} & $100$ & \textbf{0.9734} & 0.962 & 0.9319 \\
    & $150$ & \textbf{0.9756} & 0.9649& 0.8753  \\
	& $300$ & \textbf{0.9728} & 0.9572& 0.8553  \\ \hline
	\multirow{3}*{CIFAR10} & $50$ & \textbf{0.4433} & 0.4246 & 0.3308\\		
		& $150$ & \textbf{0.4368} & 0.4241 & 0.2560 \\
		& $300$ & \textbf{0.3970} & 0.3468& 0.2029 \\
	\hline
  \end{tabular}
\end{table}

\textbf{Comparisons on EMNIST MNIST.} Three conclusions are obtained from \figurename \ref{MNISTPK} and Table \ref{ASStaleness} (rows 3-5).
\textit{Firstly}, WKAFL has the fastest training speed compared to algorithms TWAFL and SASGD under the three levels of staleness as shown in \figurename \ref{MNISTPK}, especially for high staleness in \figurename \ref{EMNIST_3}. This validates the idea of exploiting the stale gradients to accelerate the training process (as discussed in Section \ref{select}).
\textit{Secondly}, \figurename \ref{MNISTPK} shows that WKAFL has higher prediction accuracy than the compared algorithms, especially for high staleness which benefits from the strategies of clipping bound the stale gradients (as discussed in Section \ref{estimate}), selecting consistent gradients (as discussed in Section \ref{select}) and adaptively adjusting the learning rate (as discussed in Section \ref{adjust}), especially when the model is going to converge (Stage two).
\textit{Thirdly}, WKAFL is more robust to staleness. Based on the decreasing prediction accuracy in each column (rows 3-5) in Table \ref{ASStaleness}, the staleness has a negative impact on the prediction accuracy for all algorithms which is consistent with the analysis of the negative effect of staleness (Theorems \ref{stageone} and \ref{stagetwo}). We note that the accuracy reduction of WKAFL is the lowest while the accuracy reduction of SASGD is the highest. Apart from it, the curves of TWAFL and SASGD have more obvious changes than the curve of WKAFL when the level of staleness increases as shown in \figurename \ref{MNISTPK}.

\begin{table*}[!ht]
  \centering
  \caption{Prediction Accuracy and Training Stability with Three Levels of Non-IID Degree Measured by $L_{num}$.}\label{ASNonIID}
  \begin{tabular}{ccccccccccccccc}
	\hline
	\multirow{2}*{Dataset} & \multirow{2}*{$L_{num}$} &
    \multicolumn{4}{c}{Final Accuracy} &  \multirow{2}*{} & \multicolumn{4}{c}{Model Stability}  \\ \cline{3-6}\cline{8-11}
	& & WKAFL  & TWAFL & GSGM & SASGD & & WKAFL  & TWAFL & GSGM & SASGD \\
	\hline
    \multirow{3}*{EMNIST MNIST} & 1 &  \textbf{0.9728} & 0.9572& 0.9057& 0.8553 & & 0.0060 & 0.0107& \textbf{0.0032}& 0.0834 \\
    & 5 &  \textbf{0.9754} & 0.9649& 0.908 & 0.8753 &  & 0.0068 & 0.0095& \textbf{0.0027}& 0.0237\\
    & 10 &  \textbf{0.9658} & 0.9623& 0.9075& 0.8619 & & 0.0036 & 0.0060& \textbf{0.0018}& 0.0206 \\ \hline
    \multirow{3}*{CIFAR10} & 1 & \textbf{0.4368} &0.4141 & 0.3568 & 0.2560 & & 0.0284 & 0.0464& \textbf{0.0135}& 0.0226  \\
    & 5 & 0.4443 &\textbf{0.4454} & 0.3568 & 0.2855 & & 0.0133 & 0.0300& \textbf{0.0083}& 0.0141  \\
    & 10 & 0.4489 & \textbf{0.4525}& 0.3430 & 0.2874 & & 0.0180 & 0.0260& \textbf{0.0144}& 0.0189 \\
	\hline
  \end{tabular}
\end{table*}

\begin{figure*}[!t]
  \centering
  \subfloat[$L_{num} =1$]{
     \includegraphics[width=2.3in]{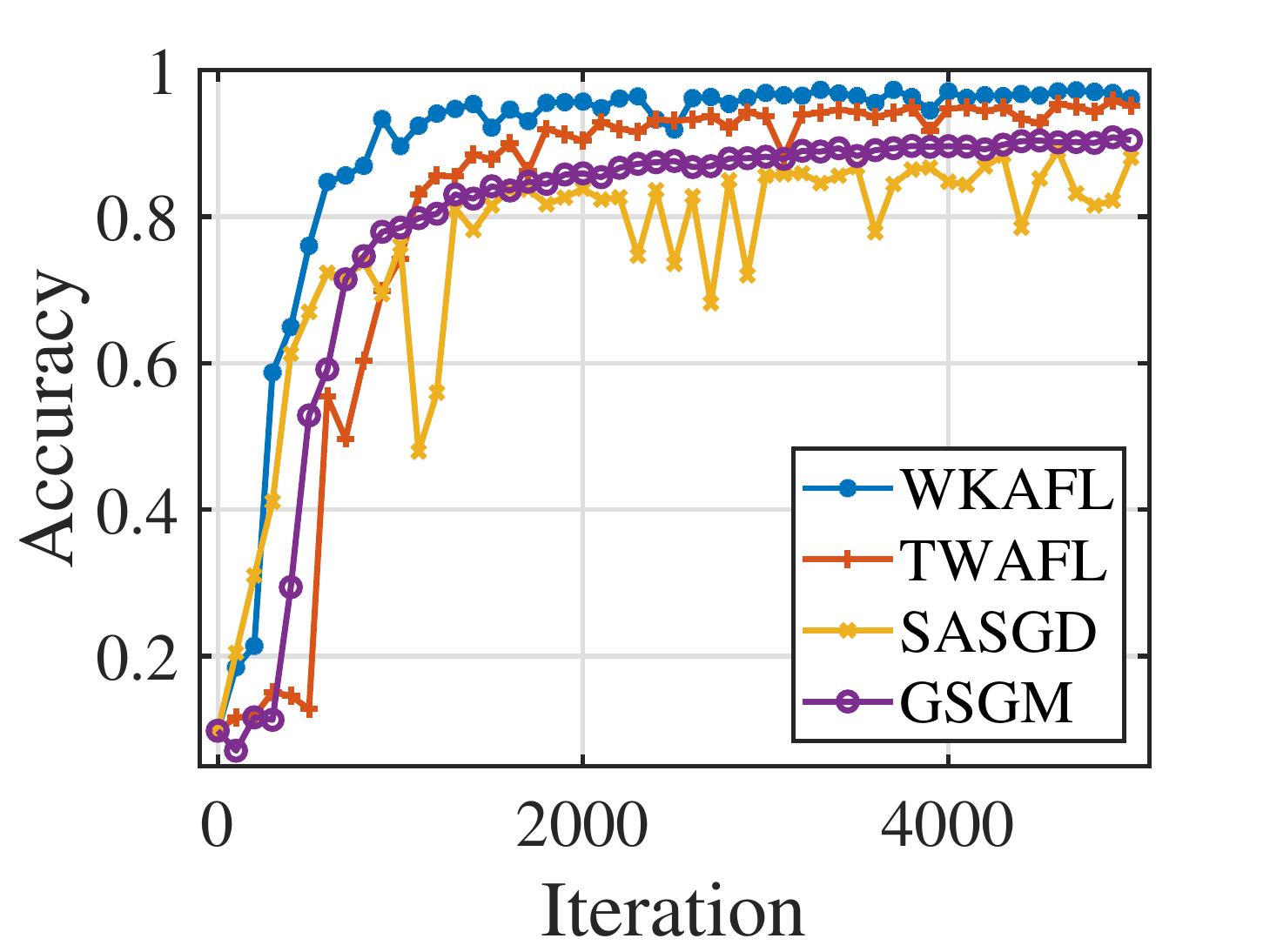}
    \label{MNIST_NonIID_1}
  }
  \hfil
  \subfloat[$L_{num} = 5$]{
    \includegraphics[width=2.3in]{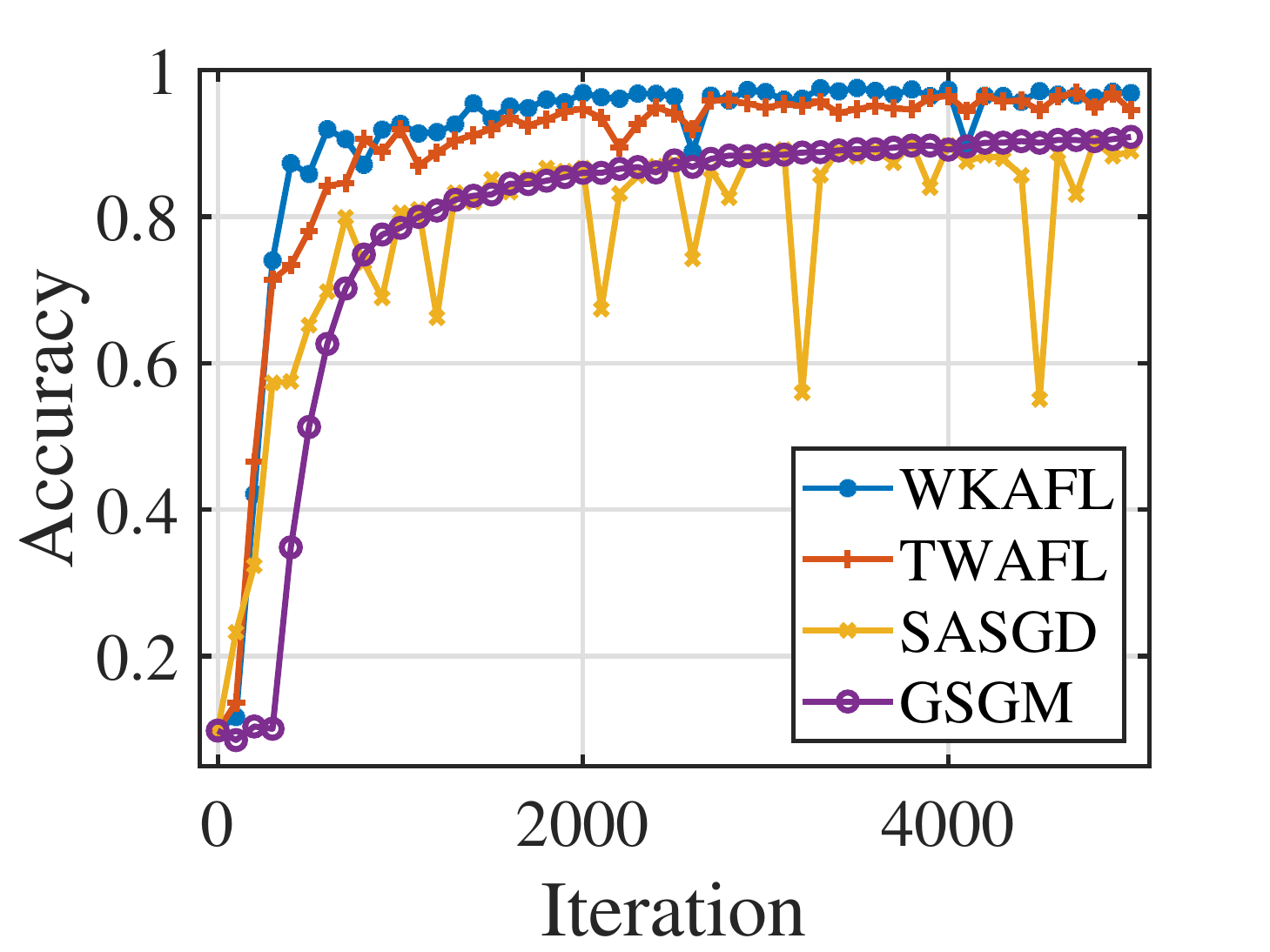}
	\label{MNIST_NonIID_5}
  }
  \hfil
  \subfloat[$L_{num} = 10$]{
    \includegraphics[width=2.3in]{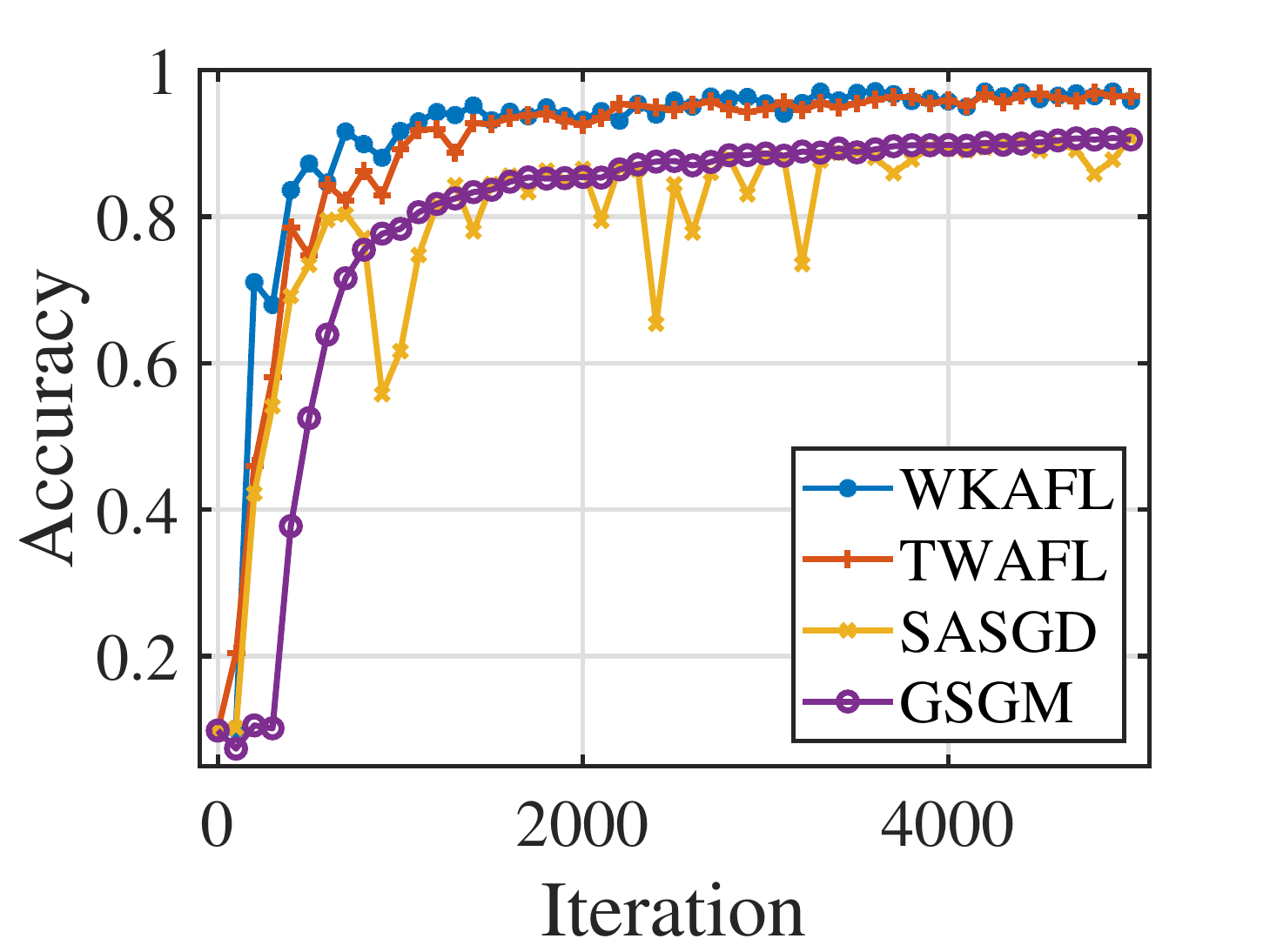}
    \label{MNIST_NonIID_10}
  }
  \caption{Relation between test accuracy and iteration on EMNIST MNIST with different levels of Non-IID degrees ($L_{num}$). A smaller $L_{num}$ means a higher level of non-IID data. The level of staleness was fixed as $300$ ($P/K=3000/10$).}\label{MNIST_NonIID}
\end{figure*}

\begin{figure*}[!t]
  \centering
  \subfloat[$L_{num} =1$]{
    \includegraphics[width=2.3in]{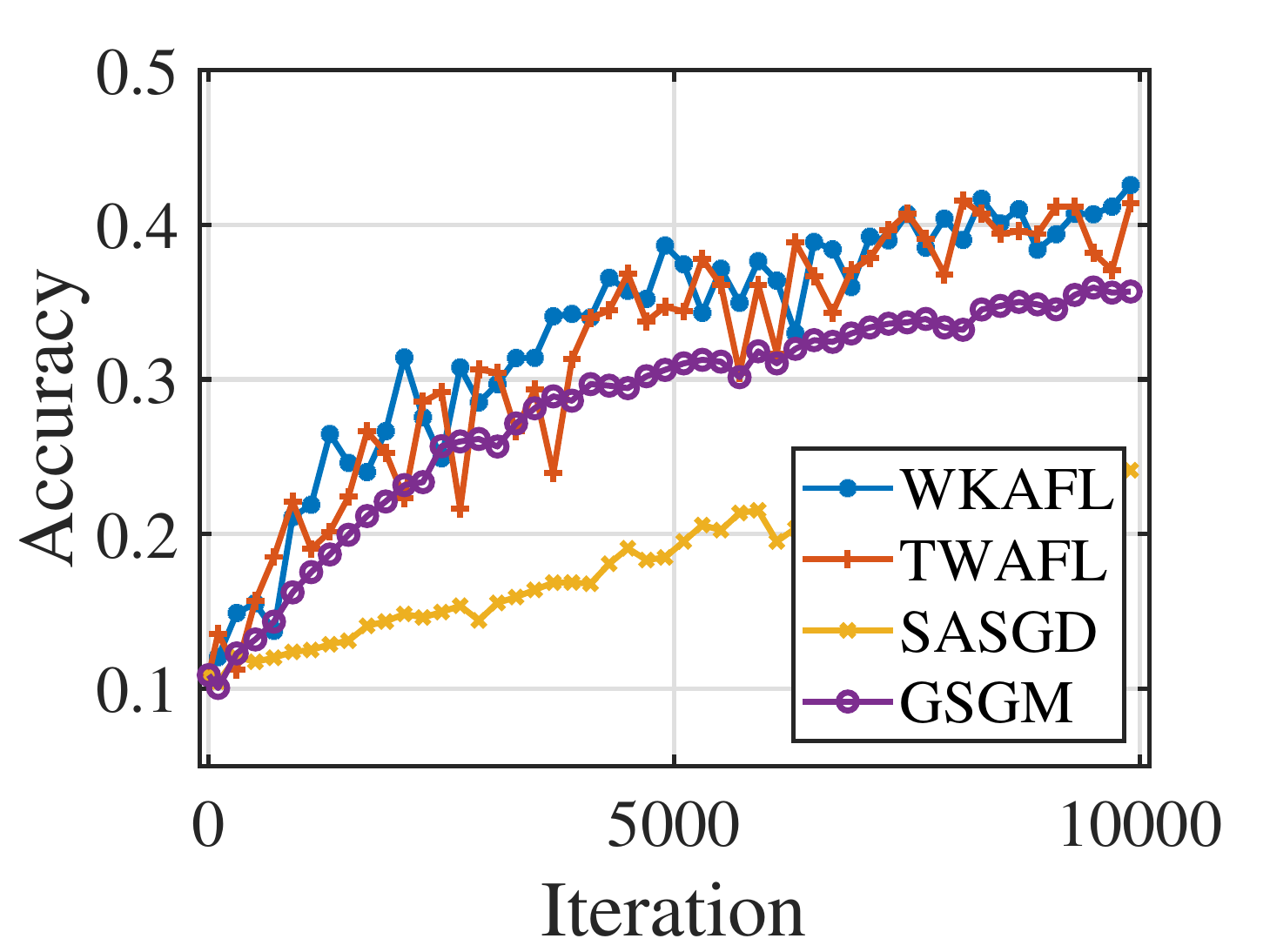}
	\label{CIFAR10_NonIID_1}
  }
  \hfil
  \subfloat[$L_{num} = 5$]{
    \includegraphics[width=2.3in]{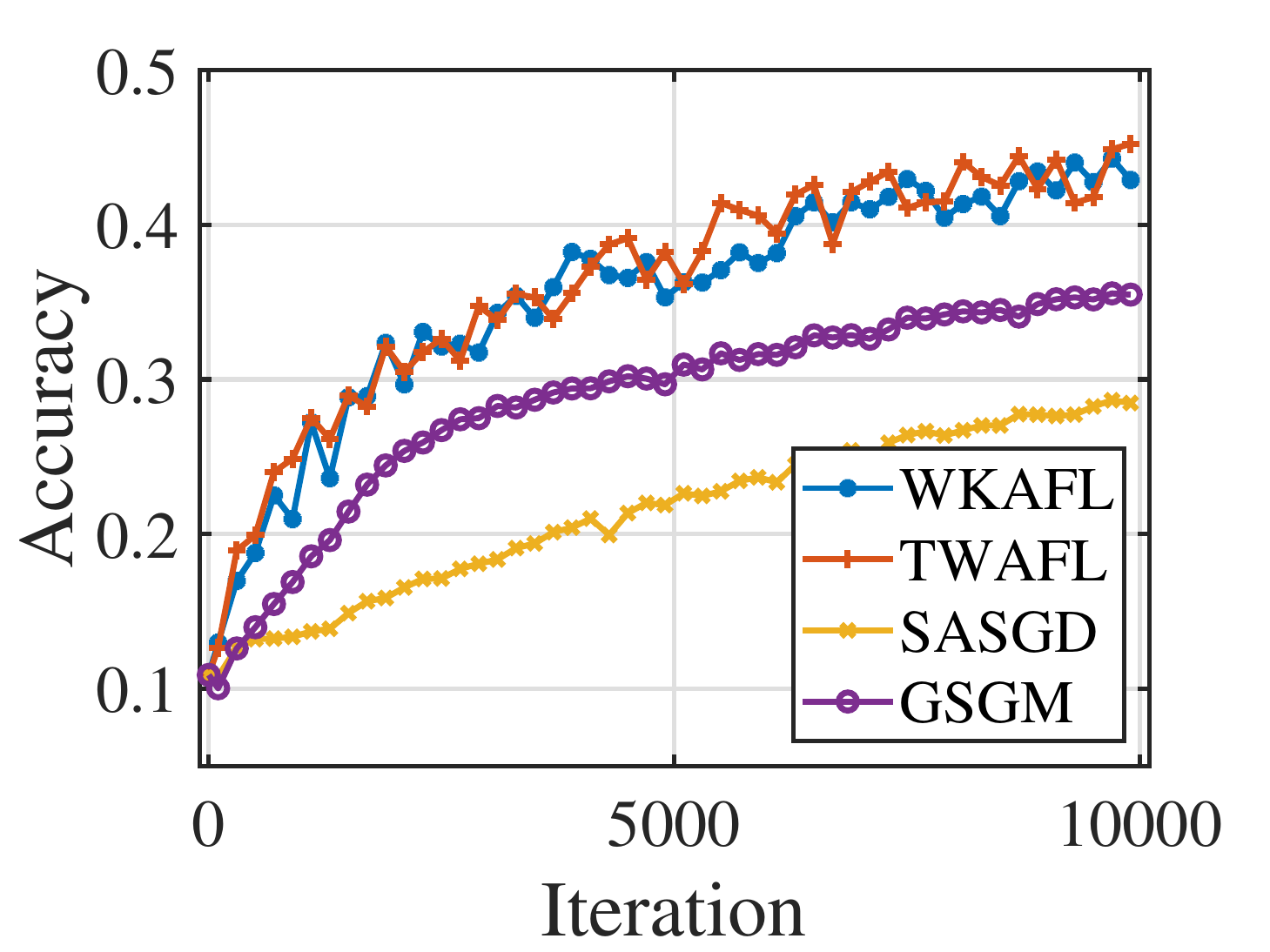}
	\label{CIFAR10_NonIID_5}
  }
  \hfil
  \subfloat[$L_{num} = 10$]{
    \includegraphics[width=2.3in]{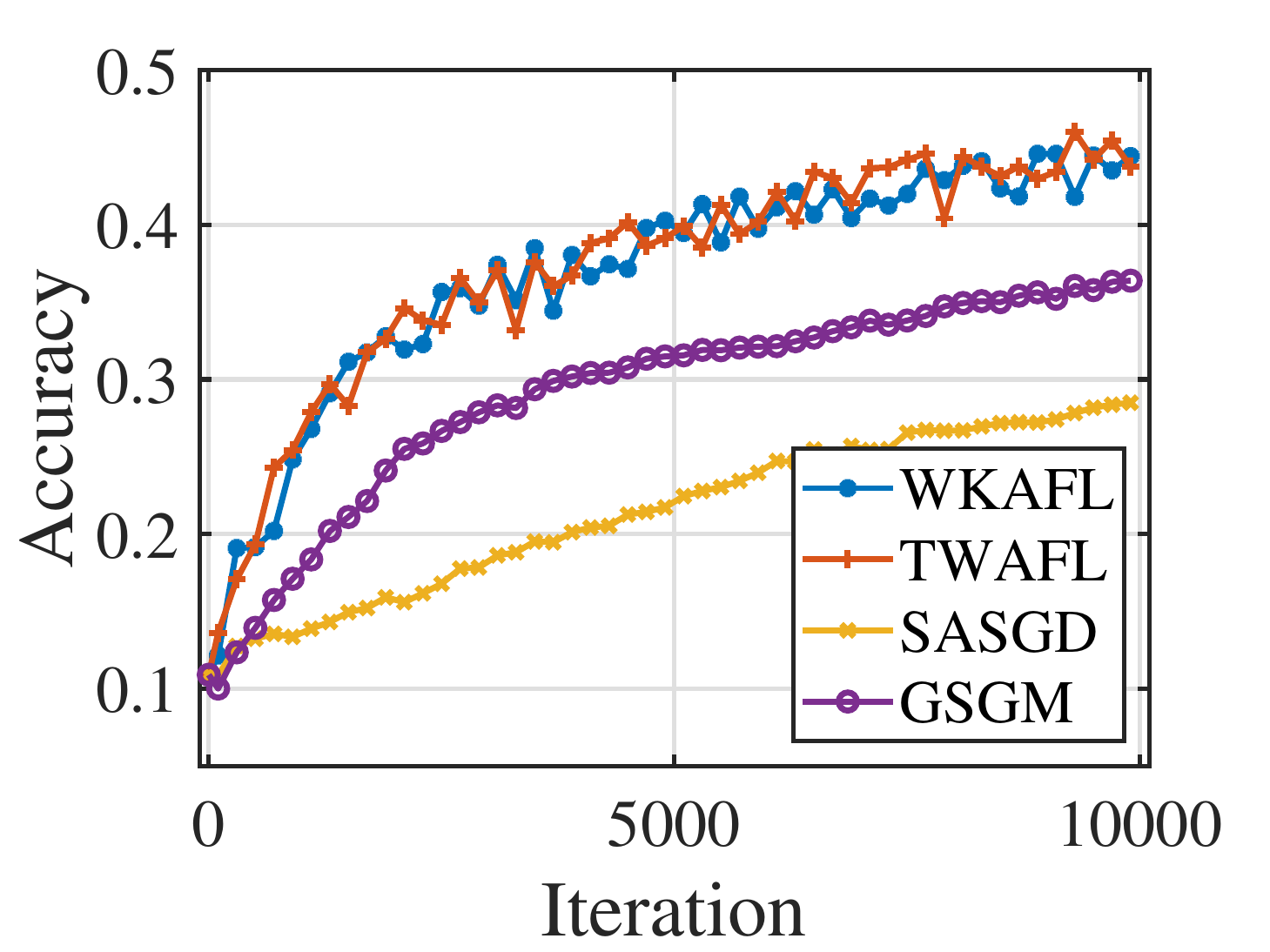}
	\label{CIFAR10_NonIID_10}
  }
  \caption{Relation between test accuracy and iteration on CIFAR10 with different levels of Non-IID degrees ($L_{num}$). A smaller $L_{num}$ means a higher level of non-IID data. The level of staleness was fixed as $150$ ($P/K=3000/20$).}\label{CIFAR10_NonIID}
\end{figure*}

\textbf{Comparisons on CIFAR10.} Similar results are obtained from \figurename \ref{CIFAR10PK} and Table \ref{ASStaleness} (rows 6-8). When it comes to dealing with stale gradients and using the adaptive learning rate, the proposed WKAFL outperforms the compared algorithms in terms of training speed and prediction accuracy. Besides, Table \ref{ASStaleness} shows that the robustness advantage of WKAFL is more significant than that on EMNIST MNIST. Particularly, when staleness increases from $100$ to $300$, the accuracy of WKAFL decreases from $0.4433$ to $0.3970$ while the accuracy significantly decreases from $0.4246$ to $0.3468$ (from $0.3308$ to $0.2029$)for TWAFL (SASGD) algorithms which validates that WKAFL is extensible for FL with large scale distributed clients in which the staleness is usually high.

Additionally, we explain the reason of low prediction accuracy (about 44.5 percent) on CIFAR10 achieved here which is significantly lower than the existing high value (around 95 percent) \footnote{https://github.com/junyuseu/pytorch-cifar-models} . The main reason is the memory restrictions of using the complicated CNN architecture presented in \cite{huang2017densely}, \cite{xie2017aggregated}. To validate the proposed WKAFL for scenarios with large scale distributed devices, the number of total clients $P$ was set as $3000$. In AFL, $3000$ duplicates of CNN are needed to be stored and computed because of the asynchronous manner. However, this requires a computer with large memory size. Therefore, we adopted a light model LeNet whose prediction accuracy under centralized machine learning is only around 67 percent \footnote{https://github.com/icpm/pytorch-cifar10}. Based on it, the achieved decentralized accuracy affected by both non-IID data and high staleness is acceptable and our main purpose is to validate the performance of WKAFL for any given model and dataset.

\subsubsection{Different Degrees of Non-IID Data}\label{difNonIID}
Non-IID data is a basic characteristic of FL, however, there is no exact metrics of it. As adopted in \cite{liu2020accelerating}, we also use the number of label classes $L_{num}$ to measure non-IID level. For example, for CIFAR10 which has labels $0,\cdots,9$, $L_{num}=1$ means that each client only owns data with one label, such as $5$. Obviously, a small value $L_{num}$ means high level of non-IID data. We set three levels of data heterogeneity on both EMNIST MNIST and CIFAR10, $L_{num}=1,5,10$. The experimental results of test accuracy on EMNIST MNIST and CIFAR10 are shown in \figurename \ref{MNIST_NonIID} and \figurename \ref{CIFAR10_NonIID} respectively. Table \ref{ASNonIID} shows the final prediction accuracy and training stability.

\textbf{Comparisons on EMNIST MNIST.} Four conclusions are drawn from \figurename \ref{MNIST_NonIID} and Table \ref{ASNonIID}.

\textit{Firstly}, \figurename \ref{MNIST_NonIID} shows that the proposed WKAFL converges faster than other three algorithms for all three levels of data heterogeneity, especially for the highest level (\figurename \ref{MNIST_NonIID_1} with $L_{num}=1$). This advantage benefits from full use of historical gradients (as discussion in Section \ref{decrease}). As the level of non-IID data increases, the direction of uploaded gradients will seriously deviate from the globally unbiased gradient and accumulated gradients can narrow the gap between uploaded gradients and consistent gradients. Therefore, the advantage of WKAFL is more significant for the high level of non-IID data.

\textit{Secondly}, the proposed WKAFL achieves the highest prediction accuracy for all three levels of data heterogeneity as shown in columns 3-6 of Table \ref{ASNonIID} (rows 3-5) which illustrates that exploiting the historical gradients can alleviate the impact of non-IID data (as explained in Section \ref{decrease}), leading to a higher accuracy.

\textit{Thirdly}, the training stability decreases as the level of non-IID data increases, i.e., $L_{num}$ from $10, 5$ to $1$, as shown in columns 7-10 of Table \ref{ASNonIID} (rows 3-5). This is because the gradient's direction of non-IID data is likely to be deviated from the globally unbiased gradient. At different iterations, the direction of aggregated gradients based on $K$ participants will differently deviate from the globally unbiased gradient. Therefore, higher heterogeneity means more uncertainty. Experiment results validate our analysis (Section \ref{analysis}).

\textit{Fourthly}, WKAFL has a better training stability as shown in columns 7-10 of Table \ref{ASNonIID} (row 3-5). A small value of training stability means a stable model. Although WKAFL does not have the best training stability, the comprehensive advantage of WKAFL is still significant, which can be described from two aspects. On the one hand, compared to GSGM with the best training stability, its prediction accuracy (row 3-5 in Table \ref{ASNonIID}) and convergence rate (Fig. \ref{MNIST_NonIID}) is significantly lower than WKAFL. For example, when $L_{num}=1$, the prediction accuracy for GSGM is only 85.53 percent, significantly lower than 97.28 percent for WKAFL. However, the training stability of WKAFL, i.e., the fluctuation degree in \figurename \ref{MNIST_NonIID_5} and \figurename \ref{MNIST_NonIID_10}, is almost the same as GSGM. On the other hand, compared to TWAFL who has a slight prediction accuracy reduction, its training stability is significantly weaker than WKAFL. For example, when $L_{num}=1$, the training stability for TWAFL is $0.0107$, almost with 44 percent increment than $0.006$ for WKAFL. Besides, \figurename \ref{MNIST_NonIID_1} shows that the fluctuation of TWAFL is more obvious than WKAFL.

\textbf{Comparisons on CIFAR10}. Similar conclusions are drawn from \figurename \ref{CIFAR10_NonIID} and Table \ref{ASNonIID}, based on the same analysis on EMNIST MNIST. The only difference is that TWAFL achieves a slight improvement of prediction accuracy than the proposed WKAFL when $L_{num}=5,10$ (columns 2-3) because of the randomness in the experiments. In fact, TWAFL and WKAFL have almost the same training process in low level of data heterogeneity as shown in \figurename \ref{CIFAR10_NonIID_5} and \figurename \ref{CIFAR10_NonIID_10}.

\section{Conclusion} \label{conclusion}
In this paper, we propose a two-stage weighted $K$-async FL (WKAFL) algorithm to improve the model utility of AFL %by exploiting staleness and improving the effects for alleviating non-IID data.
These improvements are achieved from three aspects.
Firstly, WKAFL estimates the globally unbiased gradient by accumulating historical gradients to alleviate the impact of non-IID data and aggregating $K$ gradients based on the staleness.
Secondly, WKAFL picks gradients consistent with the estimated gradient and assigns them with a high weight and vice versa to improve the effect for mitigating non-IID data while preventing the stale gradients to bring down model utility.
Thirdly, by further clipping the stale gradient in the second stage and adjusting the learning rate based on staleness, WKAFL improves the final prediction accuracy.
The experiment results on four FL datasets validate that WKAFL can accelerate the training process and improve final prediction accuracy while guaranteeing a stable model, especially in settings with high staleness or high level of non-IID data.

\bibliographystyle{IEEEtran}
{\small \bibliography{ref}}

\begin{IEEEbiography}[{\includegraphics[width=0.85in,height=1.25in, clip,keepaspectratio]{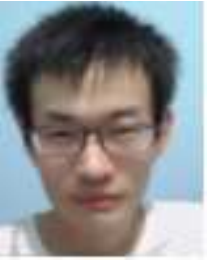}}] {Zihao Zhou} received his Bachelor degree from the school of mathematics and statistics at Xi'an Jiaotong University of China in 2020. He is currently working towards the PhD degree in the School of Mathematics and Statistics at Xi'an Jiaotong University. His research interests include federated learning and edge-cloud intelligence.
\end{IEEEbiography}
\vspace{-1cm}

\begin{IEEEbiography}[{\includegraphics[width=1in,height=1.25in, clip,keepaspectratio]{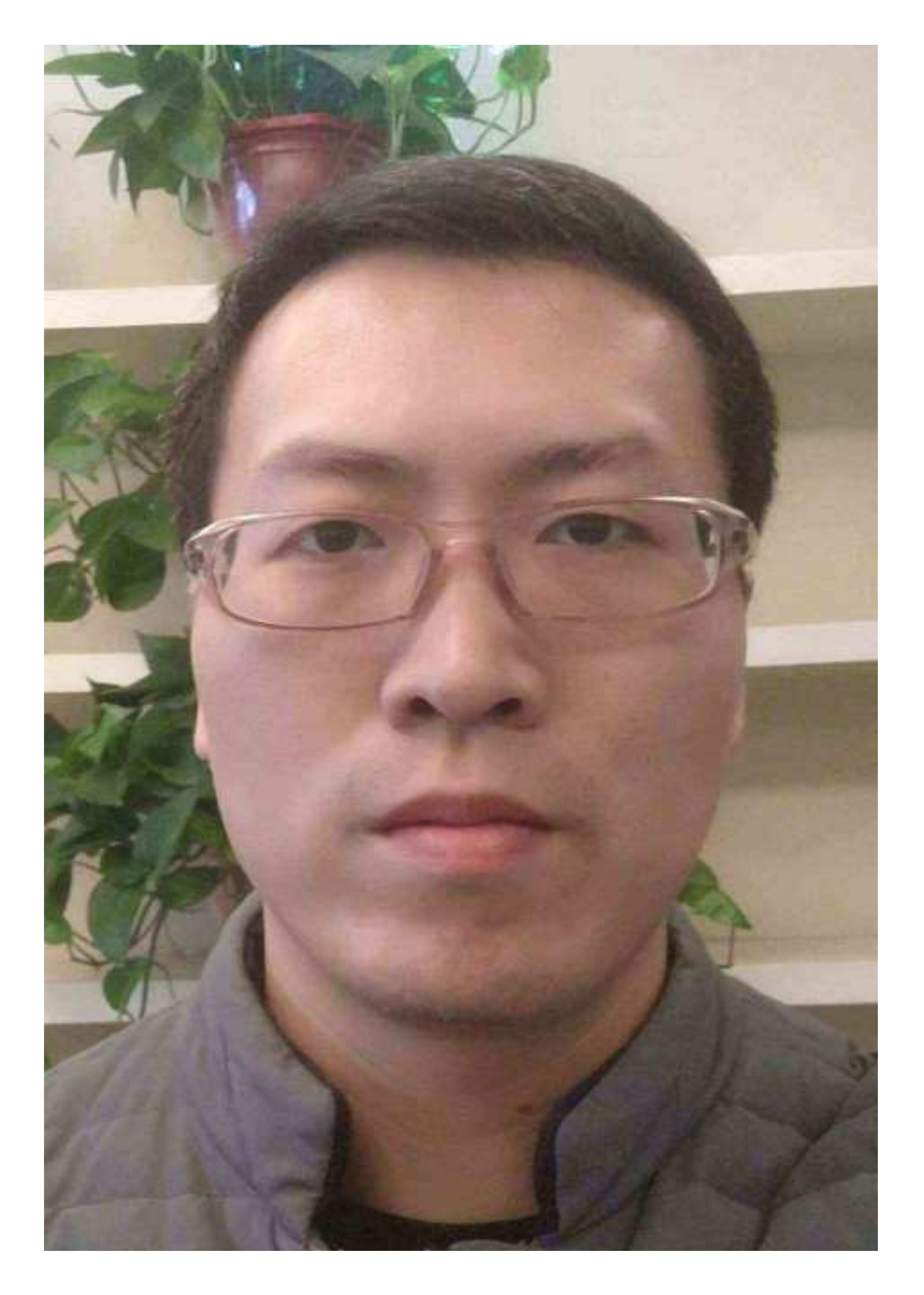}}] {Yanan Li} received his Bachelor and Master degree from Henan Normal University of China in 2004 and 2007, respectively. He is currently working towards the PhD degree in the School of Mathematics and Statistics at Xi'an Jiaotong University. Before that, he worked as a lecturer in Henan Polytechnic University from 2007 to 2017. His research interests include machine learning, federated learning, and edge-cloud intelligence.
\end{IEEEbiography}
\vspace{-1cm}

\begin{IEEEbiography}[{\includegraphics[width=1in,height=1.25in,clip,keepaspectratio]{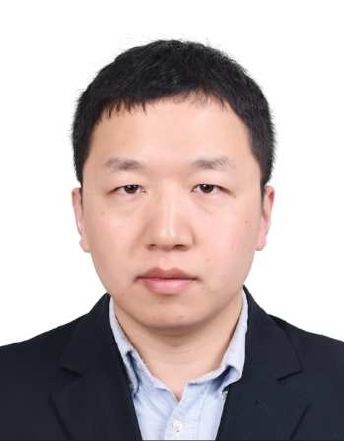}}]{Xuebin Ren} received his PhD degree in the Department of Computer Science and Technology from Xi'an Jiaotong University (XJTU), China in 2017. Currently, he is an associate professor in the School of Computer Science and Technology and a member of National Engineering Laboratory for Big Data Analytics (NEL-BDA), both at XJTU, Xi'an, China. He has been a visiting PhD student in the Department of Computing at Imperial College London, UK from 2016 to 2017. His research interests focus on data privacy protection, federated learning and privacy-preserving machine learning. He is a member of the IEEE and the ACM.
\end{IEEEbiography}
\vspace{-1cm}
\begin{IEEEbiography}[{\includegraphics[width=1in,height=1.25in, clip,keepaspectratio]{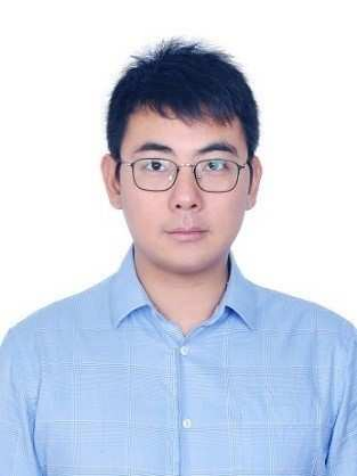}}] {Shusen Yang} received his Ph.D. degree in Computing from Imperial College London in 2014. He is a professor and director of the National Engineering Laboratory for Big Data Analytics, and deputy director of Ministry of Education(MoE) Key Lab for Intelligent Networks and Network Security, both at Xi'an Jiaotong University (XJTU), Xi'an, China. Before joining XJTU, He worked as a lecturer (assistant professor) at the University of Liverpool from 2015 to 2016, and a research associate at Intel Collaborative Research Institute (ICRI) on sustainable connected cities from 2013 to 2014. Shusen is a DAMO Academy Young Fellow, and an honorary research fellow at Imperial College London. He is a senior member of IEEE and a member of ACM. %His research focuses on distributed systems and data sciences for industrial artificial intelligence, including industrial intelligence algorithms, Edge-Cloud intelligence, and industrial security.
His research focuses on distributed systems and data sciences, and their applications in industrial scenarios, including data-driven network algorithms, distributed machine learning, Edge-Cloud intelligence, industrial internet and industrial intelligence.
\end{IEEEbiography}
\vspace{-1cm}

\newpage
\appendices

\section{Comparison Experiment of WKAFL and ZENO++}
\label{Appendix:ZENO}
In this section, we conducted experiments to compare the performance of WKAFL to that of ZENO++. Note that, ZENO++ is proposed to prevent byzantine workers for distributed ML from diverging the model, instead of a federated learning approach for privacy-preserving distributed ML.
It works as follows. Before training, the server collects some non-private data and derives the benchmark gradients by pre-training. Then, the server scores every received candidate gradient by comparing with the benchmark gradients, and picks those with high scores to update the global model. To alleviate the computational burden on the server, ZENO++ adopts a lazy update mode, in which, the server estimates the benchmark gradients every $k$ ($k\in \mathbb{N}^{+}$) iterations.

\subsection{Experiment Setup}
%In this section, our purpose is to introduce ZENO++ briefly and claim some important parameters.

%\textbf{Brief Introduction of ZENO++.}

%\textbf{Parameters Settings.}
We also used EMNIST MNIST dataset in the comparison experiment. We varied the number of classes of collected data $L_{num} \in \{ 1, 10 \}$ and the frequency of lazy update $k \in \{2, 5 \}$ on the server for ZENO++. Table \ref{Exp:Comp} shows the detailed parameter settings. Particularly, since ZENO++ is a fully asynchronous distributed ML algorithm, and the number of participant clients $K$ in each iteration is set to one. However, WKAFL adopts $K$-async FL and gradients selection and $K$ must be set to larger than one. Therefore, the number of total clients and participant clients for ZENO++ and WKAFL are set different in Table \ref{Exp:Comp}. %The reason is that their scenarios are different. Since ZENO++ is an algorithm of fully asynchronous FL, the number of participant clients $K$ in each iteration is set to one.
To guarantee the approximately equivalent staleness, the ratio $P/K$ is set to the same value for WKAFL and ZENO++.

\begin{table}[!ht]
  \renewcommand{\arraystretch}{1.3}
  \caption{Notations and parameters.}\label{Exp:Comp}
  \centering
   \begin{tabular}{cccc}
     \hline
     Algorithm & Variables & Meaning & Values \\
     \hline
     \multirow{4}*{WKAFL} & $P$ & Number of total clients & 3000 \\
     & $K$ & Number of participant gradients & 20 \\
     & $J$ & Number of iterations & 1500 \\  \hline
     \multirow{6}*{ZENO++} & $P$ & Number of total clients & 100 \\
     & $K$ & Number of participant gradients & 1  \\
     & $J$ & Number of iterations & 1500 \\ %\cline{2-4}
     & $\gamma$ & The parameter of score function & 0.1 \\
     & $\rho$ & The parameter of score function & 0.02 \\
     & $\varepsilon$ & The parameter of score function & 0.1  \\
     \hline
   \end{tabular}
\end{table}

\subsection{Experiment Results and Analysis}
\begin{figure}[!ht]
  \centering
  \includegraphics[width=0.35\textwidth]{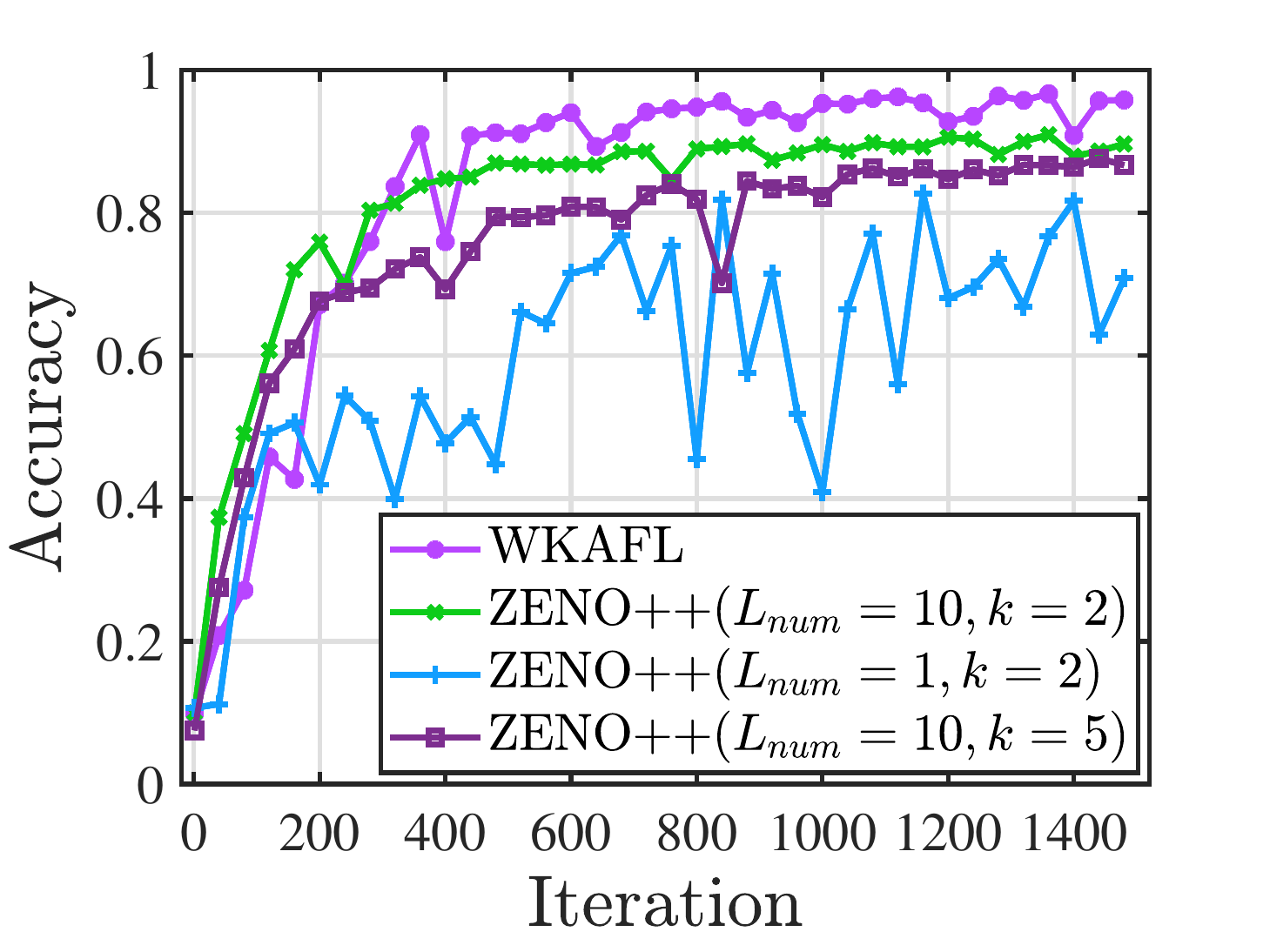}
  \caption{Comparison between WKAFL and ZENO++.}\label{Exp:ZENO}
\end{figure}

Fig. \ref{Exp:ZENO} shows the relationship between the test accuracy and the number of iterations, for ZENO++ and WKAFL on EMNIST MNIST dataset. %It indicates two aspects.
\emph{On the one hand, ZENO++ and WKAFL have similar performance only when the collected data can represent the global data distribution and the server updates the benchmark gradients frequently}. For $L_{num}=10$ and $k=2$, ZENO++ converges faster than WKAFL when the iteration number is less than 300. However, WKAFL converges to a higher final accuracy. Both ZENO++ and WKAFL can converge stably. \emph{On the other hand, WKAFL behaves much better than ZENO++ when the collected data of ZENO++ can not represent the overall data or the server has infrequent lazy update}. When $k=5$, the model utility decreases due to the fact that the benchmark gradients have larger staleness and become inaccuracy. When $L_{num}=1$, the model will greatly decrease and fluctuate heavily around a low accuracy.

It should note that,  the server has to collect some high-quality data before the collaborative training for ZENO++. However, this is infeasible in federated learning scenarios, because clients are prohibited to upload raw data due to privacy in design. Therefore, ZENO++ cannot be directly applied in federated learning scenarios. %In contrast, WKAFL proposed in this paper does not have these restrictions.

\section{Proof of Theorem \ref{stageone}}\label{proofStageone}
In this section, we provide a proof of Theorem \ref{stageone}. To convey the proof clearly, it would be necessary to prove one certain useful lemma.

\begin{lemma}\label{lemma2}
  Assume Assumption \ref{assumption3} holds.
  Then, we have:
  \begin{IEEEeqnarray}{Cl}\label{lemma2Eq}
  \mathbb{E} & [||\sum_{i=1}^{K} p_{j,i} g(w_{j,i},\xi_{j,i})||_2^2] \leq \frac{\sigma_c^2}{m} \sum_{i=1}^{K} \mathbb{E} [p_{j,i}^{2}]     \nonumber \\
  & \quad + \mathbb{E}[\sum_{i=1}^{K} p_{j,i} (\frac{M_c}{m} p_{j,i} + 1) ||\nabla F_i(w_{\tau(j)})||_2^2].
  \end{IEEEeqnarray}
\end{lemma}
\begin{IEEEproof}[Proof of Lemma \ref{lemma2}]
Observe that,
\begin{IEEEeqnarray}{Cl}
    & \mathbb{E}[||\sum_{i=1}^{K} p_{j,i} \nabla F_i(w_{\tau(j)})||_2^2] =  \mathbb{E} [ \sum_{0\leq i,k\leq K} p_{j,i}p_{j,k} \nonumber \\
    & \quad  \nabla F_i(w_{\tau(j)})\nabla F_k(w_{\tau(j)})]  \nonumber
    \\
    & \leq \frac{1}{2}\mathbb{E} [ \sum_{0\leq i,k\leq K} p_{j,i}p_{j,k}  (||\nabla F_i(w_{\tau(j)})||_2^2+||\nabla F_k(w_{\tau(j)})||_2^2)] \nonumber \\
    & = \mathbb{E} [ \sum_{0\leq i,k\leq K} p_{j,i}p_{j,k} ||\nabla F_i(w_{\tau(j)})||_2^2]  \nonumber \\
    & \leq \mathbb{E}[\sum_{i=1}^{K} p_{j,i} ||\nabla F_i(w_{\tau(j)})||_2^2] \label{xiaofang}
\end{IEEEeqnarray}

Based on $\sum_{i=1}^K p_{j,i} = 1$, we have
\begin{IEEEeqnarray}{Cl}
  \mathbb{E} & [||\sum_{i=1}^{K} p_{j,i} g(w_{j,i},\xi_{j,i})||_2^2] = \mathbb{E}[||\sum_{i=1}^{K} p_{j,i} (g(w_{j,i},\xi_{j,i})-  \nonumber \\
  & \quad \nabla F_i(w_{\tau(j)}))+ \sum_{i=1}^{K}p_{j,i} \nabla F_i(w_{\tau(j)})||_2^2]    \nonumber \\
  & = \mathbb{E}[||\sum_{i=1}^{K} p_{j,i} (g(w_{j,i},\xi_{j,i})-\nabla F_i(w_{\tau(j)}))||_2^2] +   \nonumber \\
  & \quad \mathbb{E}[||\sum_{i=1}^{K} p_{j,i}  \nabla F_i(w_{\tau(j)})||_2^2]  \label{lemma2Tag1}\\
  & \leq \sum_{i=1}^{K} \mathbb{E}[p_{j,i}^{2} ||(g(w_{j,i},\xi_{j,i})-\nabla F_i(w_{\tau(j)}))||_2^2]   \nonumber \\
  & \quad + \mathbb{E}[\sum_{i=1}^{K} p_{j,i} || \nabla F_i(w_{\tau(j)})||_2^2]   \label{lemma2Tag2} \\
  & \leq \frac{\sigma_c^2}{m} \sum_{i=1}^{K} \mathbb{E} [p_{j,i}^{2}] + \mathbb{E}[\sum_{i=1}^{K} p_{j,i} (\frac{M_c}{m} p_{j,i} + 1 )   \nonumber  \\
  & \quad ||\nabla F_i(w_{\tau(j)})||_2^2].     \label{lemma2Tag3}
\end{IEEEeqnarray}
Equation \eqref{lemma2Tag1} holds since the cross term is 0 as derived below.
\begin{IEEEeqnarray}{Cl}
  & \mathbb{E} [(\sum_{i=1}^{K} p_{j,i} (g(w_{j,i},\xi_{j,i})-\nabla F_i(w_{\tau(j)})))  (\sum_{k=1}^{K}p_{j,k} \nabla F_k(w_{\tau(j)}) )]  \nonumber  \\
  & = \sum_{i=1}^{K} \sum_{k=1}^{K} p_{j,i} p_{j,k} \mathbb{E} [(g(w_{j,i},\xi_{j,i})-\nabla F_i(w_{\tau(j)}))  \nabla F_k(w_{\tau(j)})]  \nonumber \\
  & = \sum_{i=1}^{K} \sum_{k=1}^{K} p_{j,i} p_{j,k} \mathbb{E} [(\mathbb{E}_{\xi_{j,i}}[g(w_{j,i},\xi_{j,i})]-\nabla F_i(w_{\tau(j)}))   \nonumber \\
  & \quad \nabla F_k(w_{\tau(j)})] = 0  \nonumber
\end{IEEEeqnarray}
Equation \eqref{lemma2Tag2} is derived from  Equation \eqref{xiaofang} and Equation \eqref{lemma2Tag3} is derived based on Assumption \ref{assumption3}.
\end{IEEEproof}

Based on Lemma \ref{lemma2}, we provide a proof of Theorem \ref{stageone}.

\begin{IEEEproof}[Proof of Theorem \ref{stageone}]
Based on Assumption \ref{assumption1} and update rule $w_{j+1} = w_{j} - \eta_j \sum_{l=1}^{j} \alpha^{j-l} \sum_{i=1}^K p_{l,i} g(w_{l,i},\xi_{l,i}) $, we have
\begin{IEEEeqnarray}{Cl}
  F & (w_{j+1})-F(w_{j})\leq \nabla F(w_{j})(w_{j+1}-w_{j})+\frac{L}{2}||w_{j+1}-  w_{j}||_2^2        \nonumber \\
  & = - \nabla F(w_{j}) \eta_j \sum_{l=1}^{j} \alpha^{j-l} \sum_{i=1}^K p_{l,i} g(w_{l,i},\xi_{l,i})  \nonumber \\
  & \quad + \frac{L}{2} ||\eta_j \sum_{l=1}^{j} \alpha^{j-l} \sum_{i=1}^K p_{l,i} g(w_{l,i},\xi_{l,i})||_2^2 \nonumber \\
  & = -\eta_j \sum_{l=1}^{j} \alpha^{j-l} \sum_{i=1}^K p_{l,i} \nabla F(w_j)g(w_{l,i},\xi_{l,i})
  \nonumber \\
  & \quad + \frac{L}{2} \eta_j^2 || \sum_{l=1}^{j} \alpha^{j-l} \sum_{i=1}^K p_{l,i} g(w_{l,i},\xi_{l,i})||_2^2. \label{one2}
\end{IEEEeqnarray}

Define $s_j = \sum_{l=1}^j \alpha^{j-l} = \frac{1-\alpha^j}{1-\alpha}$ and take expectation on both sides of Equation \eqref{one2}:
\begin{IEEEeqnarray}{Cl}
  \mathbb{E} & [F(w_{j+1})]-F(w_j) = -\eta_j \sum_{l=1}^{j} \alpha^{j-l} \mathbb{E} [\sum_{i=1}^K p_{l,i} \nabla F(w_j)
  \nonumber \\
  & \quad \nabla F_i (w_{\tau(l)})] + \frac{L}{2} \eta_j^2 s_j^2 \mathbb{E} [ || \sum_{l=1}^{j} \frac{\alpha^{j-l}}{s_j} \sum_{i=1}^K p_{l,i} g(w_{l,i},\xi_{l,i})||_2^2]  \nonumber  \\
  & \leq - \frac{\eta_j}{2} \sum_{l=1}^{j} \alpha^{j-l} \mathbb{E} [ \sum_{i=1}^K p_{l,i} (||\nabla F(w_{j})||_2^2 + || \nabla F_i (w_{\tau(l)})||_2^2  \nonumber \\
  & \quad - ||\nabla F(w_j) - \nabla F_i (w_{\tau(l)})||_2^2)]  \nonumber \\
  & \quad + \frac{L}{2} \eta_j^2 s_j^2 \sum_{l=1}^{j} \frac{\alpha^{j-l}}{s_j} \mathbb{E} [ || \sum_{i=1}^K p_{l,i} g(w_{l,i},\xi_{l,i})||_2^2].  \label{one3}
  \\
  & \leq - \frac{\eta_j}{2} \sum_{l=1}^{j} \alpha^{j-l} \mathbb{E} [ \sum_{i=1}^K p_{l,i} (||\nabla F(w_{j})||_2^2 + ||\nabla F_i (w_{\tau(l)})||_2^2  \nonumber \\
  & \quad - ||\nabla F(w_j) - \nabla F_i (w_{\tau(l)})||_2^2)]  + \frac{L}{2} \eta_j^2 s_j \sum_{l=1}^{j} \alpha^{j-l} \nonumber \\
  & \quad \mathbb{E}[\frac{\sigma_c^2}{m} \sum_{i=1}^{K} p_{l,i}^2 + \sum_{i=1}^{K} p_{l,i} (\frac{M_c}{m} p_{l,i} + 1 )||\nabla F_i(w_{\tau(l)})||_2^2].  \label{one4}
\end{IEEEeqnarray}

Equation \eqref{one3} follows from $2a^T b = ||a||_2^2+||b||_2^2 - ||a-b||_2^2$ and Jensen's inequality.
Equation \eqref{one4} is derived by Lemma \ref{lemma2}.

Notice that $\sum_{i=1}^K p_{l,i} = 1$, then we have
\begin{IEEEeqnarray}{Cl}
  \mathbb{E} & [F(w_{j+1})]-F(w_j) \leq \frac{L \sigma_c^2}{2m} \eta_j^2 s_j \sum_{l=1}^{j} \alpha^{j-l} \sum_{i=1}^{K} \mathbb{E} [p_{l,i}^{2}] \nonumber \\
  & \quad -\frac{\eta_j s_j}{2} ||\nabla F(w_{j})||_2^2  - \mathbb{E}[ \frac{\eta_j}{2} \sum_{l=1}^{j} \alpha^{j-l} \sum_{i=1}^K p_{l,i} (1 -    \nonumber  \\
  & \quad L \eta_j s_j (1 + \frac{M_c}{m} p_{l,i}) ) ||\nabla F_i(w_{\tau(l)})||_2^2]   \nonumber \\
  & \quad + \frac{\eta_j}{2} \sum_{l=1}^{j} \alpha^{j-l} \mathbb{E} [\sum_{i=1}^K p_{l,i} ||\nabla F(w_j) - \nabla F_i (w_{\tau(l)})||_2^2]  \nonumber
  \\
  & \leq \frac{L \sigma_c^2}{2m} \eta_j^2 s_j \sum_{l=1}^{j} \alpha^{j-l} \sum_{i=1}^{K} \mathbb{E} [p_{l,i}^{2}] - \frac{\eta_j s_j}{2} ||\nabla F(w_{j})||_2^2  \nonumber  \\
  & \quad + \frac{\eta_j}{2} \sum_{l=1}^{j} \alpha^{j-l} \underbrace{\mathbb{E} [\sum_{i=1}^K p_{l,i} ||\nabla F(w_j) - \nabla F_i (w_{\tau(l)})||_2^2]}_{\mathcal{A}}  \nonumber \\
  & \quad - \mathbb{E}[ \frac{\eta_j}{4} \sum_{l=1}^{j} \alpha^{j-l} \sum_{i=1}^{K} p_{l,i} ||\nabla F_i(w_{\tau(l)})||_2^2]). \label{one5}
\end{IEEEeqnarray}

Equation \eqref{one5} is derived because learning rate $\eta_j$ satisfies $\eta_j \leq \frac{1}{2 L s_j (1 + \frac{M_c}{m} ) } \leq \frac{1}{2 L s_j (1 + \frac{M_c}{m} p_{l,i}) }$.
Now, we aim to bound the term $\mathcal{A}$.

With respect to term $\mathcal{A}$,
\begin{IEEEeqnarray}{Cl}
  \mathcal{A} & = \mathbb{E}[\sum_{i=1}^{K} p_{l,i} ||\nabla F(w_j) - \nabla F_i (w_j)  \nonumber  \\
  & \quad + \nabla F_i(w_j) - \nabla F_i(w_{\tau(l)})||_2^2] \nonumber
  \\
  & \leq 2 \mathbb{E}[\sum_{i=1}^{K} p_{l,i} ||\nabla F(w_j) - \nabla F_i (w_j)||_2^2]  \nonumber
  \\
  & \quad + 2 \mathbb{E}[\sum_{i=1}^{K} p_{l,i}||\nabla F_i(w_j)-\nabla F_i(w_{\tau(l)})||_2^2] \nonumber  \\
  & \leq 2 G^2 + 2L^2 \underbrace{\mathbb{E}[\sum_{i=1}^{K} p_{l,i} ||w_j - w_{l-\tau_{l,i}}||_2^2]}_{\mathcal{B}}; \label{partA}
\end{IEEEeqnarray}
Equation \ref{partA} is derived based on Assumptions \ref{assumption1} and \ref{assumption3}.

Define $\tau_l = \max (1, l-\tau_{max})$. With respect to term $\mathcal{B}$,
\begin{IEEEeqnarray}{Cl}
  & \mathcal{B} \leq \mathbb{E}[\sum_{i=1}^{K} p_{l,i} ||w_j - w_{l-\tau_{l,i}}||_2^2]  \nonumber  \\
  & = \mathbb{E}[\sum_{i=1}^{K} p_{l,i} ||\sum_{t=l-\tau_{l,i}}^{j-1} (w_{t+1} - w_{t}) ||_2^2]  \nonumber \\
  & \leq  \mathbb{E}[\sum_{i=1}^{K} p_{l,i} (j-\tau_l) \sum_{t=\tau_l}^{j-1} \eta_t^2 || \sum_{k=1}^{t} \alpha^{t-k}   \nonumber \\
  & \quad \sum_{q=1}^K p_{k,q} g(w_{k,q},\xi_{k,q}) ||_2^2]  \nonumber \\
  & \leq (j-\tau_{l}) \mathbb{E}[ \sum_{t=\tau_l}^{j-1} \eta_t^2 s_t \sum_{k=1}^{t} \alpha^{t-k} || \sum_{i=1}^K p_{k,i} g(w_{k,i},\xi_{k,i}) ||_2^2]  \nonumber  \label{partC}  \\
  & \leq (j-\tau_{l}) \mathbb{E}[ \sum_{t=\tau_l}^{j-1} \eta_t^2 s_t \sum_{k=1}^{t} \alpha^{t-k} (\frac{\sigma_c^2}{m} \sum_{i=1}^{K} p_{k,i}^{2}  \nonumber \\
  & \quad + \sum_{i=1}^{K} p_{k,i}(\frac{M_c}{m} p_{k,i} +1) || \nabla F_i(w_{\tau(k)})||_2^2 )]. \label{partB}
\end{IEEEeqnarray}
Equation \eqref{partB} is derived from Lemma \ref{lemma2}.

By replacing Equations \eqref{partA}, \eqref{partB} into Equation \eqref{one5}, we have
\begin{IEEEeqnarray}{Cl}
  \mathbb{E} & [F(w_{j+1})] - F(w_{j}) \leq \frac{L \sigma_c^2}{2m} \eta_j^2 s_j \sum_{l=1}^{j} \alpha^{j-l} \sum_{i=1}^{K} \mathbb{E} [p_{l,i}^{2}] - \frac{\eta_j s_j}{2} \nonumber  \\
  & ||\nabla F(w_{j})||_2^2 + \eta_j G^2 \sum_{l=1}^{j} \alpha^{j-l} + \eta_j \sum_{l=1}^j \alpha^{j-l} L^2 (j-\tau_{l}) \nonumber \\
  & \sum_{t=\tau_l}^{j-1} \eta_t^2 s_t \sum_{k=1}^{t} \alpha^{t-k}
  \frac{\sigma_c^2}{m} \mathbb{E}[\sum_{i=1}^{K} p_{k,i}^{2}] + \eta_j L^2 \sum_{l=1}^{j}  (j-\tau_{l}) \nonumber \\
  & \alpha^{j-l} \mathbb{E}[ \sum_{t=\tau_l}^{j-1} \eta_t^2 s_t \sum_{k=1}^{t} \alpha^{t-k} \sum_{i=1}^{K} p_{k,i} (1 + \frac{M_c}{m} p_{k,i}) \nonumber \\
  & ||\nabla F_i(w_{\tau(k)})||_2^2)] ) )  - \frac{\eta_j}{4} \mathbb{E}[ \sum_{l=1}^{j} \alpha^{j-l} \sum_{i=1}^{K} p_{l,i} ||\nabla F_i(w_{\tau(l)})||_2^2])   \nonumber
  \\
  & \leq \frac{L \sigma_c^2}{2m} \eta_j^2 s_j \sum_{l=1}^{j} \alpha^{j-l} \sum_{i=1}^{K} \mathbb{E} [p_{l,i}^{2}] - \frac{\eta_j s_j}{2} ||\nabla F(w_{j})||_2^2  \nonumber  \\
  & + \eta_j s_j G^2 + \eta_j \sum_{l=1}^{j} \alpha^{j-l}  L^2 (j-\tau_{l}) \sum_{t=\tau_l}^{j-1} \eta_t^2 s_t \sum_{k=1}^{t} \alpha^{t-k} \nonumber \\
  & \frac{\sigma_c^2}{m} \mathbb{E}[\sum_{i=1}^{K} p_{k,i}^{2}] + \eta_j L^2  \sum_{l=1}^{j-1} \sum_{t=l}^{j-1}(j-\tau_{t}) \alpha^{j-t} \sum_{k=l}^{t} \eta_{k}^2 s_{k} \alpha^{k-l} \nonumber \\
  & \mathbb{E}[\sum_{i=1}^{K} p_{l,i} (1 + \frac{M_c}{m}p_{l,i}) ||\nabla F_i(w_{\tau(l)})||_2^2)] ) ) \nonumber \\
  & - \frac{\eta_j}{4} \mathbb{E}[ \sum_{l=1}^{j} \alpha^{j-l} \sum_{i=1}^{K} p_{l,i} ||\nabla F_i(w_{\tau(l)})||_2^2])    \nonumber
  \\
  & \leq \frac{L \sigma_c^2}{2m} \eta_j^2 s_j \sum_{l=1}^{j} \alpha^{j-l} \sum_{i=1}^{K} \mathbb{E} [p_{l,i}^{2}] - \frac{\eta_j s_j}{2} ||\nabla F(w_{j})||_2^2  \nonumber  \\
  & + \eta_j s_j G^2 + \eta_j \sum_{l=1}^{j} \alpha^{j-l}  L^2  \sum_{t=\tau_l}^{j-1} (j-\tau_{t}) \eta_t^2 s_t \sum_{k=1}^{t} \alpha^{t-k}  \nonumber \\
  & \frac{\sigma_c^2}{m} \mathbb{E}[\sum_{i=1}^{K} p_{k,i}^{2}] - \eta_j \sum_{l=1}^{j} (\sum_{i=1}^{K} \alpha^{j-l} p_{l,i}/4 -L^2 \sum_{t=l}^{j} (j-\tau_{t}) \nonumber \\
  & \alpha^{j-t}  \sum_{k=l}^{t} \eta_{k}^2 s_{k} \alpha^{k-l} \sum_{i=1}^{K} p_{l,i} ( 1 + \frac{M_c}{m} p_{l,i} ) ||\nabla F_i(w_{\tau(l)})||_2^2.  \nonumber
\end{IEEEeqnarray}
Since the learning rate satisfies that $L^2  \sum_{t=l}^{j} (J-\tau_{t}) \alpha^{-t} \sum_{k=l}^{t} \eta_{k}^2 s_{k} \alpha^{k} ( 1 + \frac{M_c}{m} ) < \frac{1}{4}$, then we have
\begin{IEEEeqnarray}{Cl}
  \alpha^{j-l}/4 -L^2  \sum_{t=l}^{j} (j-\tau_{t}) \alpha^{j-t} \sum_{k=l}^{t} \eta_{k}^2 s_{k} \alpha^{k-l}
   ( 1 + \frac{M_c}{m} p_{l,i}) > 0.   \nonumber
\end{IEEEeqnarray}

%By the law of total expectation,
%\begin{IEEEeqnarray}{Cl}
%  \mathbb{E} [||\nabla F_i & (w_{\tau(l)})||_2^2] = q_{l,i} \mathbb{E} [||\nabla F_i (w_{\tau(l)})||_2^2 | \tau(l)=l]  \nonumber  \\
%  & \quad + (1-q_{l,i}) \mathbb{E} [||\nabla F_i (w_{\tau(l)})||_2^2 | \tau(l) \neq l]   \nonumber  \\
%  & \geq q_0 \mathbb{E} [||\nabla F_i (w_l)||_2^2]
%\end{IEEEeqnarray}

Then, we have
\begin{IEEEeqnarray}{Cl}
  \mathbb{E} & [F(w_{j+1})] - F(w_{j}) \leq \frac{L \sigma_c^2}{2m} \eta_j^2 s_j \sum_{l=1}^{j} \alpha^{j-l} \sum_{i=1}^{K} \mathbb{E} [p_{l,i}^{2}] \nonumber  \\
  & - \frac{\eta_j s_j}{2} ||\nabla F(w_{j})||_2^2 + \eta_j s_j G^2 + \eta_j \sum_{l=1}^{j} \alpha^{j-l}  L^2 (j-\tau_{l}) \nonumber  \\
  &  \sum_{t=\tau_l}^{j-1} \eta_t^2 s_t   \sum_{k=1}^{t} \alpha^{t-k}
  \frac{\sigma_c^2}{m} \mathbb{E}[\sum_{i=1}^{K} p_{k,i}^{2}].     \nonumber
\end{IEEEeqnarray}

Taking summation with respect to $j \in \{1,2,\cdots,J\}$ on both sides, we obtain the following result:
\begin{IEEEeqnarray}{Cl}
  & \frac{1}{J} \sum_{j=1}^J \frac{\eta_j s_j}{2} \mathbb{E}[||\nabla F(w_j)||_2^2] \leq  \frac{1}{J} \sum_{j=1}^J  (\frac{L \sigma_c^2}{2m} \eta_j^2 s_j \sum_{l=1}^{j} \alpha^{j-l} \nonumber  \\
  & \mathbb{E} [\sum_{i=1}^{K} p_{l,i}^{2}] + \eta_j s_j G^2 + \eta_j \sum_{l=1}^{j} \alpha^{j-l}  L^2 (j-\tau_{l}) \sum_{t=\tau_l}^{j-1} \eta_t^2 s_t \sum_{k=1}^{t} \alpha^{t-k}  \nonumber  \\
  & \quad  \frac{\sigma_c^2}{m} \mathbb{E}[\sum_{i=1}^{K} p_{k,i}^{2}]) + \frac{F(w_1)-F(w^*)}{J}. \nonumber
\end{IEEEeqnarray}
\end{IEEEproof}

\section{Proof of Theorem \ref{stagetwo}}\label{proofStagetwo}
\begin{IEEEproof}[Proof of Theorem \ref{stagetwo}]
Based on Assumption \ref{assumption1} and update rule $w_{j+1} = w_{j} - \eta_j \sum_{l=1}^{j} \alpha^{j-l} \sum_{i=1}^K p_{l,i} \bar{g}(w_{l,i},\xi_{l,i}) $, we have
\begin{IEEEeqnarray}{Cl}
  F & (w_{j+1})-F(w_{j})\leq \nabla F(w_{j})(w_{j+1}-w_{j})+\frac{L}{2}||w_{j+1}-  w_{j}||_2^2        \nonumber \\
  & = - \nabla F(w_{j}) \eta_j \sum_{l=1}^{j} \alpha^{j-l} \sum_{i=1}^K p_{l,i} \bar{g} (w_{l,i},\xi_{l,i})  \nonumber \\
  & \quad + \frac{L}{2} ||\eta_j \sum_{l=1}^{j} \alpha^{j-l} \sum_{i=1}^K p_{l,i} \bar{g} (w_{l,i},\xi_{l,i})||_2^2 \nonumber \\
  & = -\eta_j \sum_{l=1}^{j} \alpha^{j-l} \sum_{i=1}^K p_{l,i} \nabla F(w_j) \bar{g}(w_{l,i},\xi_{l,i})
  \nonumber \\
  & \quad + \frac{L}{2} \eta_j^2 || \sum_{l=1}^{j} \alpha^{j-l} \sum_{i=1}^K p_{l,i} \bar{g}(w_{l,i},\xi_{l,i})||_2^2   \nonumber
  \\
  & \quad = - \frac{\eta_j}{2} \sum_{l=1}^{j} \alpha^{j-l} \sum_{i=1}^K p_{l,i} (||\nabla F(w_{j})||_2^2 \nonumber \\
  &  \quad + ||\bar{g}(w_{l,i}, \xi_{l,i})||_2^2 - ||\nabla F(w_j) - \bar{g}(w_{l,i}, \xi_{l,i})||_2^2)    \nonumber \\
  &  \quad + \frac{L}{2} \eta_j^2 || \sum_{l=1}^{j} \alpha^{j-l} \sum_{i=1}^K p_{l,i} \bar{g}(w_{l,i},\xi_{l,i})||_2^2.  \label{two1}
\end{IEEEeqnarray}

Define $s_j = \sum_{l=1}^j \alpha^{j-l} = \frac{1-\alpha^j}{1-\alpha}$ and take expectation on both sides of Equation \eqref{two1}:
\begin{IEEEeqnarray}{Cl}
  \mathbb{E} & [F(w_{j+1})]-F(w_j) = - \frac{\eta_j}{2}  \sum_{l=1}^{j} \alpha^{j-l} \mathbb{E}[\sum_{i=1}^K p_{l,i} (||\nabla F(w_{j})||_2^2  \nonumber \\
  & \quad + ||\bar{g}(w_{l,i}, \xi_{l,i})||_2^2 -    ||\nabla F(w_j) - \bar{g}(w_{l,i}, \xi_{l,i})||_2^2)]
  \nonumber \\
  & \quad + \frac{L}{2} \eta_j^2 s_j^2 \mathbb{E} [ || \sum_{l=1}^{j} \frac{\alpha^{j-l}}{s_j} \sum_{i=1}^K p_{l,i} \bar{g}(w_{l,i},\xi_{l,i})||_2^2]  \nonumber
  \\
  & \leq - \frac{\eta_j}{2} \sum_{l=1}^{j} \alpha^{j-l} \mathbb{E} [ \sum_{i=1}^K p_{l,i} (||\nabla F(w_{j})||_2^2 + ||\bar{g}(w_{l,i}, \xi_{l,i})||_2^2  \nonumber \\
  & \quad - ||\nabla F(w_j) - \bar{g}(w_{l,i}, \xi_{l,i})||_2^2)]  \nonumber \\
  & \quad + \frac{L}{2} \eta_j^2 s_j \sum_{l=1}^{j} \alpha^{j-l} \mathbb{E} [ || \sum_{i=1}^K p_{l,i} \bar{g}(w_{l,i},\xi_{l,i})||_2^2]   \label{two2}
  \\
  & \leq - \frac{\eta_j}{2} \sum_{l=1}^{j} \alpha^{j-l} \mathbb{E} [ \sum_{i=1}^K p_{l,i} (||\nabla F(w_{j})||_2^2 + ||\bar{g}(w_{l,i}, \xi_{l,i})||_2^2  \nonumber \\
  & \quad - ||\nabla F(w_j) - \bar{g}(w_{l,i}, \xi_{l,i})||_2^2)]  \nonumber \\
  & \quad + \frac{L}{2} \eta_j^2 s_j \sum_{l=1}^{j} \alpha^{j-l} \mathbb{E} [ \sum_{i=1}^K p_{l,i} ||\bar{g}(w_{l,i},\xi_{l,i})||_2^2]   \label{two3}
  \\
  & = - \frac{\eta_j s_j}{2} ||\nabla F(w_j)||_2^2 - \frac{\eta_j}{2} \mathbb{E} [ \sum_{l=1}^{j} \sum_{i=1}^{K} \alpha^{j-l} p_{l,i} (1 - L \eta_j s_j)   \nonumber   \\
  & \quad ||\bar{g}(w_{l,i}, \xi_{l,i})||_2^2 ] + \frac{\eta_j}{2} \sum_{l=1}^{j} \alpha^{j-l}  \nonumber   \\
  & \quad \mathbb{E} [ \sum_{i=1}^{K} p_{l,i} ||\nabla F(w_j) - \bar{g}(w_{l,i}, \xi_{l,i})||_2^2 ]  \nonumber \\
  & \leq - \frac{\eta_j s_j}{2} ||\nabla F(w_j)||_2^2 + \nonumber   \\
  & \quad \frac{\eta_j}{2} \sum_{l=1}^{j} \alpha^{j-l} \underbrace{\mathbb{E} [ \sum_{i=1}^{K} p_{l,i} ||\nabla F(w_j) - \bar{g}(w_{l,i}, \xi_{l,i})||_2^2 ]}_{\mathcal{C}}. \label{two4}
\end{IEEEeqnarray}

Equations \eqref{two2} and \eqref{two3} are derived based on Jensen's inequality.
Equation \eqref{two4} is derived because the learning rate satisfies that $\eta_j \leq \frac{1}{L s_j}$.

With respect to term $\mathcal{C}$,
\begin{IEEEeqnarray}{Cl}
  \mathcal{C} & = \mathbb{E} [ \sum_{i=1}^{K} p_{l,i} ||\nabla F(w_j) - \nabla F(w_l) + \nabla F(w_l) - \bar{g}(w_l)   \nonumber  \\
  & \quad  + \bar{g}(w_l) - \bar{g}(w_{l,i}, \xi_{l,i})||_2^2 ]  \nonumber  \\
  & \leq 3||\nabla F(w_j) - \nabla F(w_l)||_2^2 + 3 ||\nabla F(w_l) - \bar{g}(w_l)||_2^2  \nonumber \\
  & \quad + 3\mathbb{E}[||\bar{g}(w_l) - \bar{g}(w_{l,i}, \xi_{l,i})||_2^2]  \nonumber   \\
  & \leq 3L^2 ||\sum_{t=l}^{j-1} \eta_{t}\sum_{k=1}^t \alpha^{t-k} \sum_{i=1}^{K} p_{k,i} \bar{g} (w_{k,i}, \xi_{k,i})||_2^2    \nonumber  \\
  & \quad + 3\sigma_c^2 + 3 M_e ||\bar{g}(w_l)||_2^2 + 3 \sigma_e^2  \nonumber \\
  & \leq 3L^2 \sum_{t=l}^{j-1} (j-l) \eta_{t}^2 s_t \sum_{k=1}^t \alpha^{t-k} \sum_{i=1}^{K} p_{k,i} ||\bar{g}(w_{k,i}, \xi_{k,i})||_2^2    \nonumber  \\
  & \quad + 3\sigma_c^2 + 3 M_e ||\bar{g}(w_l)-\nabla F(w_l) + \nabla F(w_l)||_2^2 + 3 \sigma_e^2  \nonumber  \\
  & \leq 3L^2 \sum_{t=l}^{j-1} (j-l) \eta_{t}^2 s_t \sum_{k=1}^t \alpha^{t-k} \sum_{i=1}^{K} p_{k,i} B^2||\bar{g}(w_{k})||_2^2    \nonumber  \\
  & \quad + 3\sigma_c^2 + 6 M_e \sigma_c^2 + 6 M_e ||\nabla F(w_l)||_2^2 + 3 \sigma_e^2 \label{partC1}
  \\
  & \leq 3L^2 \sum_{t=l}^{j-1} (j-l) \eta_{t}^2 s_t \sum_{k=1}^t \alpha^{t-k} B^2 (2||\nabla F(w_k)||_2^2+2\sigma_c^2)    \nonumber  \\
  & \quad + 3\sigma_c^2 + 6 M_e \sigma_c^2 + 6 M_e ||\nabla F(w_l)||_2^2 + 3 \sigma_e^2 \nonumber  \\
  & \leq 6L^2 B^2 \sum_{t=l}^{j-1}(j-l)\eta_t^2 s_t \sum_{k=1}^t \alpha^{t-k} ||\nabla F(w_k)||_2^2  \nonumber  \\
  & \quad + 6 M_e ||\nabla F(w_l)||_2^2 + 6L^2 \sigma_c^2 B^2 \sum_{t=l}^{j-1}(j-l)\eta_t^2 s_t^2 +  3\sigma_c^2   \nonumber  \\
  & \quad + 6 M_e \sigma_c^2 + 3 \sigma_e^2.  \label{partC2}
\end{IEEEeqnarray}

By replacing Equation \eqref{partC2}  into Equation \eqref{two4}, we have
\begin{IEEEeqnarray}{Cl}
  \mathbb{E} & [F(w_{j+1})]-F(w_j) \leq - \frac{\eta_j s_j}{2} ||\nabla F(w_j)||_2^2 + \nonumber   \\
  & \quad \frac{\eta_j}{2} \sum_{l=1}^{j} \alpha^{j-l} (3\sigma_c^2 + 6 M_e \sigma_c^2 + 6 M_e ||\nabla F(w_l)||_2^2 + 3 \sigma_e^2) \nonumber \\
  & \quad + 3 \eta_j L^2 B^2 \sum_{l=1}^{j} \alpha^{j-l} \sum_{t=l}^{j-1}(j-l)\eta_t^2 s_t \sum_{k=1}^t \alpha^{t-k} ||\nabla F(w_k)||_2^2    \nonumber  \\
  & \quad + 3 \eta_j L^2 \sigma_c^2 B^2 \sum_{l=1}^{j} \alpha^{j-l} \sum_{t=l}^{j-1}(j-l)\eta_t^2 s_t^2 \nonumber  \\
  & = - \frac{\eta_j s_j}{2} ||\nabla F(w_j)||_2^2 + \frac{3\eta_j s_j}{2} (\sigma_c^2 + 2M_e\sigma_c^2+\sigma_e^2)  \nonumber  \\
  & \quad + 3\eta_j M_e \sum_{l=1}^{j} \alpha^{j-l} ||\nabla F(w_l)||_2^2  \nonumber  \\
  & \quad + 3 \eta_j L^2 B^2 \sum_{l=1}^{j} \sum_{t=l}^{j-1} \sum_{k=1}^{t} \alpha^{j-k} (j-k) \eta_t^2 s_t \alpha^{t-l} ||\nabla F(w_l)||_2^2  \nonumber  \\
  & \quad + 3 \eta_j L^2 \sigma_c^2 B^2 \sum_{l=1}^{j} \alpha^{j-l} \sum_{t=l}^{j-1}(j-l)\eta_t^2 s_t^2. \nonumber
\end{IEEEeqnarray}

Taking summation with respect to $j$ on both sides, we obtain
\begin{IEEEeqnarray}{Cl}
  F & (w^*) - F(w_1) \leq  \sum_{j=1}^J (\frac{3\eta_j s_j}{2} (\sigma_c^2 + 2M_e\sigma_c^2+\sigma_e^2) \nonumber  \\
  & \quad + 3 \eta_j L^2 \sigma_c^2 B^2 \sum_{l=1}^{j} \alpha^{j-l} \sum_{t=l}^{j-1}(j-l)\eta_t^2 s_t^2)  \nonumber  \\
  & \quad - \sum_{j=1}^J \frac{\eta_j s_j}{2}||\nabla F(w_j)||_2^2 +  \sum_{j=1}^J 3\eta_j \sum_{l=1}^{j} \alpha^{j-l}  (M_e+ \nonumber  \\
  & \quad L^2 B^2  \sum_{t=l}^{j-1} \sum_{k=1}^{t} \alpha^{t-k} (j-k) \eta_t^2 s_t) ||\nabla F(w_l)||_2^2 \nonumber
  \\
  & = \sum_{j=1}^J (\frac{3\eta_j s_j}{2} (\sigma_c^2 + 2M_e\sigma_c^2+\sigma_e^2) + 3 \eta_j L^2 \sigma_c^2 B^2 \nonumber  \\
  & \quad \sum_{l=1}^{j} \alpha^{j-l} \sum_{t=l}^{j-1}(j-l)\eta_t^2 s_t^2)  - \sum_{j=1}^J \frac{\eta_j s_j}{2}||\nabla F(w_j)||_2^2 \nonumber  \\
  & \quad +  \sum_{j=1}^J \sum_{l=j}^{J} 3 M_e \eta_l \alpha^{l-j} ||\nabla F(w_j)||_2^2+   \sum_{j=1}^{J} 3L^2 B^2  \sum_{l=j}^{J} \eta_l \nonumber  \\
  & \quad \sum_{t=j}^{l-1} \eta_t^2 s_t \sum_{k=1}^{t} (l-k) \alpha^{l+t-j-k}  ||\nabla F(w_j)||_2^2. \label{two5}
\end{IEEEeqnarray}
Equation \eqref{two5} amounts to:
\begin{IEEEeqnarray}{Cl}
  \frac{1}{J} & \sum_{j=1}^J (\frac{\eta_j s_j}{2} - \sum_{l=j}^{J} 3 M_e \eta_l \alpha^{l-j} -   3L^2 B^2  \sum_{l=j}^{J} \eta_l \sum_{t=j}^{l-1} \eta_t^2 s_t \nonumber  \\
  & \quad \sum_{k=1}^{t} (l-k) \alpha^{l+t-j-k} ) ||\nabla F(w_j)||_2^2  \nonumber  \\
  & \leq \frac{1}{J} \sum_{j=1}^J (\frac{3\eta_j s_j}{2} (\sigma_c^2 + 2M_e\sigma_c^2+\sigma_e^2) + 3 \eta_j L^2 \sigma_c^2 B^2 \nonumber  \\
  & \quad \sum_{l=1}^{j} \alpha^{j-l} \sum_{t=l}^{j-1}(j-l)\eta_t^2 s_t^2) + \frac{F(w_1) - F(w^*)}{J}. \nonumber
\end{IEEEeqnarray}
\end{IEEEproof}

\section{Minimal Estimation of Averaged Staleness}
\begin{theorem}\label{experiproof}
In $K$-async FL, assume that there are totally $P$ clients participating in the training process and $K$ clients uploading their gradients at one iteration. Then, after $P/K$ iterations, the averaged staleness of gradients is more than $\frac{K}{2P}(1+[P/K])*[P/K]$.
\end{theorem}
\begin{IEEEproof}[Proof of Theorem \ref{experiproof}]
At $j$-th iteration, define a series of numbers from small to large $\tau_{j,1}, \tau_{j,2}, \cdots, \tau_{j,P}$ as the staleness of $P$ clients' gradients. After $P/K$ iterations, the first $K$ numbers $\tau_{j,1}, \tau_{j,2}, \cdots, \tau_{j,K}$ are 1. And the second $K$ numbers $\tau_{j,K+1}, \tau_{j,K+2}, \cdots, \tau_{j,2K}$ cannot be smaller than 2 because the gradients have been calculated for at least two iterations. Then, we have:
\begin{IEEEeqnarray}{Cl}
\tau_{j,1} & + \tau_{j,2} + \cdots + \tau_{j,P} = K+\tau_{j,K+1}+ \tau_{j,K+2}+ \cdots+ \tau_{j,P} \nonumber \\
& \geq K+\underbrace{2+2+\cdots+2}_{K}+\tau_{j,2K+1}+\cdots+\tau_{j,P} \nonumber \\
& \geq K+2K+\cdots+[P/K]*K \nonumber\\
& +([P/K]+1)*(P-[P/K]*K)  \nonumber \\
& \geq K+2K+\cdots+[P/K]*K  \nonumber \\
& \geq \frac{K}{2}(1+[P/K])*[P/K];  \nonumber
\end{IEEEeqnarray}
Therefore, the least averaged staleness is:
\begin{IEEEeqnarray}{Cl}
  (\tau_{j,1} & + \tau_{j,2} + \cdots + \tau_{j,P})/P \geq \frac{K}{2P}(1+[P/K])*[P/K]. \nonumber
\end{IEEEeqnarray}
\end{IEEEproof}
\begin{remark}
  When $P$ is much higher than $K$, the lower bound of averaged staleness is around $\frac{P}{2K}$.
\end{remark}

\end{document}